%% file: main_camera.tex
\documentclass[10pt,twocolumn,letterpaper]{article}

\usepackage{wacv}              % To produce the CAMERA-READY version

\definecolor{wacvblue}{rgb}{0.21,0.49,0.74}
\usepackage[pagebackref,breaklinks,colorlinks,allcolors=wacvblue]{hyperref}

\usepackage{amsmath,amssymb,amsfonts,dsfont}
\usepackage{algorithmic}
\usepackage{graphicx,color}
\usepackage{textcomp}
\usepackage{xcolor}
\usepackage{algorithm,algorithmic}
\usepackage{arydshln}
\usepackage{booktabs}
\usepackage{float} 
\usepackage{mathtools}
\usepackage{caption}
\usepackage{subcaption}

\def\wacvPaperID{1536} % *** Enter the WACV Paper ID here
\def\confName{WACV}
\def\confYear{2026}

\newcommand*{\Scale}[2][4]{\scalebox{#1}{$#2$}}%
\title{Unsupervised Modular Adaptive Region Growing and RegionMix Classification for Wind Turbine Segmentation}

\author{Raül Pérez-Gonzalo$^{1,3,\dagger}$, \hspace{0.3cm}Riccardo Magro$^{1,2,\dagger}$, \hspace{0.3cm}Andreas Espersen$^{3}$, \hspace{0.3cm}Antonio Agudo$^{1}$\\
$^{1}$Institut de Robòtica i Informàtica Industrial, CSIC-UPC, Barcelona, Spain\\$^{2}$Politecnico di Milano, Milan, Italy\\$^{3}$Wind Power LAB, Copenhagen, Denmark \\
{\tt\small \{rpg,ace\}@windpowerlab.com,  riccardo.magro@mail.polimi.it, aagudo@iri.upc.edu}
}

\begin{document}
\maketitle
\input{WACV26/chapters/Abstract}
\input{WACV26/chapters/Introduction}

\input{WACV26/chapters/RelatedWork}
\input{WACV26/chapters/Methods}
\input{WACV26/chapters/Results}
\input{WACV26/chapters/Conclusion}

{
    \small
    \bibliographystyle{ieeenat_fullname}
    \bibliography{main}
}

\input{supplementary_arxiv}

\end{document}

%% file: WACV26/chapters/Abstract.tex
\begin{abstract}
%Periodic inspection of wind turbines are crucial for ensuring its operation, as even minor surface defects impact significantly aerodynamic efficiency, power output, and blade longevity. The foundation of an automated assessment system is the precise recognition of wind turbine blades within images, a challenge referred to image segmentation. State-of-the-art image segmentation algorithms rely on end-to-end learning-based pixel-dense methods to identify object boundaries. Their weakness lies in the huge amount of required annotated data. To mitigate data-reliance, we propose reducing the complex pixel-wise problem to a simplified binary classification via a region-based classifier. Regions are synthesized through a fully unsupervised and interpretable Modular-Adaptive Region-Growing strategy. The precision of the identified regions stem from a novel Adaptive Thresholding method, setting thresholds based on individual image traits, and a Region Merging algorithm, fusing smaller pieces into more meaningful, coherent segments. The region classification performance is boosted by RegionMix which blends distinct regions to produce new mixed-regions. This comprehensive framework achieves top-performing accuracy and shows great generalization by consistently segmenting blades across different windfarms.
Reliable operation of wind turbines requires frequent inspections, as even minor surface damages can degrade aerodynamic performance, reduce energy output, and accelerate blade wear. Central to automating these inspections is the accurate segmentation of turbine blades from visual data. This task is traditionally addressed through dense, pixel-wise deep learning models. However, such methods demand extensive annotated datasets, posing scalability challenges. In this work, we introduce an annotation-efficient segmentation approach that reframes the pixel-level task into a binary region classification problem. Image regions are generated using a fully unsupervised, interpretable Modular Adaptive Region Growing technique, guided by image-specific Adaptive Thresholding and enhanced by a Region Merging process that consolidates fragmented areas into coherent segments. To improve generalization and classification robustness, we introduce RegionMix, an augmentation strategy that synthesizes new training samples by combining distinct regions. Our framework demonstrates state-of-the-art segmentation accuracy and strong cross-site generalization by consistently segmenting turbine blades across distinct windfarms. \begingroup % Start a group to localize changes
\renewcommand\thefootnote{\(\dagger\)} % Set the footnote mark to a dagger symbol
\footnotetext{These authors contributed equally.}
\endgroup % End the group, restoring the default footnote mark (usually numbers)
\end{abstract}

% df
% gfgf

%% file: WACV26/chapters/Introduction.tex
\vspace{-0.3cm}
\section{Introduction}
\vspace{-0.15cm}

Wind energy is a crucial component in the shift towards renewable energy, but its efficiency is hindered by the high cost of operation and maintenance, which accounts for 30\% of the total energy production cost~\cite{opex30}. Wind turbine blades, essential yet vulnerable components, are frequently subjected to damage due to harsh environmental conditions~\cite{maintenance}, making their timely inspection and repair imperative. Traditional manual inspection methods are labor-intensive, costly, and require turbine downtime. Drone-based inspections have emerged as a promising alternative, capturing extensive high-resolution imagery~\cite{PerezGonzaloIcip2024}. However, the rapid growth of the wind industry demands automated and scalable solutions~\cite{woodmac9}. Blade image segmentation is a fundamental step in this process, simplifying complex image-based tasks such as defect detection~\cite{bunet,maskrcnn} and autonomous navigation~\cite{realtime,automateddrone}. By isolating the blades from complex backgrounds, segmentation can enhance the accuracy and efficiency of subsequent analysis~\cite{bunet}, facilitating faster, cost-effective maintenance and minimizing downtime. This innovation not only supports the operational efficiency of wind turbines but also contributes to the broader goal of sustainable energy production.

%Therefore, accurate classification of each pixel belonging to the wind turbine blade is crucial.

\begin{figure}[t!]
  \centering
    \includegraphics[width=\linewidth]{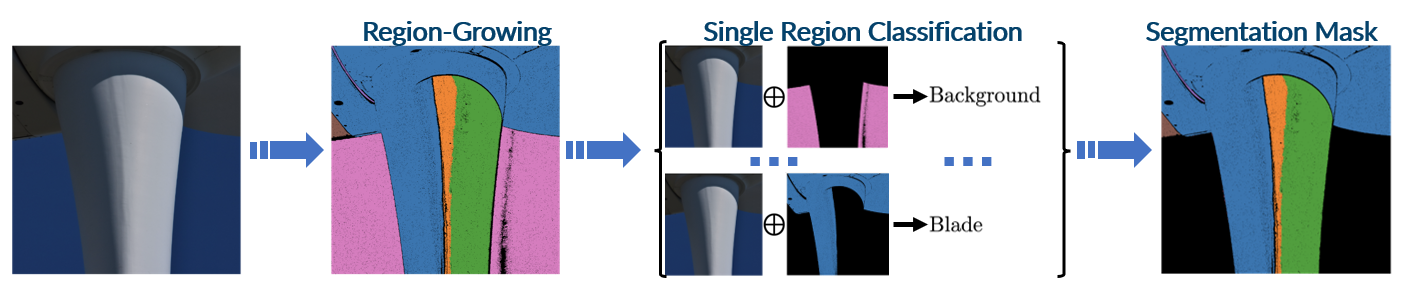} 
  \vspace{-0.8cm}
      \caption{\textbf{Schematic overview of our proposed framework}, refer to \cref{fig:marg-class} for details. The wind turbine blade is segmented by first identifying salient regions and then, classifying these regions into blade or background. Regions obtained by our MARG are depicted with different colors.}
  \label{fig:intro}
\vspace{-0.6cm}
\end{figure}

Current image segmentation solutions primarily use end-to-end learning algorithms and depend heavily on vast amounts of hand-annotated data~\cite{clipseg,mask2former,fourier-aug}. Just to give an idea on a popular model, SAM~\cite{sam} relies on one billion of hand-annotated masks. In contrast, the images available in this context is quite limited. Still, achieving state-of-the-art performance is imperative for automated wind turbine assessments. This work aims to create an innovative high-performance image segmentation framework despite the limited dataset. The approach simplifies the complex task of pixel-dense image segmentation into binary classification. To do that, we propose to create a bespoke region-growing algorithm to partition the images into salient regions, followed by training a deep learning model to classify these regions as either blade or background, ultimately combining these regions to identify the blade.

\newpage
In this work, the following methodologies are presented:

\begin{enumerate}

    \item A custom Dual-Threshold Modular Region-Growing (DTMRG) strategy for unsupervised salient region segmentation that generates quality segments. 
    \item An Adaptive Thresholding (AT) algorithm that seeks pairs of thresholds based on local image characteristics, fundamental to address a highly diversified dataset.
    \item A Region Merging (RM) strategy that reduces the amount of regions per image by almost 60\%, drastically enhancing region expressiveness and increasing the region's classifier accuracy by 5\%.
    \item A holistic Modular Adaptive Region-Growing (MARG) strategy that integrates DTMRG, AT, and RM, creating an effective region-growing algorithm.
    \item A novel data augmentation technique, termed RegionMix, designed to enhance region classification by shifting the model's learning approach: from binary classification to predicting the proportion of the blade region.
\end{enumerate}

All these components form a highly innovative image segmentation framework for limited-data scenarios. In wind turbine applications, this algorithm achieves state-of-the-art performance, surpassing existing methods across various metrics. Furthermore, it enhances the explainability of image segmentation, essential for industrial applications. This underscores the effectiveness of leveraging regions to streamline the segmentation problem into a binary classification task. Our approach is assessed across different windfarms, demonstrating consistently high and robust performance, along with remarkable generalization capabilities.

%% file: WACV26/chapters/RelatedWork.tex
\vspace{-0.2cm}
\section{Related Work}
\vspace{-0.1cm}

\textbf{Learning-based Image Segmentation}. Encoder-decoder architectures are among the most popular techniques for image segmentation, exemplified by DeepLab with atrous convolutions~\cite{deeplab,deeplabv3}, fast fully CNN networks with pyramid upsampling~\cite{fam_1_3}, and high-resolution bilateral reference models such as BiRefNet~\cite{birefnet}. These models offer direct pixel-dense categorization but often struggle in its generalization without extensive fine-tuning~\cite{fan,volpi}. Other approaches use feature maps for classification followed by dense pixel labeling, incorporating skip-layer connections~\cite{fam_2_1} or a CNN-based segmentation tree~\cite{fam_2_2}.

A bottom-up strategy forms superpixels—clusters of similar properties, such as bounding box identification~\cite{fam_3_1} or object recognition~\cite{fam_3_2}. For instance, \cite{fam_3_3} uses selective search to generate region proposals for object detection, then classifies them using a deep network. Other works introduce superpixel generation~\cite{xiang} for radar imagery classification or through neutrosophic clustering~\cite{hu}. 

Recently, transformers via attention mechanisms~\cite{mvanet} have emerged for their capabilities to capture long-range dependencies~\cite{resnest,vit_b_16,sw,mask2former}. Zero-shot models~\cite{clipseg,diffseg} like SAM~\cite{sam,sam2} demonstrate transformers' adaptability in diverse scenarios without domain-specific tuning. While requiring substantial computational resources and large datasets, techniques such as transfer learning~\cite{transfer} and lightweight models~\cite{chitty2023survey,mobilevit,efficientformer} reduce these demands. Other remarkable transformers combine channel-wise attention with a multi-path network layout~\cite{attentionunet} or with spatial hierarchies~\cite{agudoICIP20,unetformer,transU}, while frequency-aware feature fusion has also been introduced for dense prediction~\cite{freqfusion}.

\textbf{Region-growing Segmentation}. Region-growing segmentation is a flexible method that groups pixels into regions based on color or texture similarity~\cite{regiongrowing}. It adapts to gradual intensity changes and achieves accurate boundaries, forming coherent regions~\cite{huang2018weakly}. Several adaptations focus on automatic seed placement~\cite{regiongrowing_seed,regiongrowing_icip} and diverse region expansion criteria~\cite{regiongrowing1,regiongrowing2}. A first remarkable approach combines color-edge extraction with seeded region growing, using edge centroids as seeds~\cite{fan2001automatic}. Another introduces a three-rule seed selection in the YCbCr colorspace~\cite{frank}. An instance-based learning module applies distance criteria for region growth and ownership tables for merging~\cite{gomez}.

%This work presents a novel region-growing algorithm that incorporates an automatic Seed Selection mechanism [\cref{SS}], Dual Threshold criteria for pixel inclusion [\cref{DTRG}], a unique definition of connected neighbors [\cref{DTMRG}], and a post-processing phase defined by Region Merging [\cref{subch: RM}].

\textbf{Wind Turbine Blade Image Segmentation}. The plethora of related works published over the past five years evidence the importance of wind turbine blade segmentation withing the wind industry~\cite{thermal_blade_citeme,first_blade,hough_blade}. Many innovative learning-based approaches are based on the U-Net model~\cite{unet}, such as~\cite{thermal_blade} which includes a hierarchical split in the convolution block. Another example further augments its architecture with ECA- and PSA-attention~\cite{eca_attention_blade}. A combination of a ensembled CNN with Otsu thresholding for blade segmentation is introduced in~\cite{ensemble_blade}. Lastly, BU-Net~\cite{bunet} complemented the U-Net with ad-hoc random forest postprocessing to refine blade boundary precision.

\textbf{Data Augmentation}. Data augmentation enhances training data diversity through transformations that help reduce overfitting and improve generalization~\cite{mash,fourier-aug}. Successful approaches involve adjusting brightness and contrast~\cite{data-aug}, or applying whitening and standardization~\cite{sw}.

Advanced techniques linearly combine image pairs~\cite{mixup,guidedmixup,puzzlemix} or generate hybrid images~\cite{diffusemix}. Another approach randomly crops and rearranges image regions~\cite{qiao2020data}. These techniques are further extended to the object-level~\cite{zhang2021objectaug} and patch-level~\cite{cutmix}, aiming for domain adaptation~\cite{putmix,afan}.

%This literature review inspired the development of a novel data augmentation method named RegionMix (see section~\ref{fig:RegionMix_labeling}), which leverages segmented regions identified through the Region-Growing algorithm detailed in this work. By blending regions from different classes and creating a label that indicates the prevalence on the blade class, RegionMix has proved to enhance classification performance in section~\ref{sec:classification-efficacy} of a DCNN consequently affecting also the blade segmentation performance of the comprehensive framework proposed in this work.

%% file: WACV26/chapters/Methods.tex
\vspace{-0.15cm}
\section{Modular Adaptive Region-Growing}  \label{sec:marg}
\vspace{-0.15cm}

This section introduces a custom unsupervised modular adaptive seeded region-growing algorithm, designed for advanced image segmentation. The discussion encompasses a dual-threshold criterion for region growth (\cref{sec:dual-threshold}), a seed selection strategy (\cref{sec:seed-selection}), the introduction of modular neighbors (\cref{sec:modular}), and adaptive thresholding to dynamically adjust to image characteristics (\cref{sec:adaptive-thresholding}). We conclude with the region merging process (\cref{sec:reg-merging}), aimed at consolidating fragmented regions into more coherent segments.

\vspace{-0.15cm}
\subsection{Dual-Threshold Region-Growing} \label{sec:dual-threshold}
\vspace{-0.15cm}

Region-growing segmentation operates on the principle of pixel aggregation, where the grouping mechanism is based on homogeneity in a feature space that often includes intensity levels, textural patterns, or spatial closeness. The algorithm commences with the selection of {\it seed} points, which serve as the starting locations for region growth. As the iterative process unfolds, the algorithm examines adjacent pixels or subregions based on a similarity criterion and decides whether to merge them with the seed's growing region.%~\cite{RevisitingRG}.

Building upon the traditional region-growing algorithm~\cite{RevisitingRG}, we present the traditional dual-threshold variant, see Alg. 2 in Supplementary. Given an RGB image $\mathbf{X} \in \mathbb{R}^{H \times W \times 3}$, where $H$ and $W$ indicates their height and width, and $3$ the color channels, the goal is to segment $\mathbf{X}$ into homogeneous regions based on predefined set of rules. In this work, a dual-threshold region-growing that exploits both local threshold $\tau^{l}$ and seed threshold $\tau^{s}$ has been implemented. For a particular RGB pixel $\mathbf{x}_{h,w}$, the local $d_{l}$ and seed $d_{s}$ color distances are defined as follows:

\vspace{-0.7cm}
\begin{align}
     d_{l}(\mathbf{x}_{h,w},\mathbf{x}_{h',w'}) &= \frac{1}{3} \sum_{c=1}^3 \left| x_{h,w,c} - x_{h',w',c} \right|, \\
    d_{s}(\mathbf{x}_{h,w}) &= \frac{1}{3} \sum_{c=1}^3 \left| x_{h,w,c} - s_{\cdot,\cdot,c} \right|,
    \vspace{-0.5\baselineskip}
    \label{eq:seed_color_distance_def}
\end{align} %\vspace{-0.35cm}
where $(h,w)$ represents a pixel's coordinates, $(h',w')$ its neighbor, and $\mathbf{s}$ a seed pixel. Neighbors are defined using the notation $\aleph_{h,w}^{(k)}$, which represents the set of neighbors of a pixel at position $(h,w)$ for a given connectivity $k$:

\vspace{-0.5cm}
\begin{equation}
\Scale[0.95]{
\aleph_{h,w}^{(k)} = \left\{ \mathbf{x}_{h + i, w + j} \in \mathbf{X} \,|\, i, j \in \{-k, \ldots, k\}, (i, j) \neq (0, 0) \right\}.
}\nonumber
\end{equation}%\vspace{-0.55cm}

The algorithm applies $\tau^{l}$ and $\tau^{s}$ to ensure both local and global color consistency, respectively. For a candidate pixel $\mathbf{x}_{h',w'}$ to be added to a region, the conditions $d_{l} \leq \tau^{l}$ and $d_{s} \leq \tau^{s}, $ must be simultaneously met. The condition $d_{l} \leq \tau^{l}$ limits region expansion across object edges by ensuring that a pixel's color is not too dissimilar from its neighbors $\aleph_{h,w}^{(k)}$. Specifically, we operate over direct neighbors with $k=1$. Similarly, $d_{s} \leq \tau^{s}$ addresses global consistency by comparing candidate pixels to the seed one. This dual-threshold strategy prevents excessive region expansion, and facilitates accurate and coherent regions.

\vspace{-0.15cm}
\subsection{Seed-Selection Strategy} \label{sec:seed-selection} 
\vspace{-0.1cm}

\begin{figure}[t!]
    \centering
    \begin{tabular}{@{}c@{\hskip 1pt}c@{\hskip 1pt}c@{}}
        \includegraphics[width=.30\linewidth]{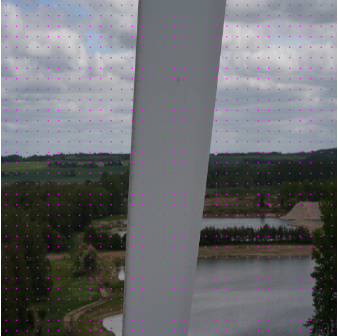} & 
        \hspace{0.02\linewidth} % Space between the images
        \includegraphics[width=.30\linewidth]{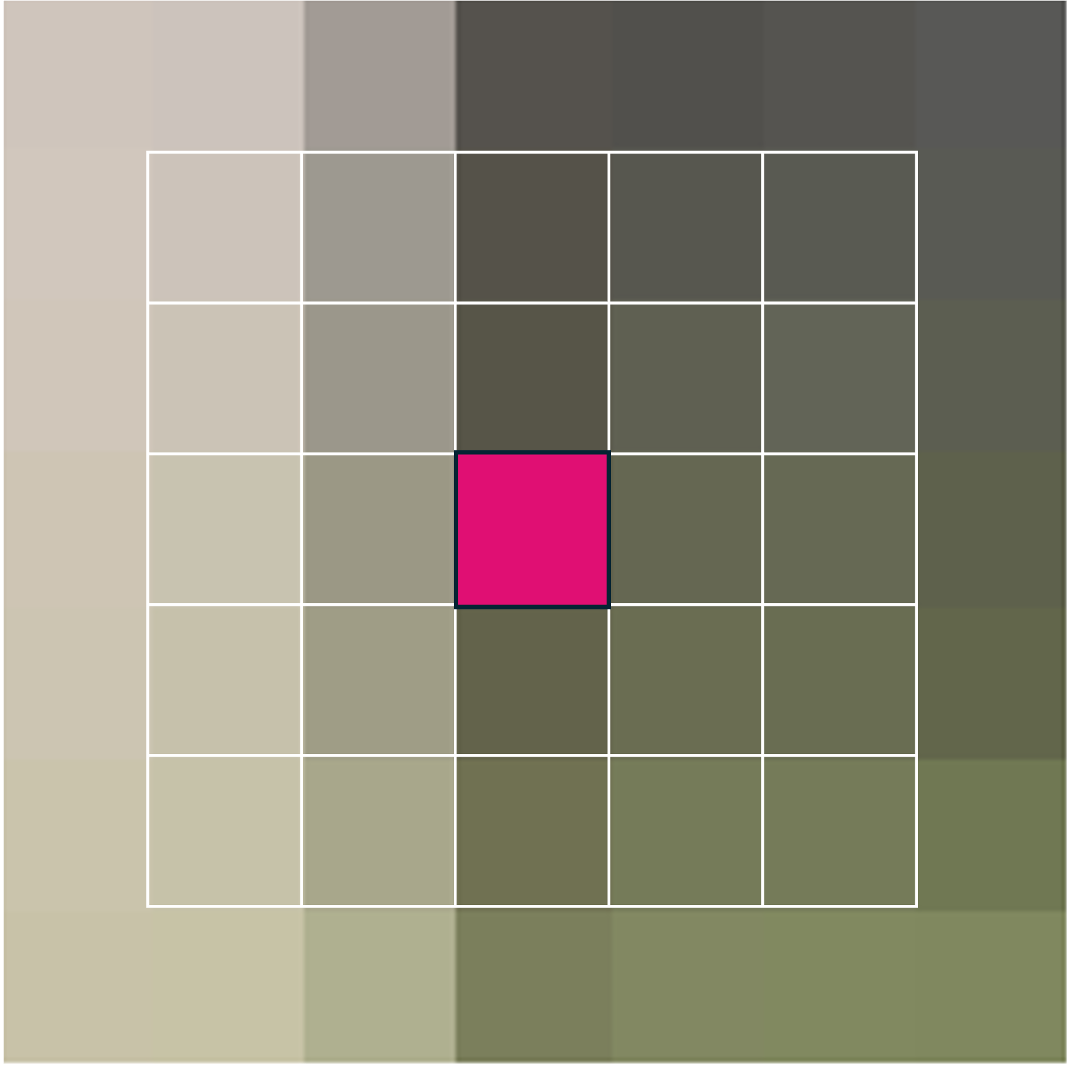} &
        \hspace{0.02\linewidth} % Space between the images
        \includegraphics[width=.30\linewidth]{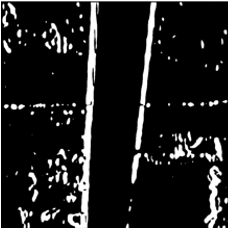} \\
    \end{tabular}
    \vspace{-0.4cm}
    \caption{\textbf{Seed selection process.} \textbf{Left:} equidistant seed candidate pixels in red. \textbf{Middle:} window $\aleph_{h,w}^{(2)}$ around a particular seed pixel $\mathbf{c}_{h,w}$. \textbf{Right:} Sobel's output.}
    \label{fig:seed-rg} 
    \vspace{-0.3cm}
\end{figure}

In the proposed work, seed pixels are selected dynamically as the segmentation process evolves, thereby reducing the risks of over- or under-segmentation through adaptation to the image features during region development. To identify suitable seed pixels, the approach involves selecting equidistant candidate seed pixels $\mathbf{c}_{h,w}$ from the image $\mathbf{X}$, refer to \cref{fig:seed-rg}.  For every candidate seed pixel, the window $\aleph_{h,w}^{(2)}$ is defined around it. This discretization facilitates the establishment of criteria for the selection of candidate seed pixels $\mathbf{c}_{h,w}$ as potential seeds --a process referred as Seed Promotion-- based on the following considerations:

%\vspace{-0.3cm}

\begin{enumerate}\label{SP}
    \item \textbf{Region Overlap Avoidance:} Define $\mathbf{X}_{\mathbf{c}}$ as the set of pixels already belonging to any previously grown region. A candidate seed pixel $\mathbf{c}_{h,w}$ is discarded if exists an overlap between its window $\aleph_{h,w}^{(2)}$ and $\mathbf{X}_{\mathbf{c}}$, i.e., the candidate is discarded if $\aleph_{h,w}^{(2)} \cap \mathbf{X}_{\mathbf{c}} \neq \phi$.

    \item \textbf{Edge-Guided Seed Displacement:} Let $\mathbf{S}$ be the Sobel operator output, which indicates the presence of edges within the image. If $\mathbf{c}_{h,w}$ is found within $\mathbf{S}$, it is moved in a random direction until $\mathbf{c}_{h,w} \notin \mathbf{S}$, ensuring seeds are not placed on high gradient color intensities.
    %\vspace{-0.35cm}
\end{enumerate}
   %\noindent

Upon satisfying these criteria, a candidate $ \mathbf{c}_{h,w} $ is promoted to a seed pixel, denoted as $ \mathbf{s}_{h,w} $, and utilized for region-growing. Sections \ref{sec:dual-threshold} and \ref{sec:seed-selection} coalesce to form Dual-Threshold Seeded Region-Growing (DTRG).

\vspace{-0.15cm}
\subsection{Modular Region-Growing} \label{sec:modular}
\vspace{-0.1cm}

\begin{figure}[t!]
\centering
\begin{tabular}{@{}c@{\hskip 2pt}c@{\hskip 2pt}c@{}}
  \includegraphics[width=0.25\linewidth]{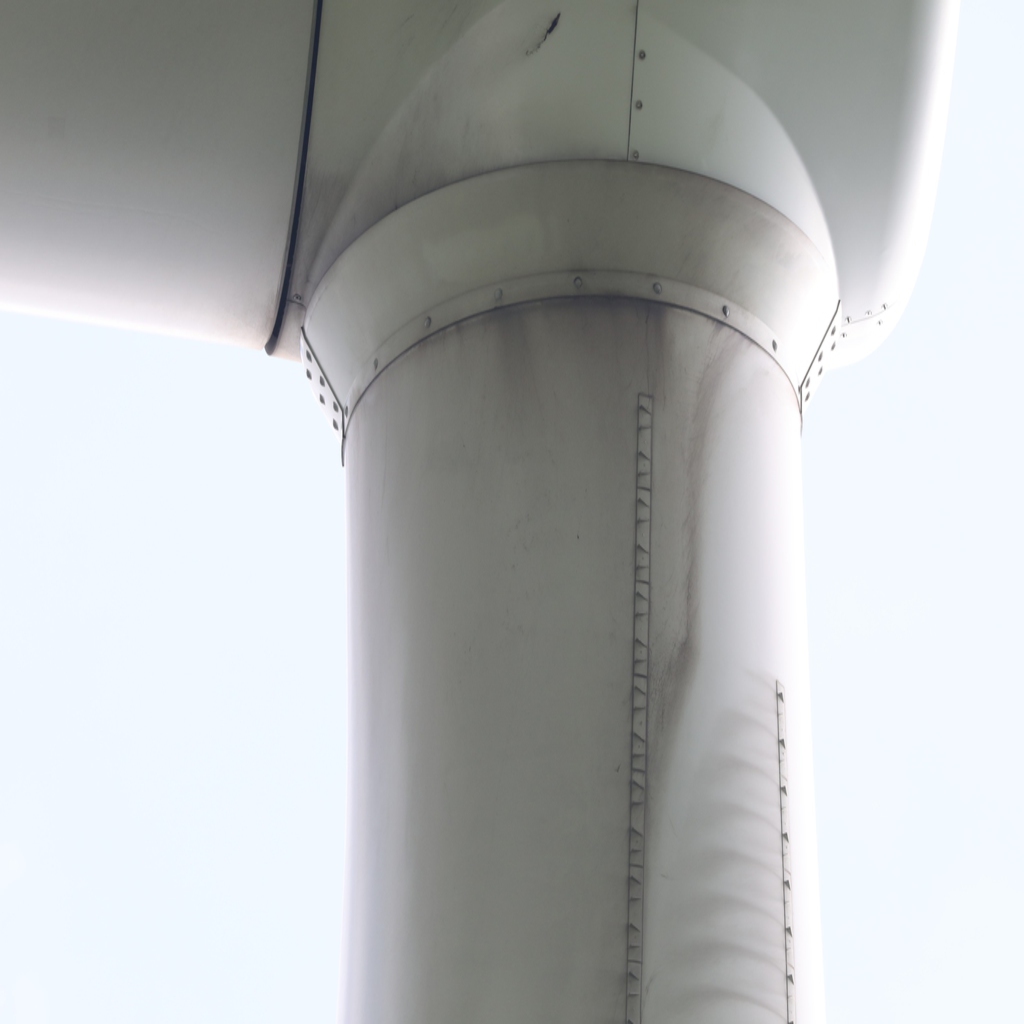} & 
  \includegraphics[width=0.25\linewidth]{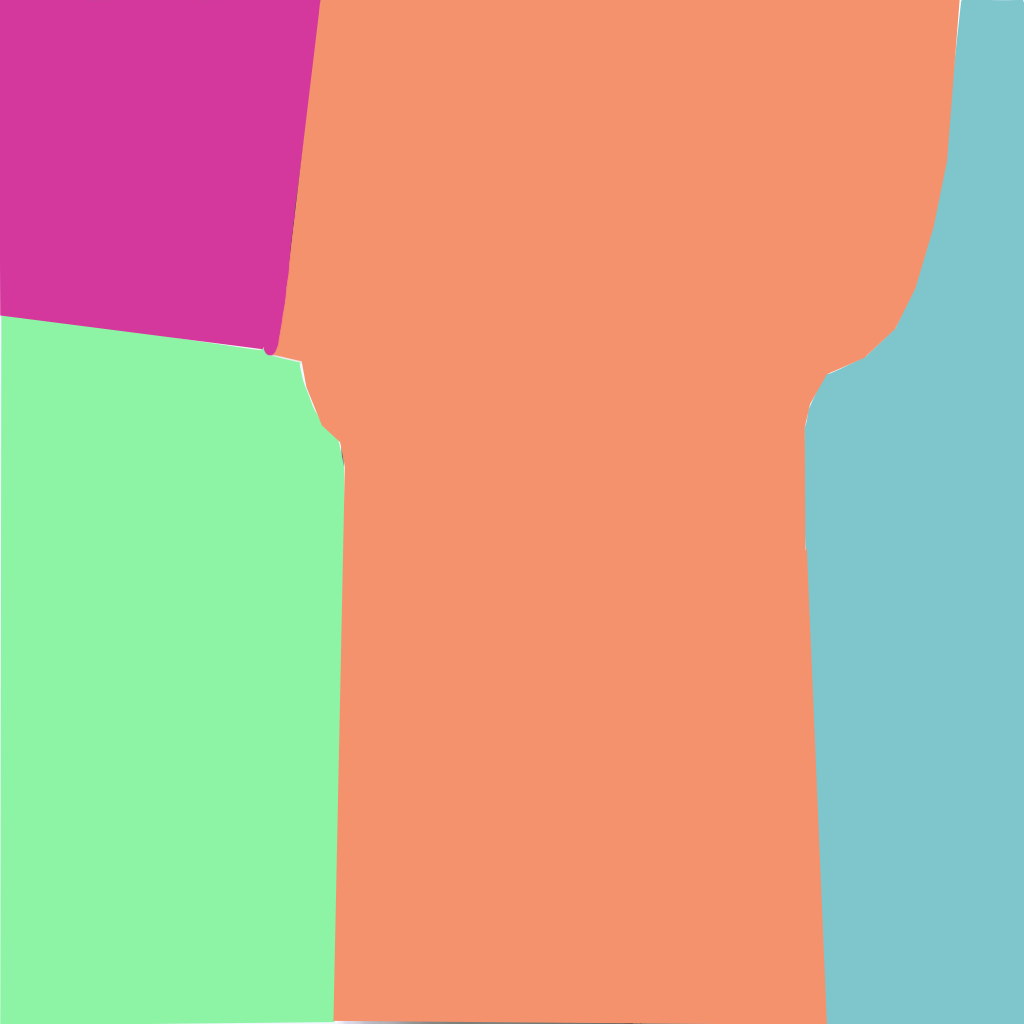} & 
  \includegraphics[width=0.25\linewidth]{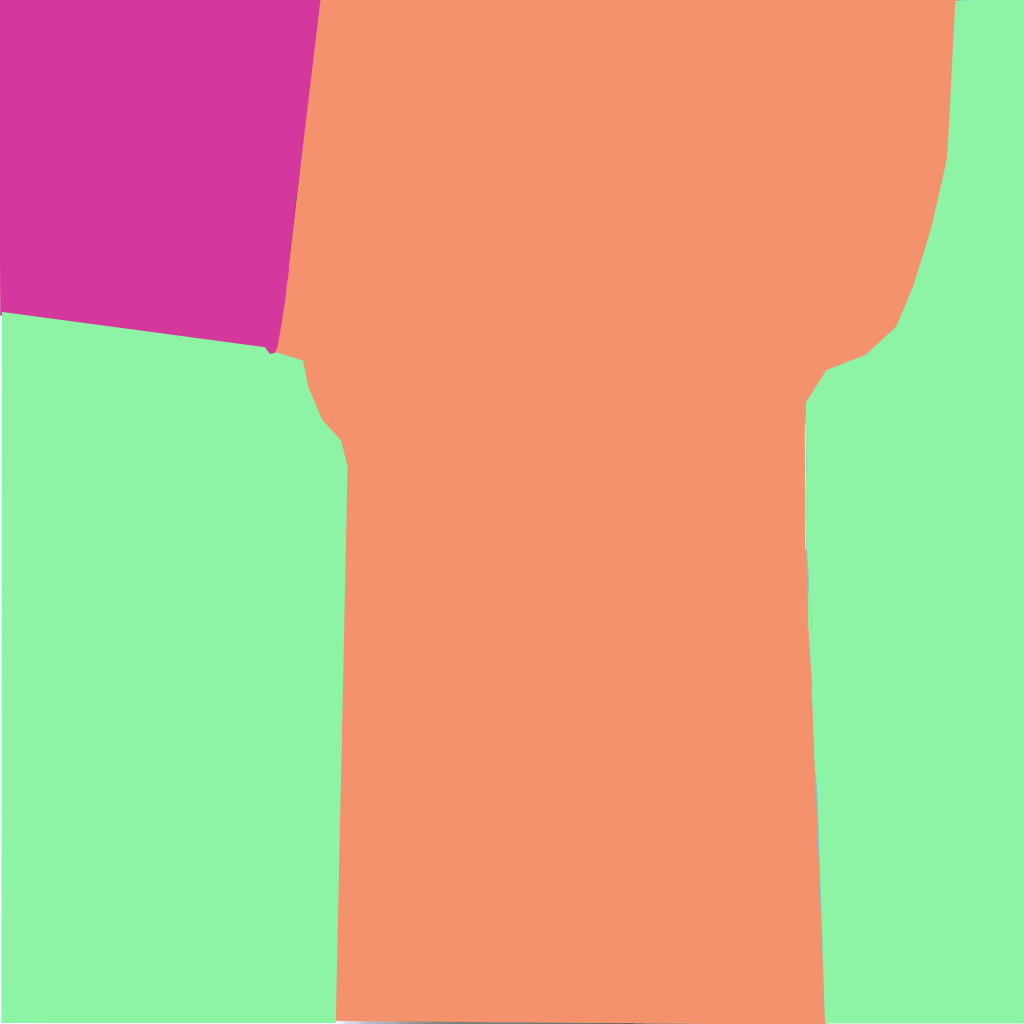} \\
  %(a) Original image & (b) Cartesian neighbors & (c) Modular neighbors \\[15pt]
  \includegraphics[width=0.25\linewidth]{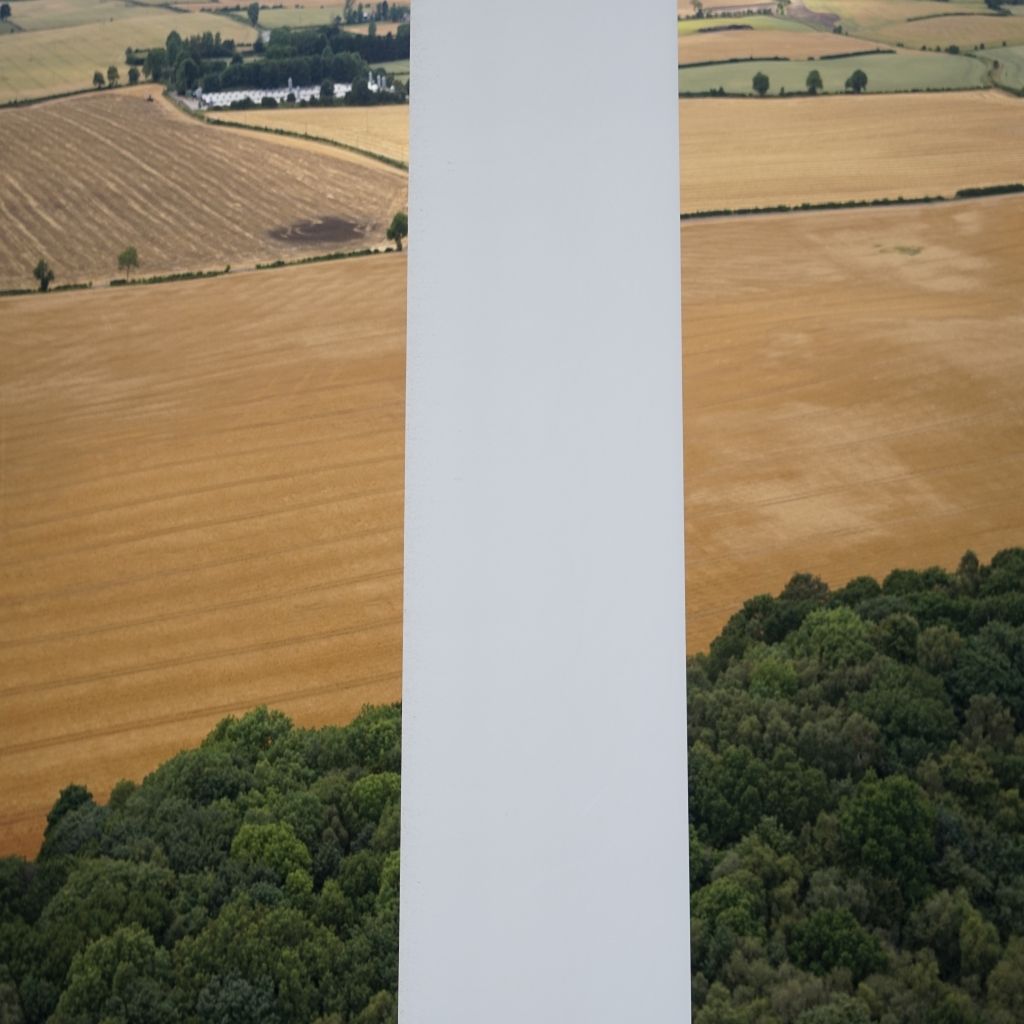} & 
  \includegraphics[width=0.25\linewidth]{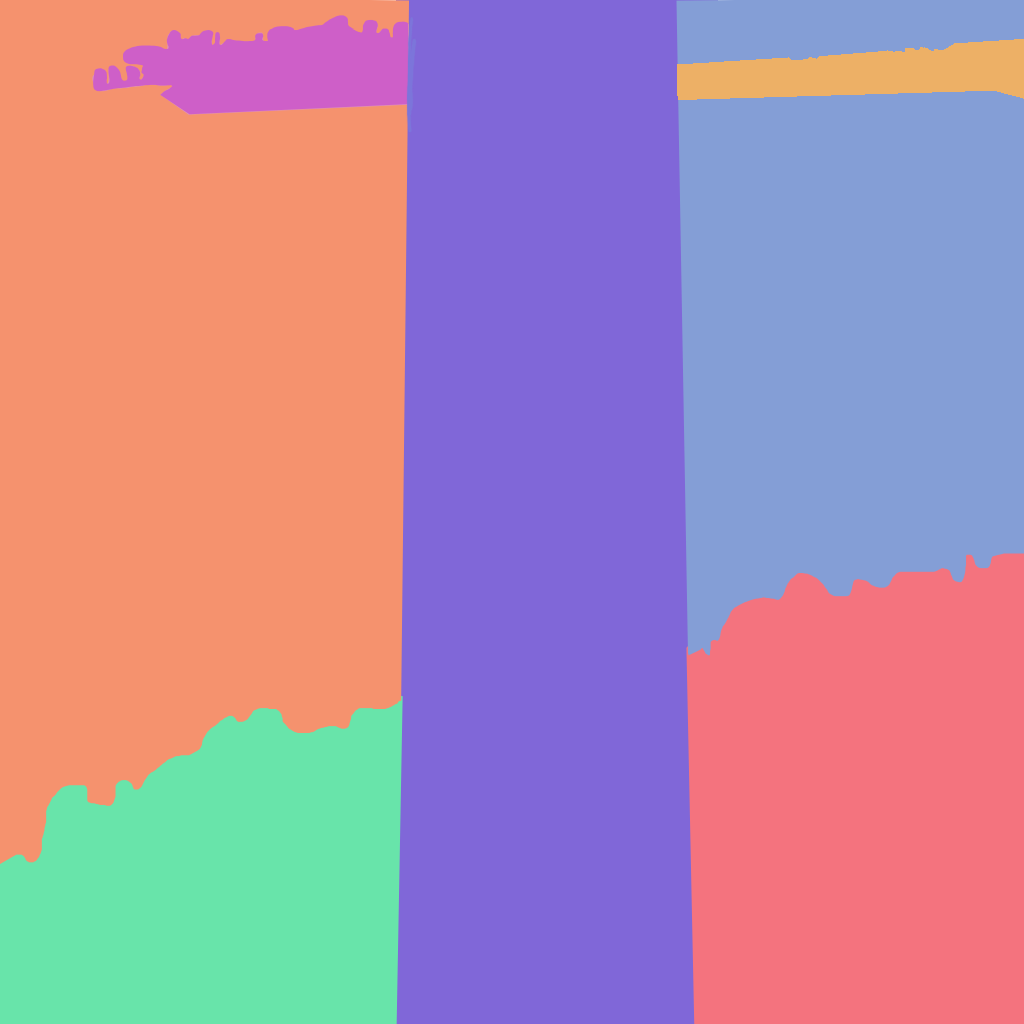} & 
  \includegraphics[width=0.25\linewidth]{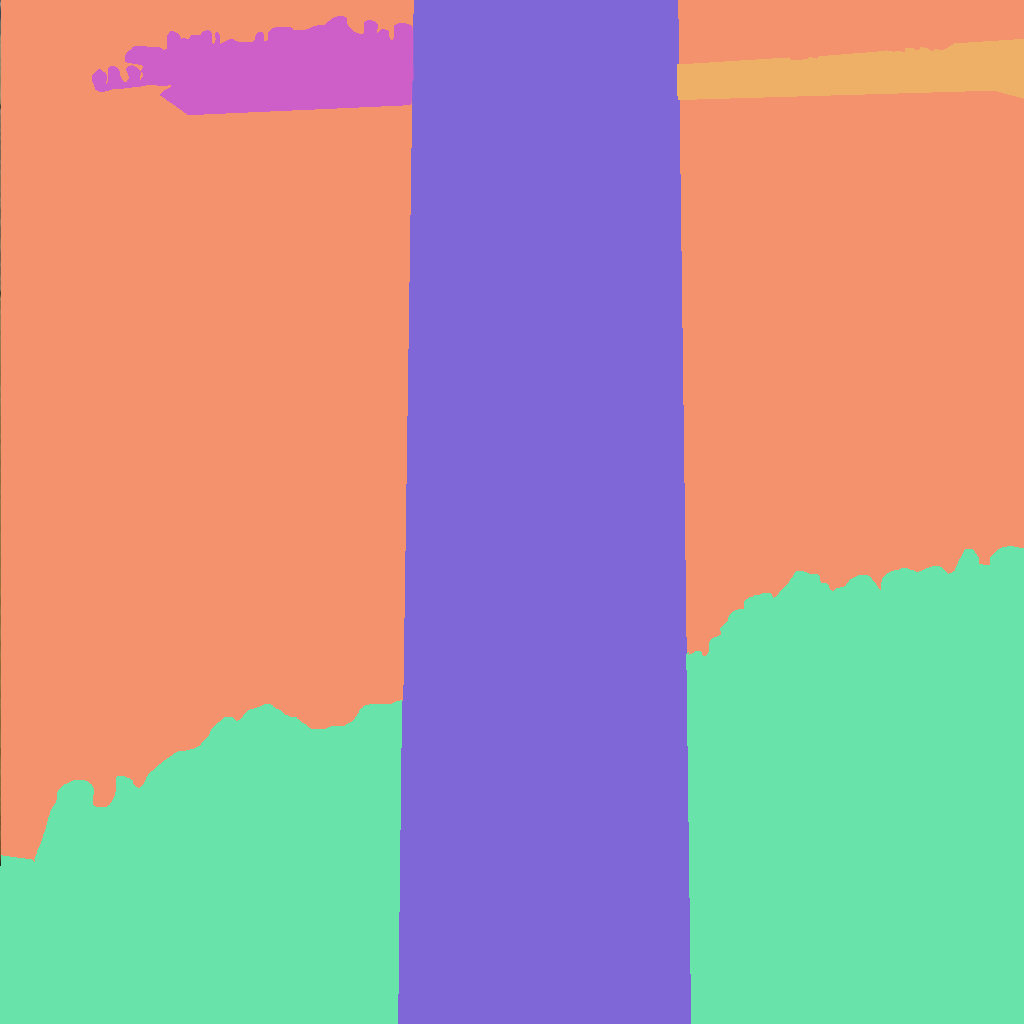} 
  %(d) Original image & (e) Cartesian neighbors  & (f) Modular neighbors \\
\end{tabular}
\vspace{-0.4cm}
\caption{\textbf{Cartesian Neighbors vs. Modular Neighbors.} Image $\mathbf{X}$, DTRG with cartesian neighbors, and DTRG with modular ones (DTMRG), respectively. Regions are shown with different colors.}
\label{fig:modular}
\vspace{-0.5cm}
\end{figure}

The Cartesian definition of neighbors, $\aleph_{h,w}^{(k)}$, is not ideal for toroidally structured images. Let $\bmod$ be the modular operator, an alternative neighbor set $\Theta_{h,w}^{(k)}$, which ensures continuity across image boundaries, can be defined as:

\vspace{-0.5cm}
\begin{equation}
\Scale[0.78]{
\Theta_{h,w}^{(k)} = \left\{ \mathbf{x}_{h + i \bmod H, w + j \bmod W} \in \mathbf{X} \,|\, i,j \in \{-k,\dots, k\}, (i, j) \neq (0, 0) \right\}}. \nonumber
\end{equation} %\vspace{-0.55cm}

This modification boosts the algorithm's effectiveness for images with toroidal boundaries, notably in wind turbine segmentation. Here, background elements like the sky, sea, or land can stretch over image edges (refer to \cref{fig:modular} for illustrative examples). The Dual-Threshold Modular Region-growing (DTMRG) works exactly as DTRG (\cref{sec:seed-selection}) but with the updated definition of neighbors $\Theta_{h,w}^{(1)}$.

%In the first row notice how the sky gets condensed into just one region using modular neighbors. In the second row, when employing modular neighbors, the fields and the treetops form only two cohesive regions as opposed to four when using Cartesian neighbors.

The first row of \cref{fig:modular} shows how the sky on either side of the blade is accurately segmented into a single cohesive region that spans the image borders. The second row similarly groups the fields and treetops on either side of the blade into one cohesive region. This nuanced alteration in neighbors' definition enables the identification of regions that more effectively capture the intrinsic structure of the image. A secondary advantage is the resultant reduction in the number of segmented regions per image, denoted as $N$.

\vspace{-0.2cm}
\subsection{Adaptive Thresholding} \label{sec:adaptive-thresholding}
\vspace{-0.15cm}

\begin{figure}[t!]
  \centering
  \begin{tabular}{@{}c@{\hskip 2pt}c@{\hskip 2pt}c@{}}
    \includegraphics[width=0.3\linewidth]{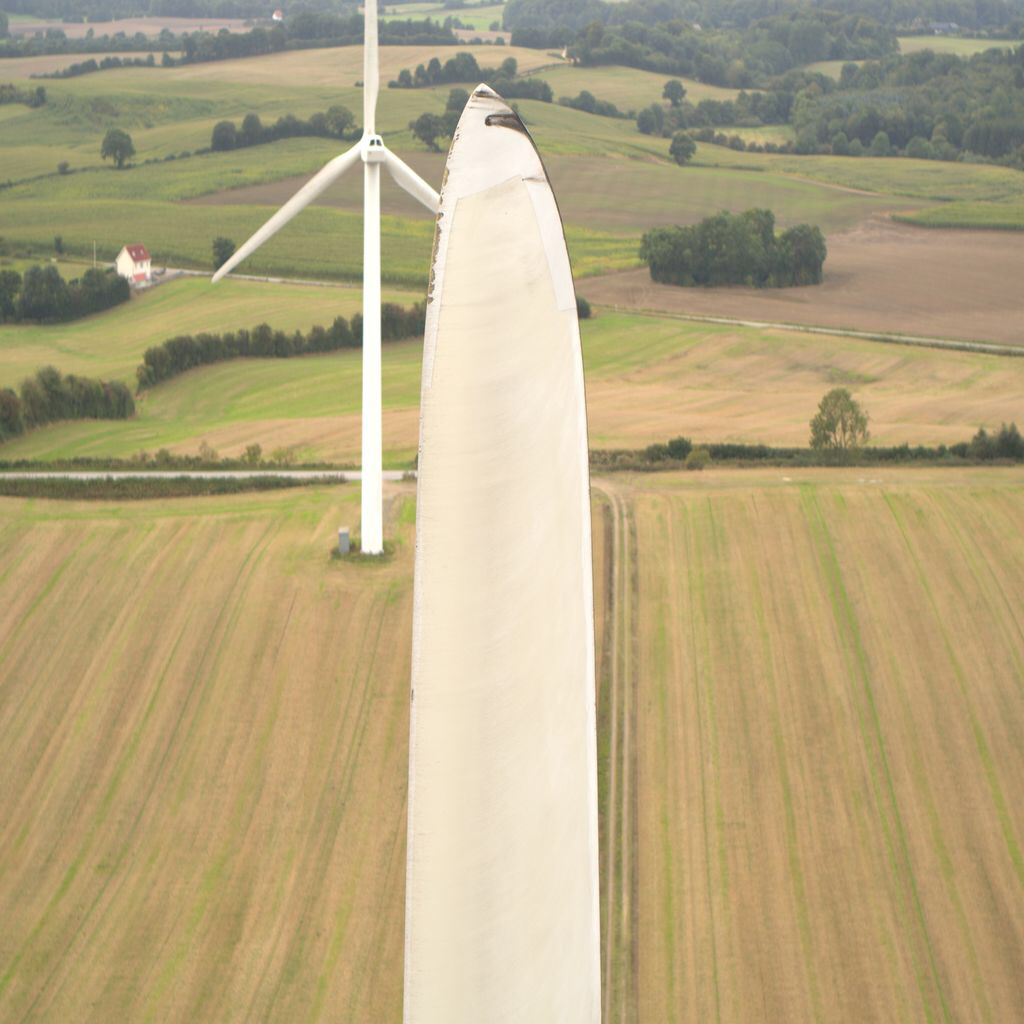} &
    \includegraphics[width=0.3\linewidth]{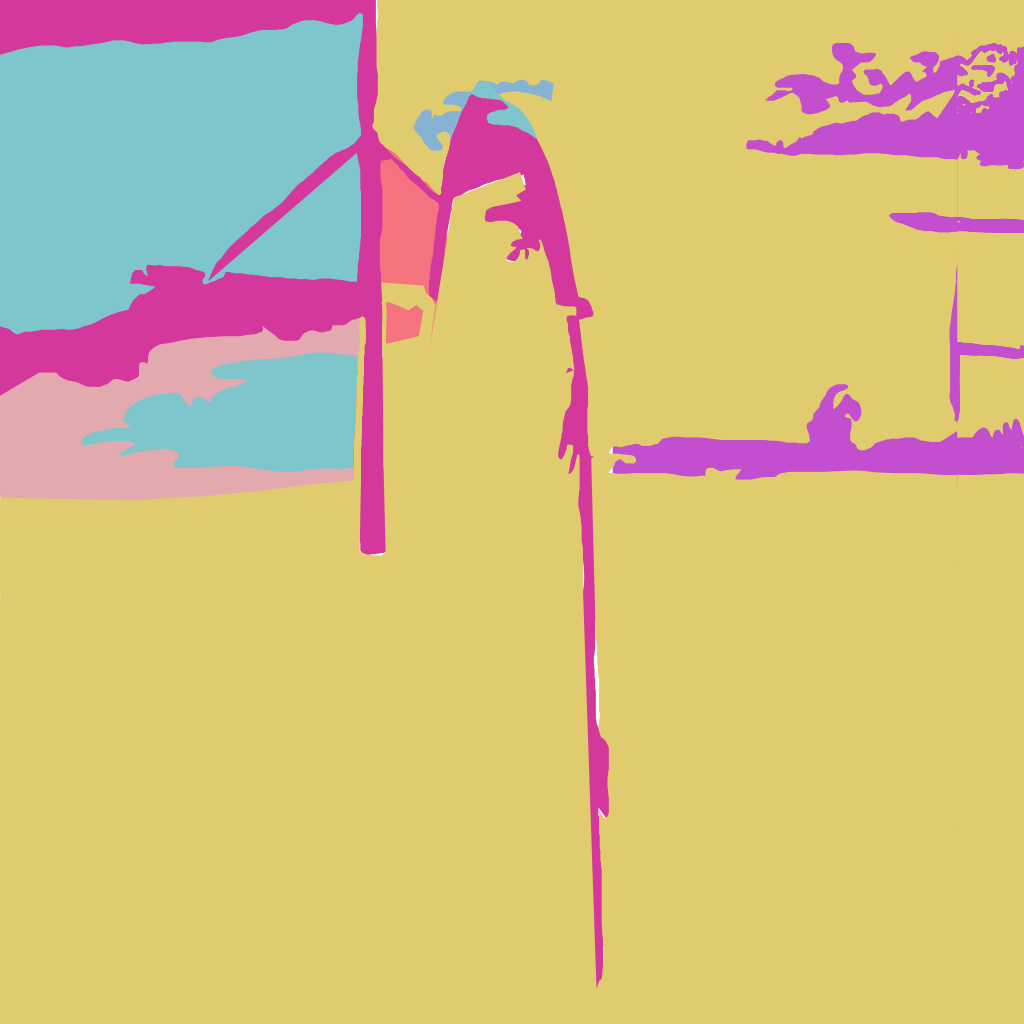} &
    \includegraphics[width=0.3\linewidth]{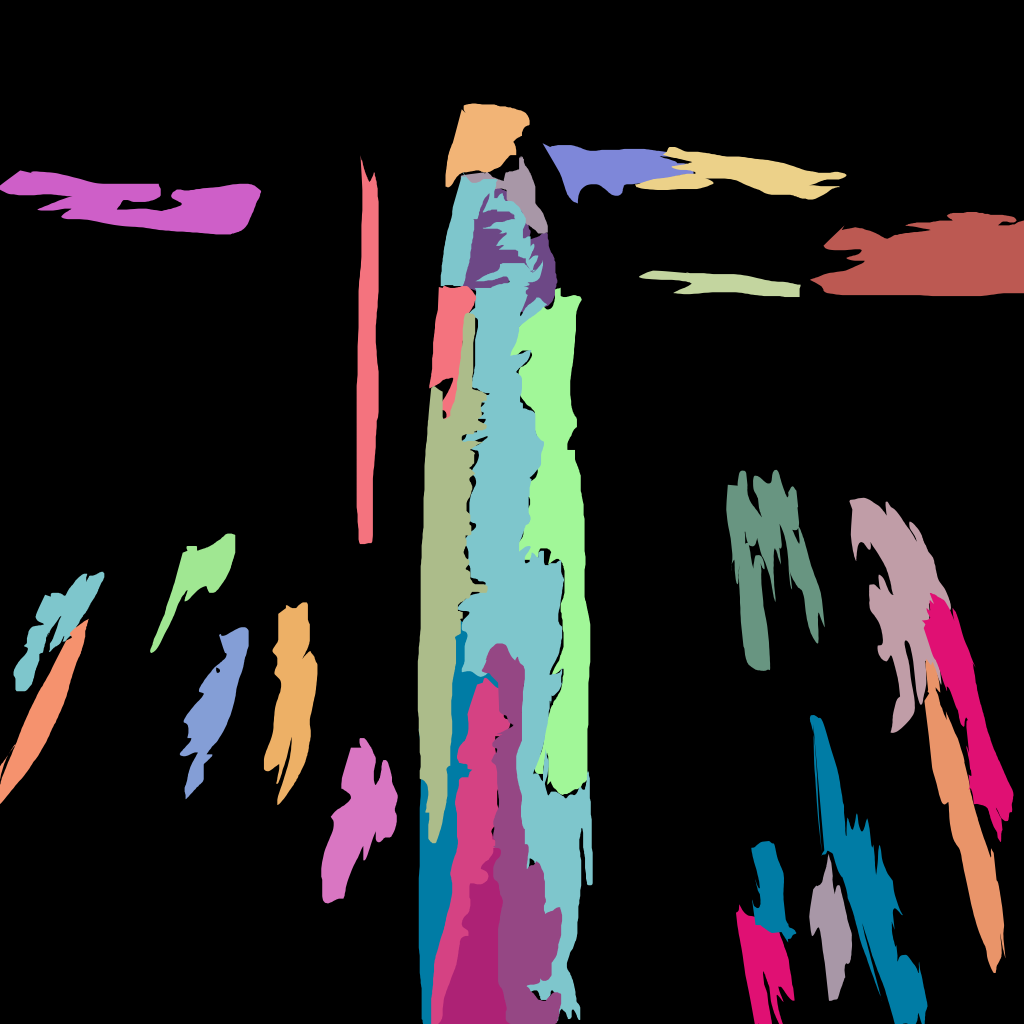} \\
    %(a) & (b) & (c) \\
  \end{tabular}
  \vspace{-0.35cm}
      \caption{\textbf{Distinct levels of coverage based on selected region-growing thresholds $\tau^{s}$ and $\tau^{l}$.} \textbf{Left:} Original image $\mathbf{X}$; \textbf{Middle:} Very high $\tau^{s}$ and $\tau^{l}$, $\Pi = 100\%$ but it generates over-grown regions; \textbf{Right:} Very stringent $\tau^{s}$ and $\tau^{l}$, leading to low coverage $\Pi$; under-segmentation. Regions are depicted with different colors.}
  \label{fig:adaptative-rg-examples}
   \vspace{-0.35cm}
\end{figure}

The principle of Adaptive Thresholding (AT) is to dynamically adjust the thresholds $\tau^{l}$ and $\tau^{s}$ based on the local image characteristics, enhancing the algorithm’s ability to accurately delineate regions of interest in a specific image $\mathbf{X}$. In complex natural scenes, Global Thresholding (GT) is inadequate due to image variability. \Cref{fig:adaptative-rg-examples} illustrates the importance of proper threshold selection.

\begin{figure}[t!]%[h!]
    \centering
    
        \begin{subfigure}[t]{0.32\linewidth}
            \centering
            \includegraphics[width=\linewidth]{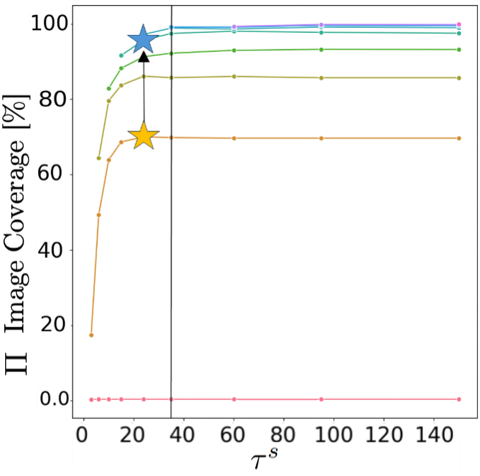} 
            \vspace{-0.5cm}
            \caption{$\Pi$ vs. $\tau^{s}$}
            \label{subfig:coverage_adaptive}
        \end{subfigure} 
        \begin{subfigure}[t]{0.32\linewidth}
            \centering
            \includegraphics[width=\linewidth]{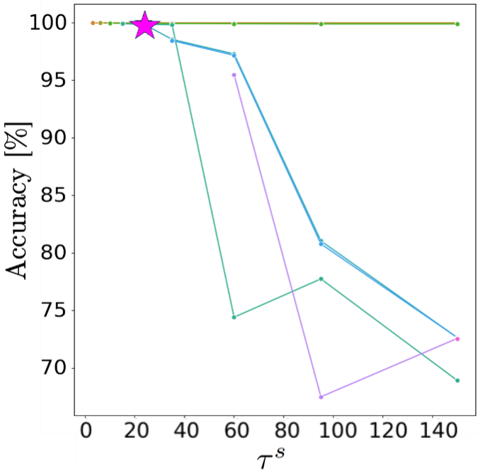}
            \vspace{-0.5cm}
            \caption{Accuracy vs. $\tau^{s}$}
            \label{subfig:accuracy_adaptive}
        \end{subfigure} 
        \begin{subfigure}[t]{0.31\linewidth}
            \centering
            \includegraphics[width=\linewidth]{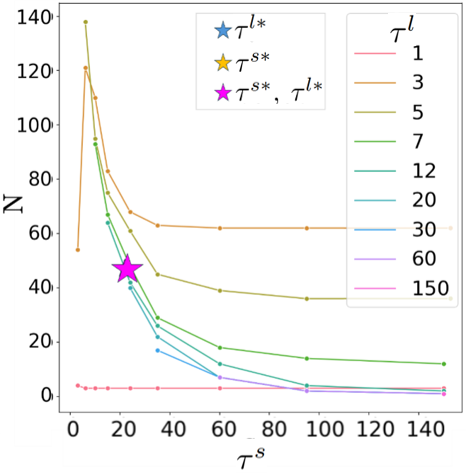}
            \vspace{-0.5cm}
            \caption{$N$ vs. $\tau^{s}$}
            \label{subfig:nregions_adaptive}
        \end{subfigure}
         \vspace{-0.4cm}

    \caption{\textbf{Comparative analysis of segmentation performance metrics across varying $\tau^{s}$ values, differentiated by $\tau^{l}$ levels.} The golden star indicates $\tau^{s*}$, the blue one $\tau^{l*}$, while the purple star indicates the couple of thresholds $\tau^{s*}$,$\tau^{l*}$.}
    \label{fig:adaptative-rg} %The left plot showcases image coverage against $\tau^{s}$, the middle plot displays accuracy in relation to $\tau^{s}$, and the right plot illustrates the number of regions $N$ versus $\tau^{s}$.
\vspace{-0.55cm}
\end{figure}

The implementation of AT allows to strike a balance between \textit{image coverage} $\Pi$, defined as the ratio of pixels of $\mathbf{X}$ that belong to a region $\Pi = \frac{|\mathbf{X}_{\mathbf{c}}|}{H \times W}$, and \textit{region accuracy}, defined as the proportion of pixels that are grown in a unique class. $\Pi$ in this context, can be seen as a control variable to monitor region growth leveraging the underlying trade-off that exists between the threshold choice and region accuracy: higher $\tau^{s}$ and $\tau^{l}$ correspond to a higher $\Pi$ at the cost of region accuracy. The adaptive strategy revolves around the idea of selecting a couple of $\tau^{s*}$ and $\tau^{l*}$ prior to any decrease in the accuracy metric by leveraging the relationship between accuracy and coverage. The $\tau^{s}$ acts as a growth-inhibiting radius around the seed pixel, limiting region growth and ensuring segments remain color consistent with the seed, adding a ``memory factor" to the growth; $\tau^{l}$'s key role instead is identifying object boundaries. 

\Cref{fig:adaptative-rg} illustrates that increments in $\tau^{s}$ beyond a certain threshold yield no further improvements in $\Pi$. Reaching this plateau, $\frac{d\Pi}{d\tau^{s}} \approx 0$, allows to select the value of $\tau^{s*}$ before any decrement in accuracy. To select the optimal $\tau^{l*}$, it is sufficient to increase $\tau^{l}$ until a satisfactory image coverage $\Pi$ is accomplished. \Cref{fig:adaptative-rg} illustrates the region accuracy with respect to $\tau^{s}$ and $\tau^{l}$. The figure shows the selection of the pair $\tau^{s*}$, $\tau^{l*}$ prior to any decrease in accuracy. Finally, the number of regions $N$ with respect to $\tau^{s}$, $\tau^{l}$ plot confirms the negative correlation between $N$ and $\tau^{s}$, validating the principle that $\tau^{s}$ constrains the extent of region growth. A qualitative example of AT algorithm is included in the Supplementary in Fig. 2, with the complete algorithm detailed in Alg. 3 of Supplementary.

\vspace{-0.15cm}
\subsection{Region Merging} \label{sec:reg-merging}
\vspace{-0.1cm}

% \begin{figure}[t!]
%   \centering
%  \includegraphics[width=\linewidth]{Plots/full_algo.png}
%   \caption{\textbf{Pipeline of the Modular Adaptive Seeded Region-Growing Algorithm (MARG).} \textbf{Top-left:} Input image. \textbf{Top-right:} Sequence of images showing the Adaptive Thresholding process with varying thresholds. \textbf{Bottom-right:} Segmented regions of DTMRG using ${s_t}^{\star}$ and ${l_t}^{\star}$. \textbf{Bottom-center:} The Region-Merging matrix indicating potential merges. \textbf{Bottom-left:} Final regions indicated by different colors.}
% \label{fig:full_algo}
% \end{figure}

Simplifying subsequent classification tasks by reducing the overall number of regions is crucial to enhance segmentation. Region Merging (RM) strategy initializes a symmetric binary matrix, denoted as $\mathbf{R}^M \in \{0,1\}^{N \times N}$, where each element $\mathbf{R}^{M}_{i,j}$ indicates the mergeability of region pairs $\mathbf{R}_i , \mathbf{R}_j \in \{0,1\}^{H \times W}$, determined by their overlap. This mergeability is formally defined by the expression:
\begin{equation}\label{eq:marg-rm}
    \mathbf{R}^{M}_{i,j} = 
    \begin{cases}
    1, & \text{if } \frac{|\mathbf{R}_i \cap \mathbf{R}_j|}{\min(|\mathbf{R}_i|, |\mathbf{R}_j|)} \geq 0.1 \\
    0, & \text{otherwise}
    \end{cases},
\end{equation}
where $|\mathbf{R}_i \cap \mathbf{R}_j|$ is the intersection area between the two regions, and $|\mathbf{R}_i|$ and $|\mathbf{R}_j|$ are their individual areas; refer to Fig. 3 in Supplementary for an illustrative example.

With $\mathbf{R}^{M}$ defined, a global merging strategy is followed for the identification of chains of interconnected regions. Through a graph traversal technique, specifically Depth-First Search, regions are categorized into chains based on their connections in $\mathbf{R}^{M}$, setting a comprehensive structure of mergeable regions for the actual merging process.

Finally, regions within each chain are iteratively merged using a union operation, updating the segmented image with larger, unified regions. Modular Adaptive Seeded Region-Growing algorithm with Region Merging (MARG) is outlined in \cref{fig:marg-class}, and detailed in Alg. 4 of Supplementary.

\vspace{-0.15cm}
\section{Region-aware Classifier} \label{ch:Classification}%
\vspace{-0.2cm}

%This section delves into the process of neural classification, beginning with a foundational binary classification model for blade or background region categorization, progressing through the novel RegionMix data augmentation technique for improved performance, and discussing the model implementation.

This section explores the binary classification model for categorizing the regions. It then introduces the novel RegionMix data augmentation technique to enhance performance and concludes with the model implementation.

\vspace{-0.2cm}
\subsection{Binary Region Labeling}
\vspace{-0.15cm}

A binary label to each region is assigned indicating whether it belongs to the blade (1) or background (0). Since MARG algorithm operates in an unsupervised manner, some regions may not be exclusively grown over a single class. Such ambiguous cases are excluded from the training.

The challenging problem of dense pixel classification, which leads to the final segmentation result, is thus reduced to the simplest form of a machine-learning task, which is particularly advantageous when the available data is scarce. In the context of wind turbines, which are challenging to access and often situated in remote and isolated locations, binary region classification is especially beneficial. 

\vspace{-0.2cm}
\subsection{RegionMix Data Augmentation}
\vspace{-0.15cm}

RegionMix is designed to augment the shape, size, and contextual placement of regions. RegionMix capitalizes on the segmented outputs of the MARG algorithm to expose the model to a virtually unlimited combination of region assemblies, significantly enriching the training variability. 

\begin{figure}[t!]
\captionsetup{font=small, skip=0\baselineskip}
    \centering
    \includegraphics[width=\linewidth]{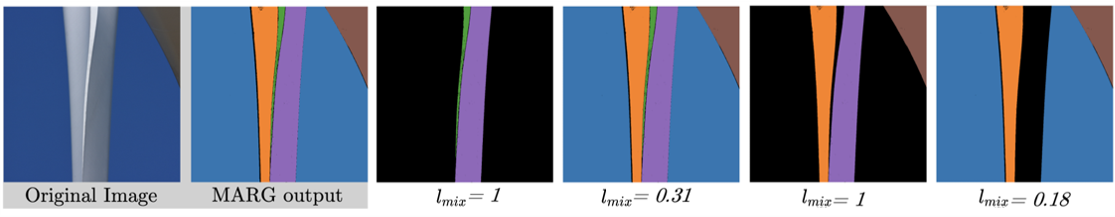}
    \caption{\textbf{Examples of RegionMix data augmentation}.}
    \label{fig:reg-mix}
    \vspace{-0.5cm}
\end{figure}

RegionMix synthesizes a new training example, $ \mathbf{R}_{\text{mix}} $, by blending existing regions. Consider the set of regions $ \{\mathbf{R}_1, \mathbf{R}_2, \ldots, \mathbf{R}_N\} $ for an image $ \mathbf{X} $, where $ N $ represents the total number of regions. First, a subset $ \mathcal{R} $ of randomly selected regions is generated. Then a composite region $ \mathbf{R}_{\text{mix}} $ is generated by the union of the regions in this subset $ \mathcal{R} $: $ \mathbf{R}_{\text{mix}} = \bigcup_{\mathbf{R}_i \in \mathcal{R}} \mathbf{R}_i $. The label $ l_{\text{mix}} $ assigned to this artificial region is the proportion of the mixed region that overlaps with the segmentation mask $\mathbf{M}$ (refer to \cref{fig:reg-mix}):

\vspace{-0.3cm}
\begin{equation}
   l_{\text{mix}}(\mathbf{R}_{\text{mix}}) = \frac{|\mathbf{R}_{\text{mix}} \bigcap \mathbf{M}|}{|\mathbf{R}_{\text{mix}}|}.
\end{equation}

%RegionMix is a novel technique that uses segmented region data with continuous labels rather than raw pixels. This approach increases the model's sensitivity to different class characteristics and boosts prediction confidence by effectively managing real-world variations. Training on diverse region combinations helps prevent overfitting and improves the generalization for complex natural images.

\vspace{-0.4cm}
\subsubsection{Implementation Details}\label{sec:implementation-reg}
\vspace{-0.1cm}

%The region classifier is implemented using a modified ResNet18 architecture~\cite{resnet} designed specifically for this task. We add a fourth channel input embedding the binary mask of the region being classified. \cref{fig:marg-class} shows a schematic overview of the whole algorithm. Key specifications include the binary cross entropy loss, a batch size of 4, and the AdamW optimizer~\cite{adamw} with a learning rate of $0.0005$ and a weight decay of $1 \times 10^{-5}$. Additionally, the Lookahead optimization strategy~\cite{lookahead} is employed, which updates ``slow weights" towards ``fast weights" every 5 steps with a step size of 0.8, providing a faster convergence. Hole-Filling~\cite{bunet} post-processing is applied to the union of all the blade regions.

The dataset used follows BU-Net~\cite{bunet} and is detailed in Section 12 of Supplementary. The region classifier is implemented using a modified EfficientNet-B4 architecture~\cite{efficientnet} that incorporates an additional fourth channel input to embed the binary mask of the region being classified. Other architectures were explored and experiments are reported in Section~10 of Supplementary. \Cref{fig:marg-class} shows a schematic overview of the whole algorithm. Key specifications include the binary cross entropy loss, a batch size of 4, and the AdamW optimizer~\cite{adamw} with a learning rate of $0.0005$ and a weight decay of $10^{-5}$. The Lookahead optimization strategy~\cite{lookahead} is employed, which updates ``slow weights" towards ``fast weights" every 5 steps with a step size of 0.8.%, providing a faster convergence. %Hole-Filling~\cite{bunet} is applied after assembling the blade regions.

\begin{figure*}[t!] %figure*
\captionsetup{font=small, skip=0.3\baselineskip}
    \centering
    \includegraphics[width=\linewidth]{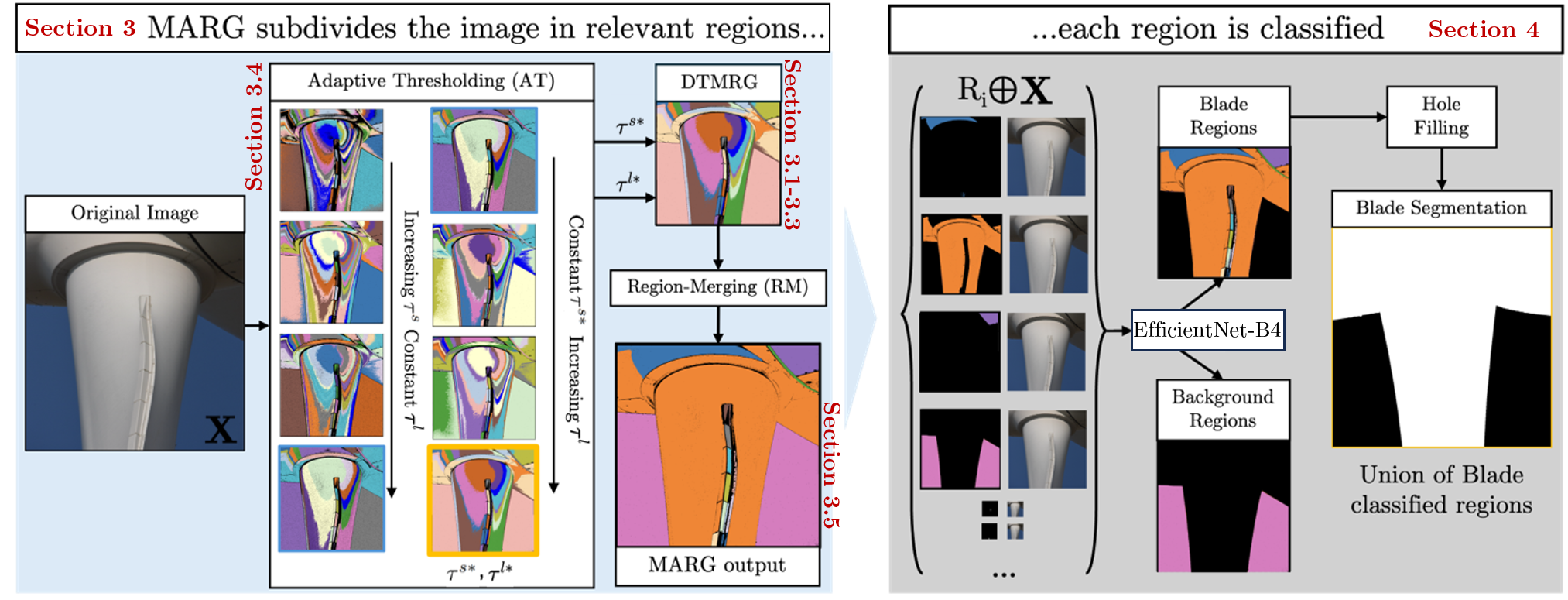}
    \vspace{-0.5cm}
    \caption{\textbf{Left, MARG:} AT selects ${\tau^{s*}}$, ${\tau^{l*}}$ for $\mathbf{X}$, DTMRG grows the regions, RM consolidates them. \textbf{Right, Region classifier:} each proposed region is classified as blade or background. Classified regions as blade set the generated segmentation mask.}
    \label{fig:marg-class}
    \vspace{-0.5cm}
\end{figure*} 

%% file: WACV26/chapters/Results.tex
%The Reference Seeded Region-Growing algorithm (RSRG), relies on a Randomized Seed Selection and Global Threhsolding (GT), which means that for every image, the same couple of thresholds \( l_t, s_t \) is applied. 

\vspace{-0.15cm}
\section{Region-Growing Evaluation} \label{sec:reg-results}
\vspace{-0.2cm}

This section delves into the analysis of the region-growing segmentation applied to wind turbine blade imagery. First, a theoretical formulation is utilized to assess region-growing segmentation performance, and then an ablation study highlights the algorithm's adaptability and performance under the various enhancements. Lastly, a qualitative analysis highlights the benefits of each algorithm module.

\vspace{-0.15cm}
\subsection{Region-Growing Segmentation Performance} \label{RG-perfect-classifier}
\vspace{-0.15cm}

An accurate region-growing algorithm should generate regions whose pixels only belong to one class. Hence, given an image $\mathbf{X}$, its associated set of regions $\{\mathbf{R}_1, \mathbf{R}_2, \ldots, \mathbf{R}_N\}$ and the ground-truth binary mask $\mathbf{M}$, we evaluate the segmentation quality of a region $\mathbf{R}_i$ for a given similarity metric $\texttt{sim}$ as:

\vspace{-0.4cm}
\begin{equation}
\Scale[0.96]{
\texttt{sim}_{\mathbf{R}_i} = \max \left[ \texttt{sim}\Big(\mathbf{R}_i^{C_1}, \mathbf{M}[\mathbf{R}_i]\Big), \texttt{sim}\Big(\mathbf{R}_i^{C_0}, \mathbf{M}[\mathbf{R}_i]\Big) \right].}
\end{equation}

The generated region $\mathbf{R}_i$ is considered either as blade $\mathbf{R}_i^{C_1}$ or as background $\mathbf{R}_i^{C_0}$ and the highest metric score is captured for each region when comparing with the region from the ground-truth mask $\mathbf{M}[\mathbf{R}_i]$. So, given the region size $|\mathbf{R}_i|$, the overall performance metric $\texttt{sim}_{\mathbf{X}}$ for the image $\mathbf{X}$ is the weighted average across all segmented regions:

\vspace{-0.4cm}
\begin{equation} \label{eq:marg-sim}
\texttt{sim}_{\mathbf{X}} = \frac{1}{|\mathbf{X}|} \sum_{i=1}^{N} |\mathbf{R}_i|  \texttt{sim}_{\mathbf{R}_i}.
\end{equation}

This method ensures that the contributions of all regions are proportionately reflected and that we can assess the region-growing algorithm using standard metrics.% such as Accuracy, Recall, F1-Score, and Intersection over Union (IoU) for each image.

\vspace{-0.1cm}
\subsection{Region-Growing Ablation Study} \label{Ablation_RG}
\vspace{-0.14cm}

An ablation study is performed to assess each region-growing module, with the baseline being the Reference Seeded Region-Growing (RSRG), which employs Global Thresholding and a Random Seed Selection. Apart from the segmentation metrics, we study the image coverage ($\Pi$) (defined in \cref{sec:adaptive-thresholding}) and the number of segmented regions ($N$). The findings are summarized in \cref{tab:reg-ablation}.%, showcasing the improvements beyond RSRG.

The baseline RSRG highlights notable limitations in recall and $\Pi$, with over $11\%$ of $\mathbf{X}$ remaining unassigned to any grown region. This shortfall is attributed to the seed pixels' failure to segment salient regions and the Global Thresholding parameters $\tau^{l}$ and $\tau^{s}$ not being optimized for specific image conditions. The introduction of Seed Selection (\cref{sec:seed-selection}) increases $\Pi$ by 4.56\% and significantly improves all performance metrics, highlighting the importance of accurate seed selection to avoid missing image areas.

Adaptive Thresholding (\cref{sec:adaptive-thresholding}) significantly enhances recall to $94.41\%$ by optimizing $\tau^{s*}$ and $\tau^{l*}$ to handle varying contrast, clutter, and noise. However, it prioritizes segmentation accuracy over minimizing $N$ or maximizing $\Pi$, leading to fragmented segmentation with $N=23.6$ regions per image, increasing misclassification risk (\cref{sec:classification-efficacy}). Fig. 4 in Supplementary depicts the optimal distribution of optimal thresholds, showcasing their variability, and Fig. 1 in Supplementary showcases an illustrative example where dual-thresholding ensures proper region growth.

The deployment of Modular Neighbors (MN, \cref{sec:modular}) results in improvements across all metrics and reduces $N$, leading to more coherent segmentation. Integrating Region Merging (RM, \cref{sec:reg-merging}) yields a slight decrease in region-growing performance, but significantly minimizes the number of regions $N$. This is necessary to generate larger and more meaningful regions, which leads to enhanced segmentation when including the classifier, as discussed in \cref{sec:classification-efficacy}.

We analyzed the sensitivity of the MARG algorithm to its hyperparameters (see Section 6 of the Supplementary) and found that moderate variations in these parameters do not significantly impact segmentation performance.

%le maintaining high performance across all metrics is the ultimate objective for an effective region-growing algorithm. It signifies that the segmented regions accurately reflect the actual content of the image. This principle will be further validated in the ablation study focused on classification performance in Section \ref{sec:classification-efficacy}, directly demonstrating the impact on the model's ability to classify regions (and thereby segment the images) effectively.

\begin{table*}[t!]
\centering
\resizebox{\linewidth}{!} {
\begin{tabular}{@{}ccccccccccccc@{}}
\toprule
\multicolumn{5}{c}{\textbf{Region-Growing Method}} & \multicolumn{6}{c}{\textbf{Image Segmentation Metrics}} \\
\cmidrule(r){1-5} \cmidrule(l){6-12}
\textbf{Algorithm} & \textbf{Seed Choice} & \textbf{Thresholding} & \textbf{Neighbors} & \textbf{RM} & Acc. & Precision & Recall & F1 & mIoU [\%] & $\Pi$ & $N$ \\
& & & & & $\uparrow$ [\%] & $\uparrow$ [\%] & $\uparrow$ [\%] & $\uparrow$ [\%] & $\uparrow$ [\%] & $\uparrow$ [\%] & $\downarrow$\\
\midrule
RSRG & Random & Global & Cartesian & No & 95.61 & 97.88 & 83.60 & 90.18 & 83.40 & 89.89 & 12.73 \\ 
DTRG & Seed Selection & Global & Cartesian & No & 97.23 & 99.15 & 88.47 & 96.30 & 92.80 & 94.45 & 26.11 \\ 
DTRG & Seed Selection & Adaptive & Cartesian & No & \textbf{98.10} & 99.25 & \textbf{94.41} & 96.77 & 93.87 & \textbf{97.14} & 23.60 \\ 
DTMRG & Seed Selection & Adaptive & Modular & No & \textbf{98.10} & \textbf{99.45} & 94.38 & \textbf{96.85} & \textbf{94.00} & 96.52 & 18.30 \\ 
MARG & Seed Selection & Adaptive & Modular & Yes & 97.87 & 98.97 & 93.77 & 96.30 & 93.37 & 96.52 & \textbf{7.81} \\
\bottomrule
\end{tabular}   
}

\vspace{-0.3cm}
\caption{\textbf{Ablation study of the region-growing algorithm}. This table details the performance of various algorithm configurations on the validation dataset. Note that the introduction of Region Merging (RM) leads to lower region-growing performance. However, RM reduces substantially the number of regions ($N$), which helps later the region classifier to enhance segmentation (\cref{sec:marg-sota}).}
\label{tab:reg-ablation}
\vspace{-0.45cm}
\end{table*}

\vspace{-0.15cm}
\subsection{Region-Growing Qualitative Analysis} \label{sec:qualitative-rg}
\vspace{-0.1cm}

%\Cref{fig:reg-qualitative} presents a visual progression of the distinct region-growing modules to provide a qualitative assessment. It substantiates the incremental enhancements from the baseline RSRG to the advanced MARG approach. To start with, RSRG (second column of \cref{fig:reg-qualitative}) cannot generate regions over all the image (low coverage $\Pi$), escaping from segmentation and leading to a loss of critical information. Notably, as shown in the first instance, the algorithm incompletely captures the blade shape.

\begin{figure}[t!]
\resizebox{\linewidth}{!} {
%\begin{tabular}{@{}c@{}c@{}c@{}c@{}c@{}c|c@{}c@{}c@{}c@{}c@{}c@{}}
\begin{tabular}{@{}c@{}c@{}c@{}c@{}c@{}c@{}}

{\setlength{\fboxsep}{0pt}\fbox{\includegraphics[width=0.0833\linewidth]{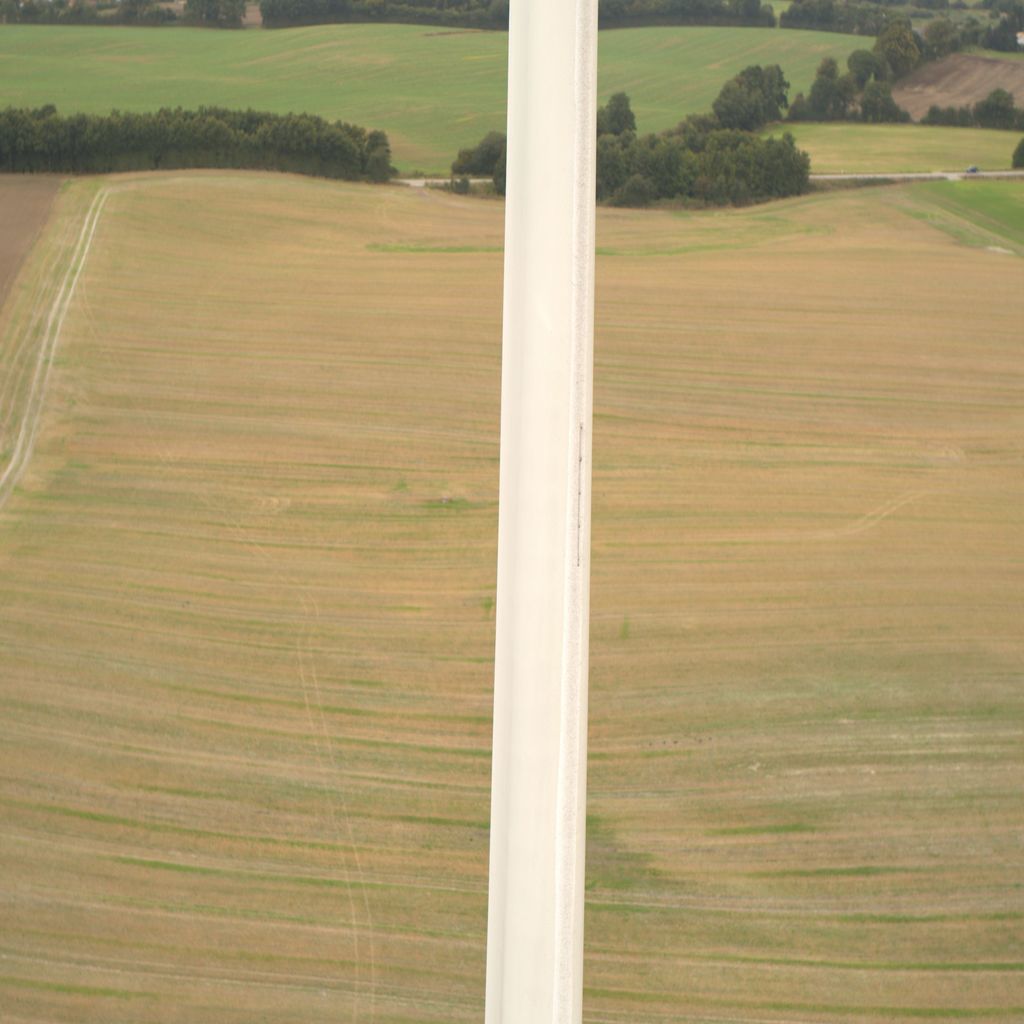}}}&
{\setlength{\fboxsep}{0pt}\fbox{\includegraphics[width=0.0833\linewidth]{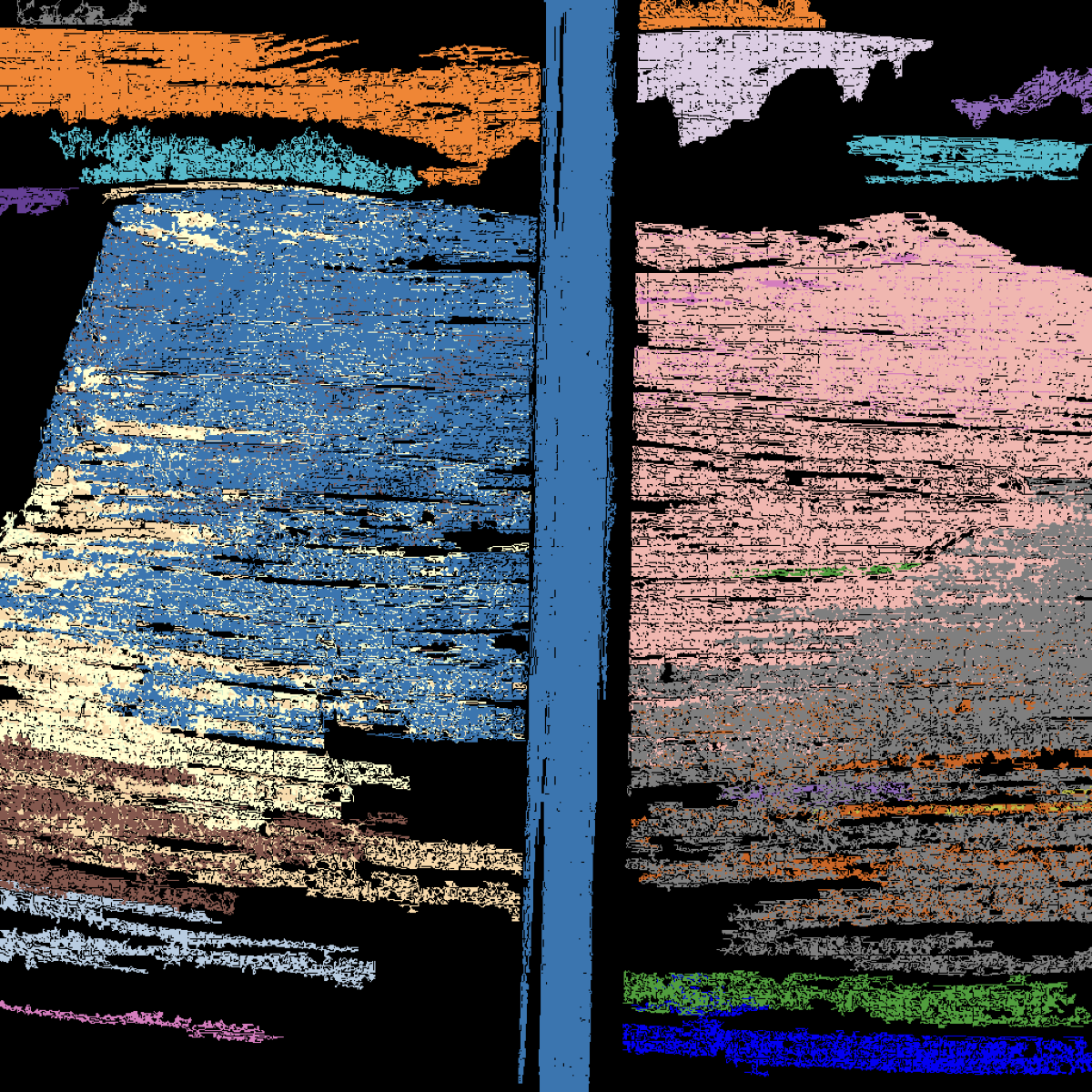}}}&
{\setlength{\fboxsep}{0pt}\fbox{\includegraphics[width=0.0833\linewidth]{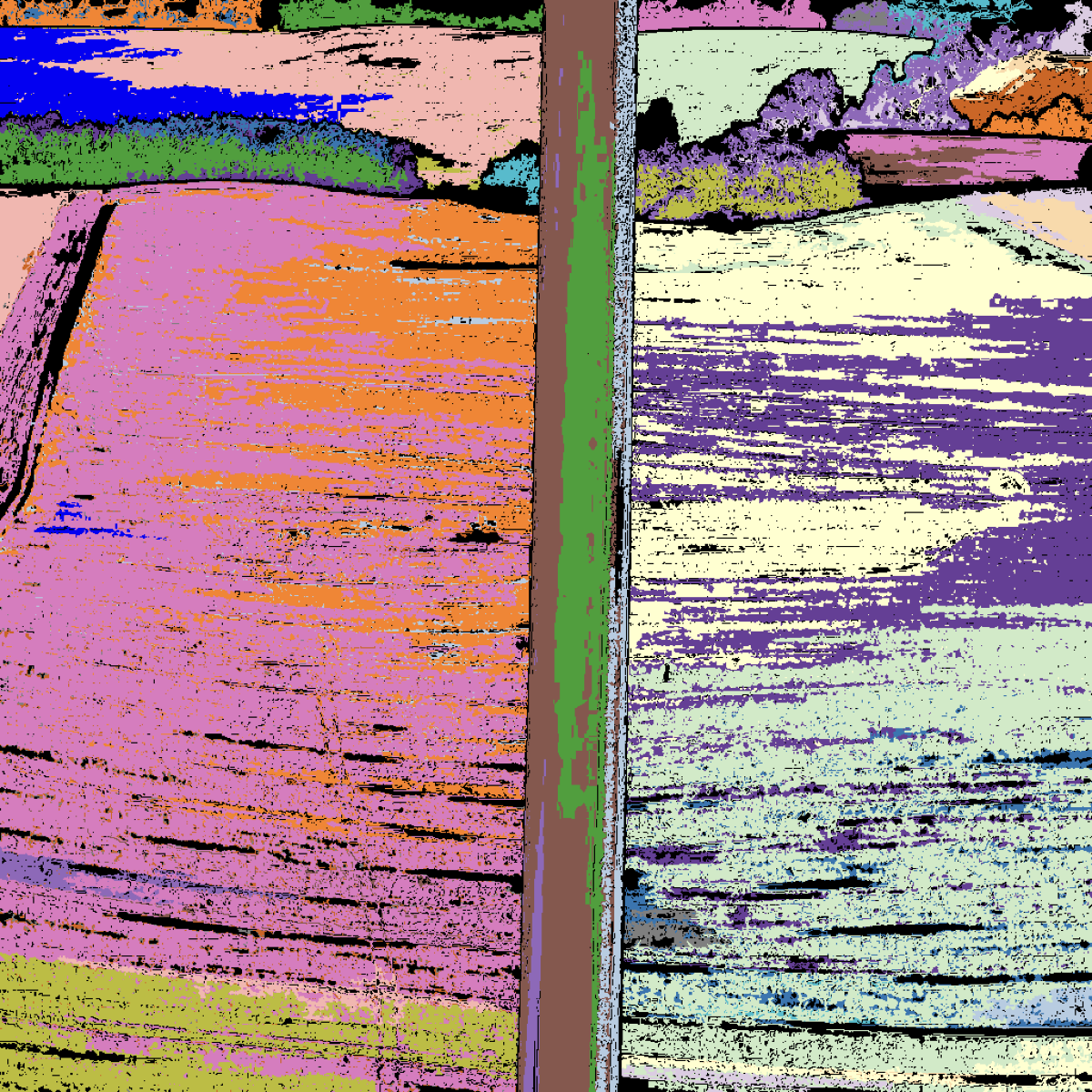}}}&
{\setlength{\fboxsep}{0pt}\fbox{\includegraphics[width=0.0833\linewidth]{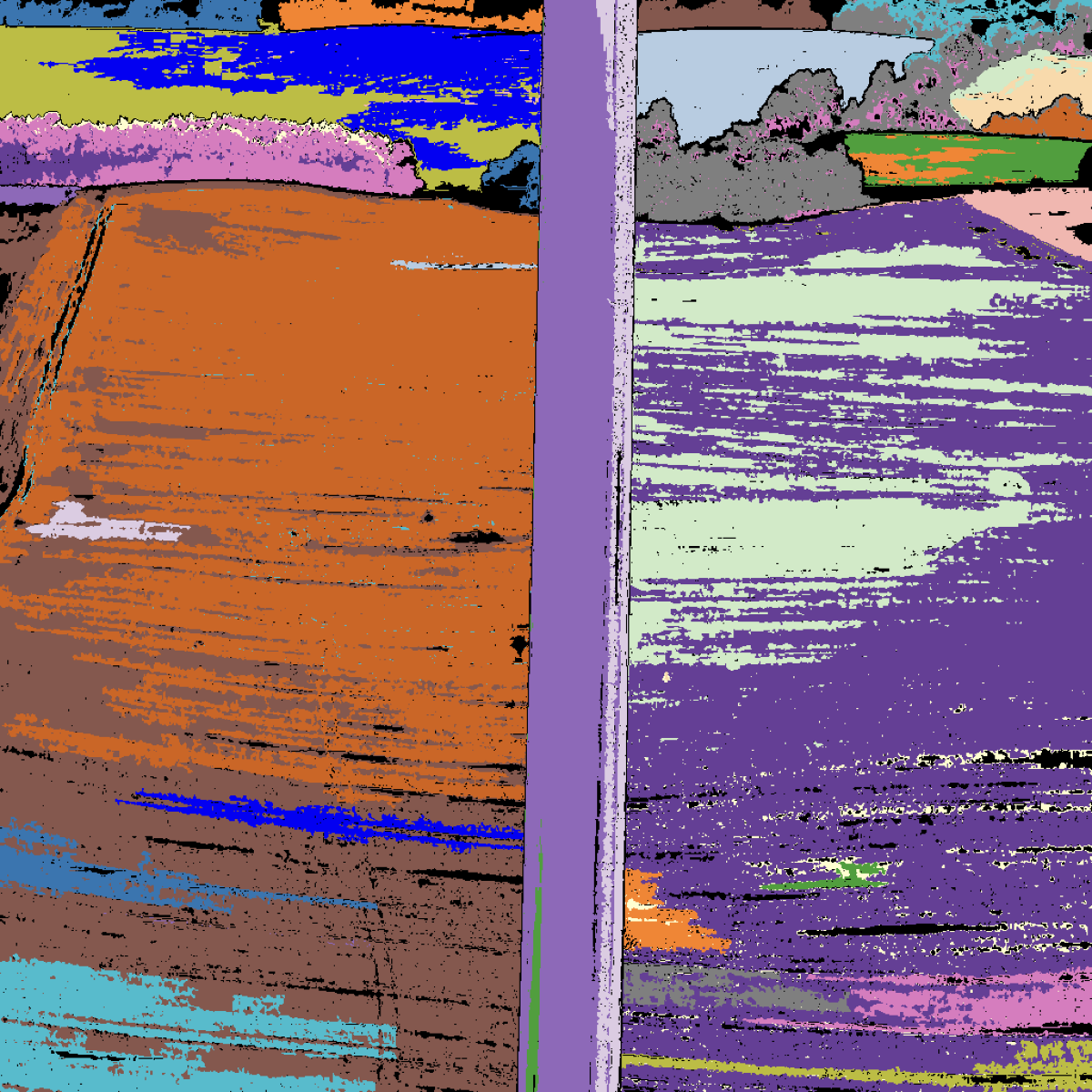}}}&
{\setlength{\fboxsep}{0pt}\fbox{\includegraphics[width=0.0833\linewidth]{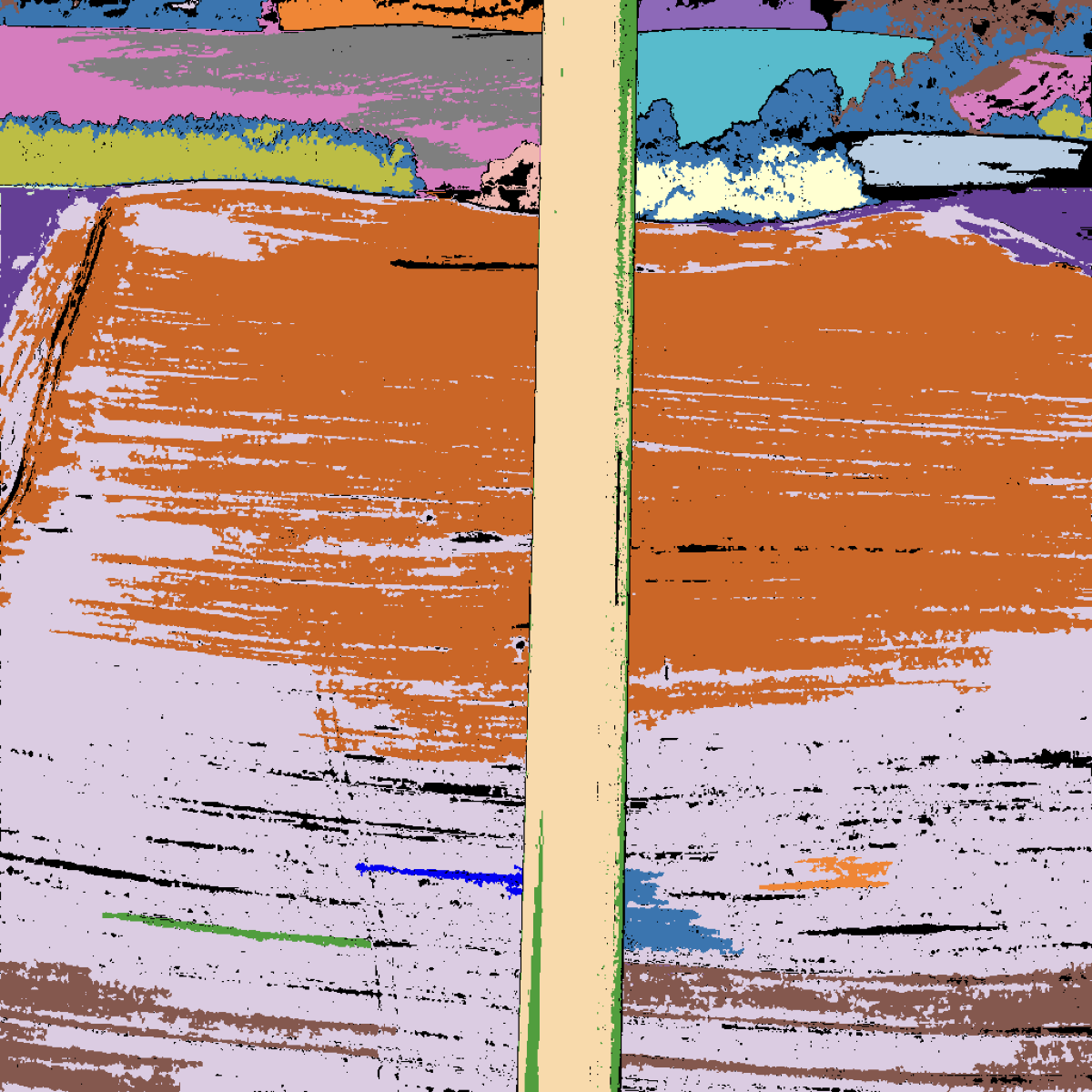}}}&
{\setlength{\fboxsep}{0pt}\fbox{\includegraphics[width=0.0833\linewidth]{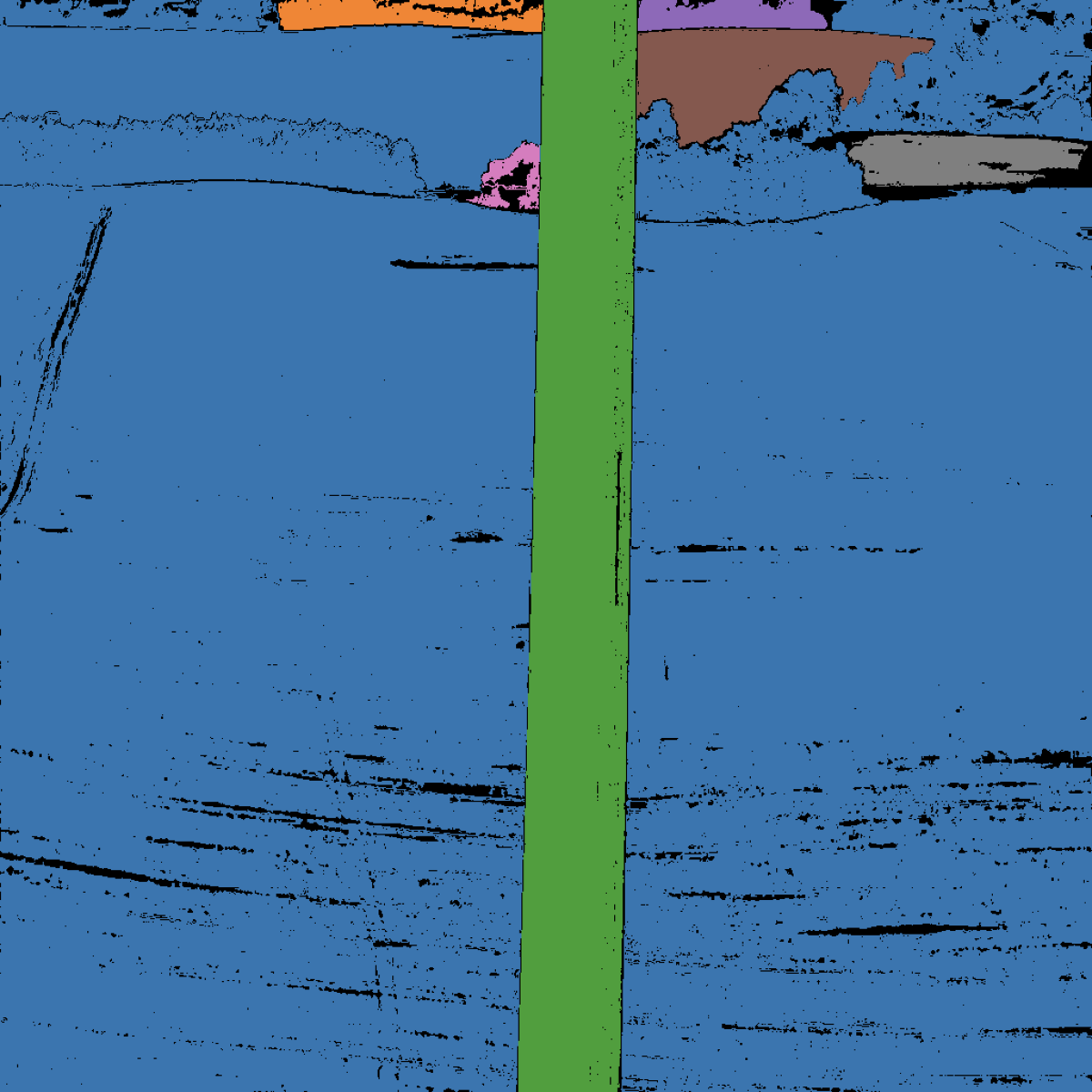}}} \vspace{-0.1cm} \\
{\setlength{\fboxsep}{0pt}\fbox{\includegraphics[width=0.0833\linewidth]{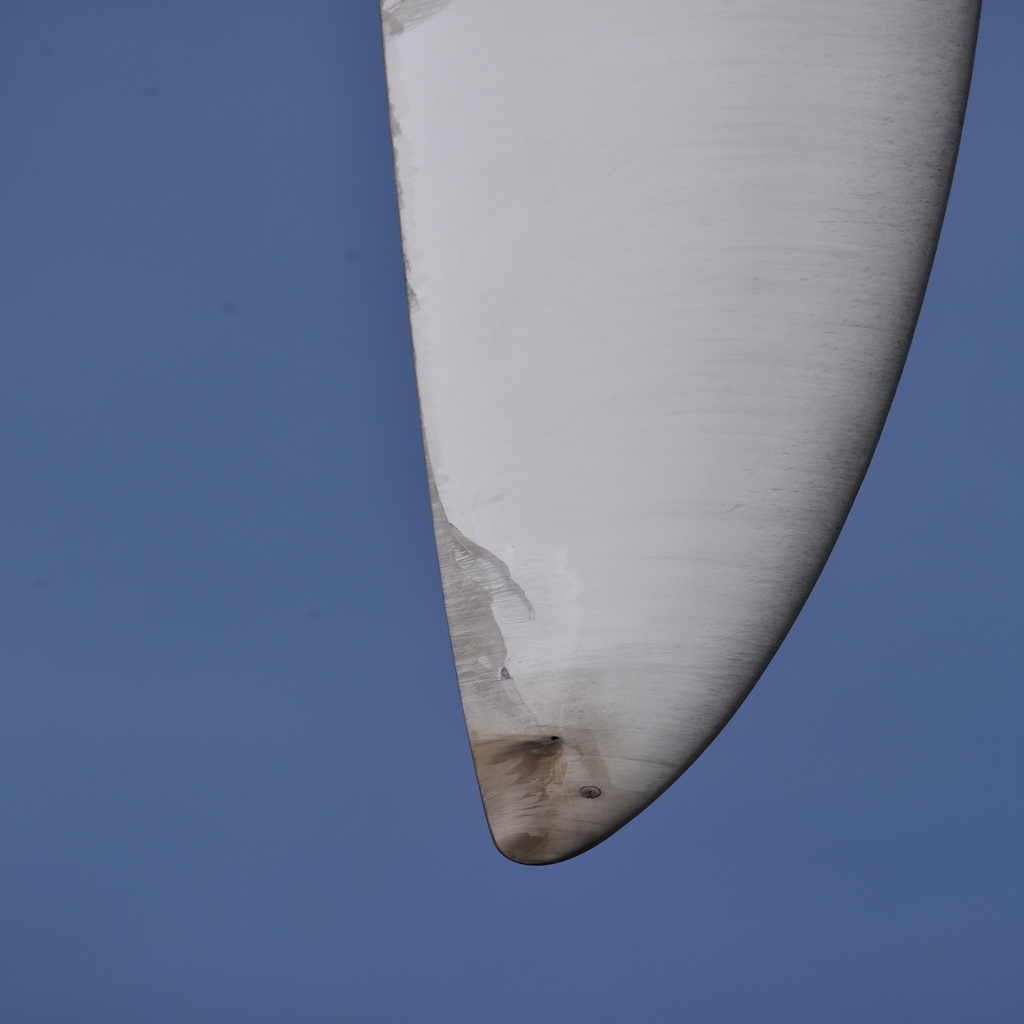}}}&
{\setlength{\fboxsep}{0pt}\fbox{\includegraphics[width=0.0833\linewidth]{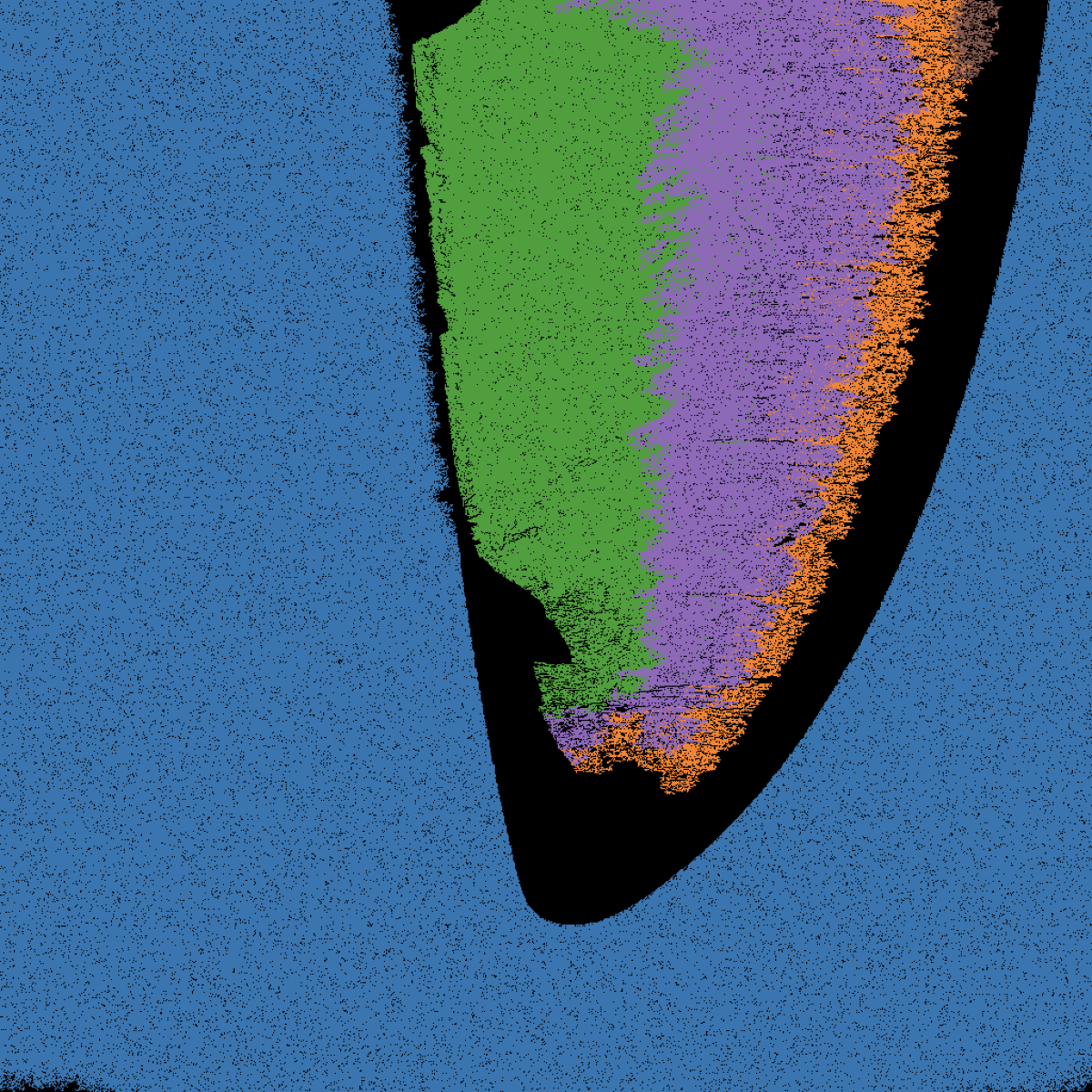}}}&
{\setlength{\fboxsep}{0pt}\fbox{\includegraphics[width=0.0833\linewidth]{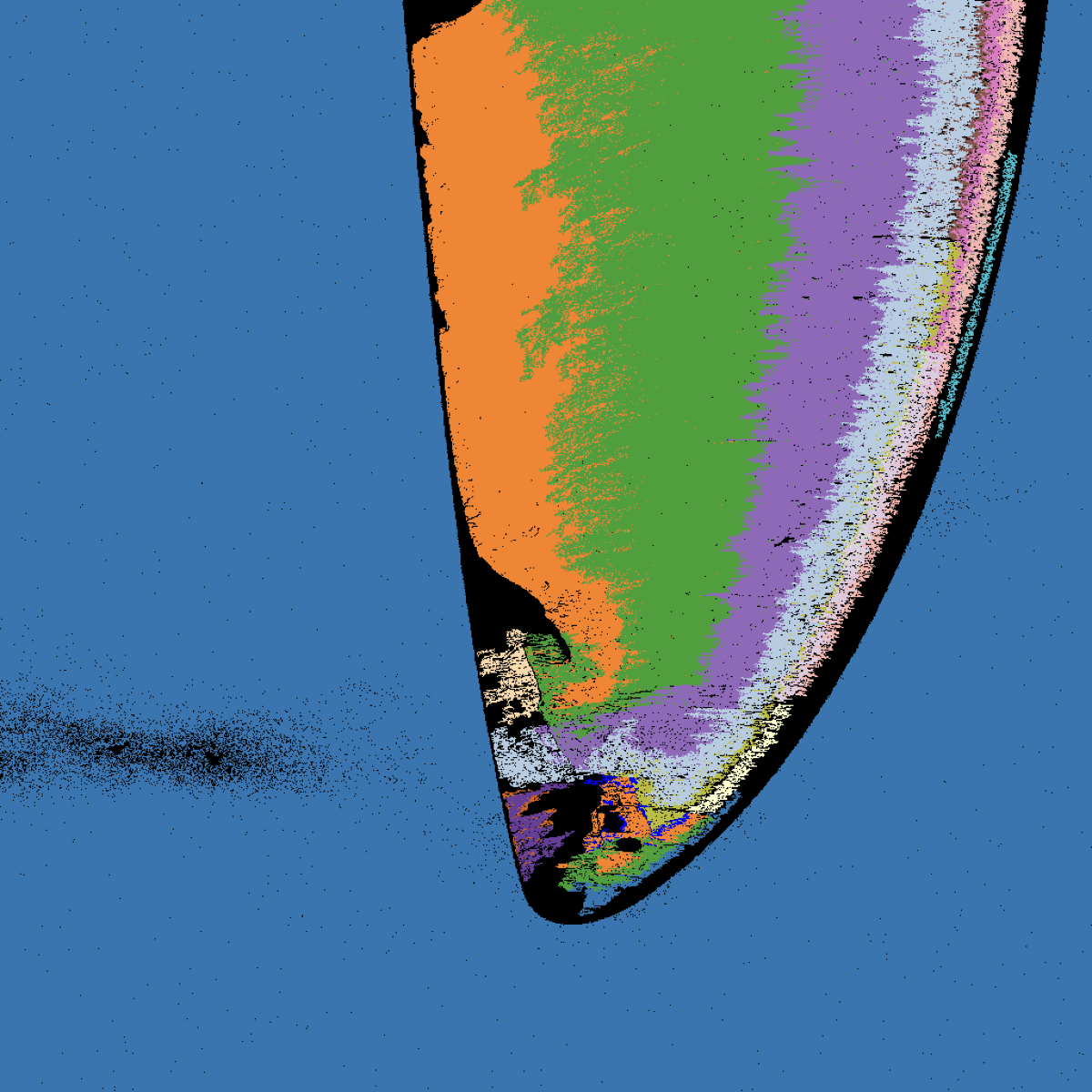}}}&
{\setlength{\fboxsep}{0pt}\fbox{\includegraphics[width=0.0833\linewidth]{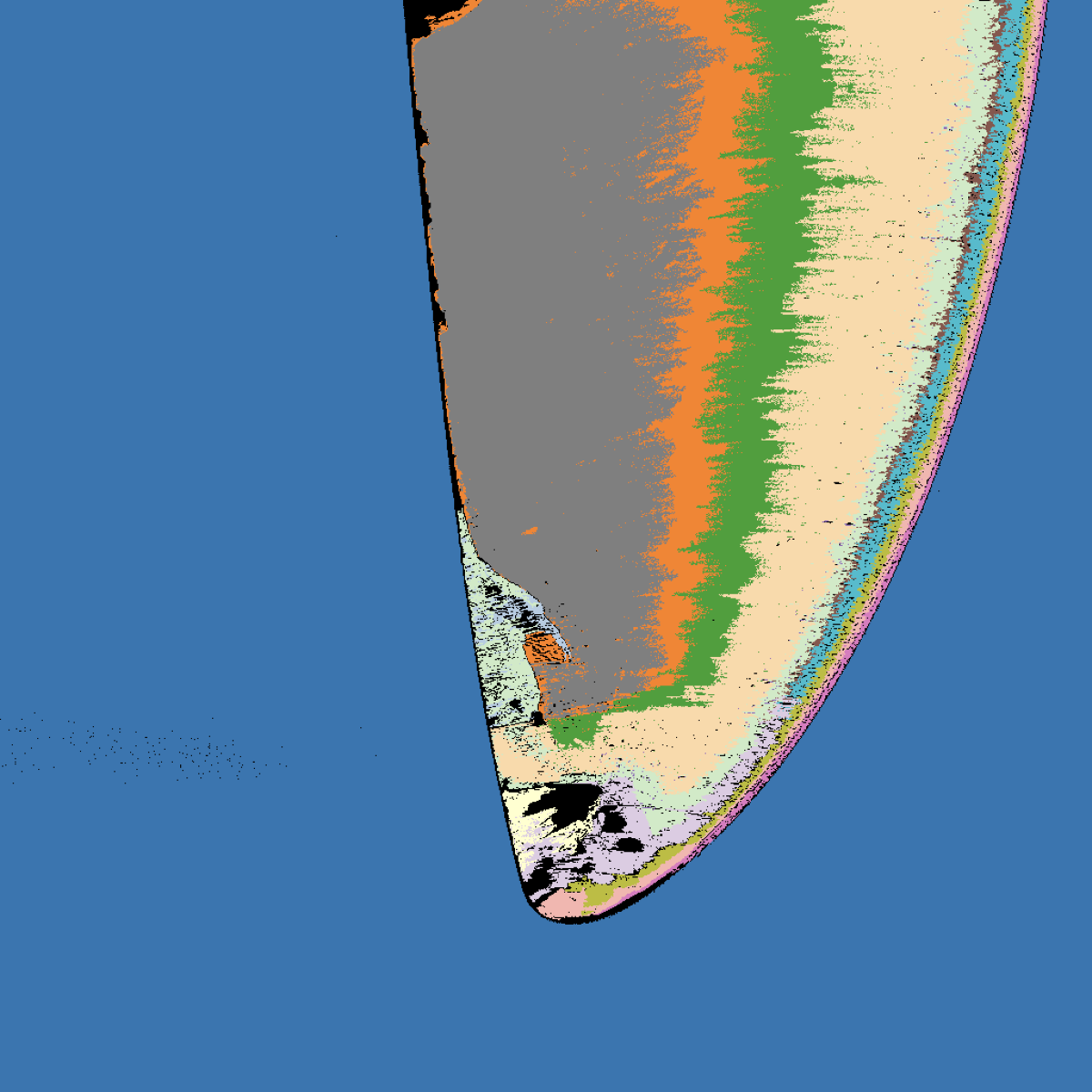}}}&
{\setlength{\fboxsep}{0pt}\fbox{\includegraphics[width=0.0833\linewidth]{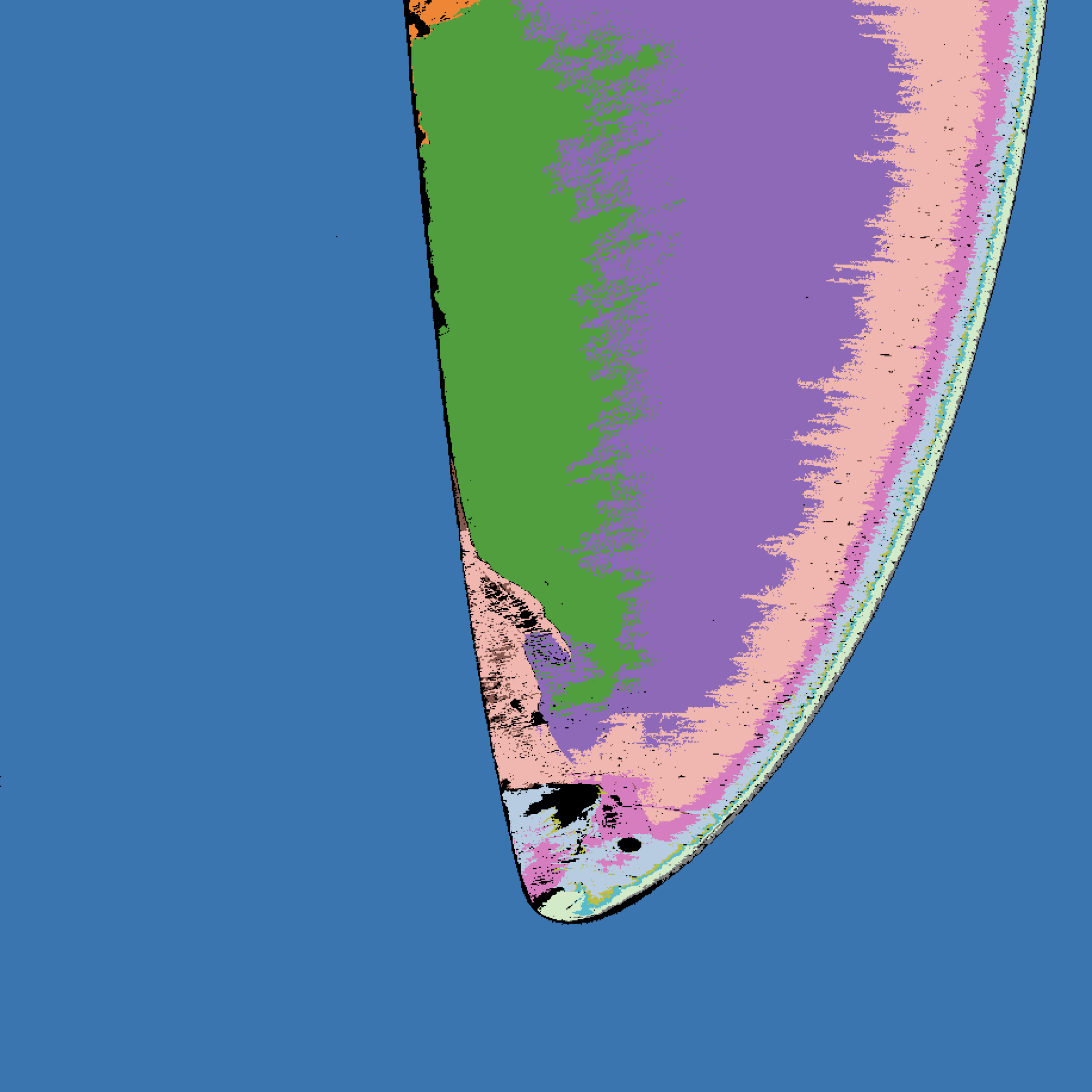}}}&
{\setlength{\fboxsep}{0pt}\fbox{\includegraphics[width=0.0833\linewidth]{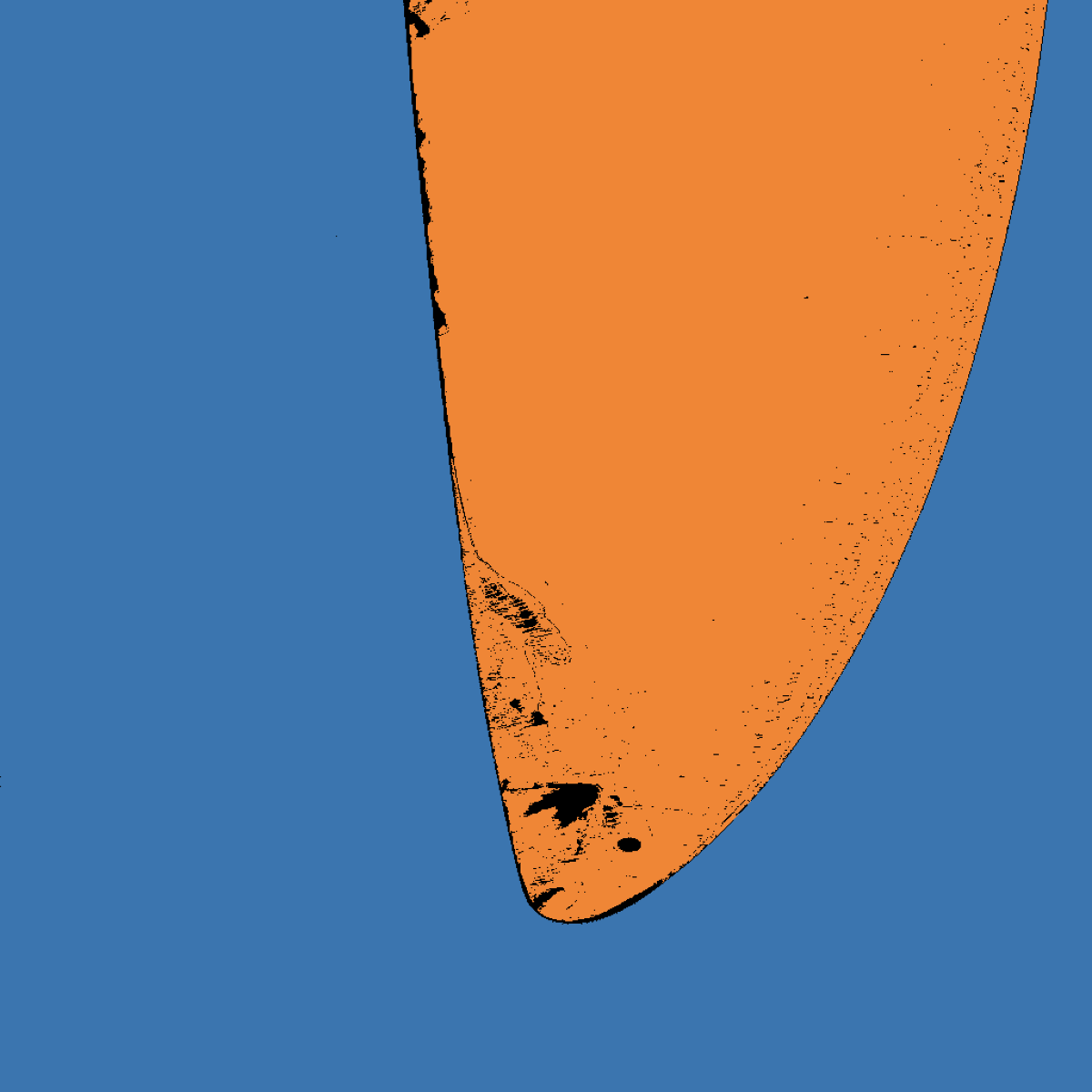}}} \vspace{-0.1cm} \\
{\setlength{\fboxsep}{0pt}\fbox{\includegraphics[width=0.0833\linewidth]{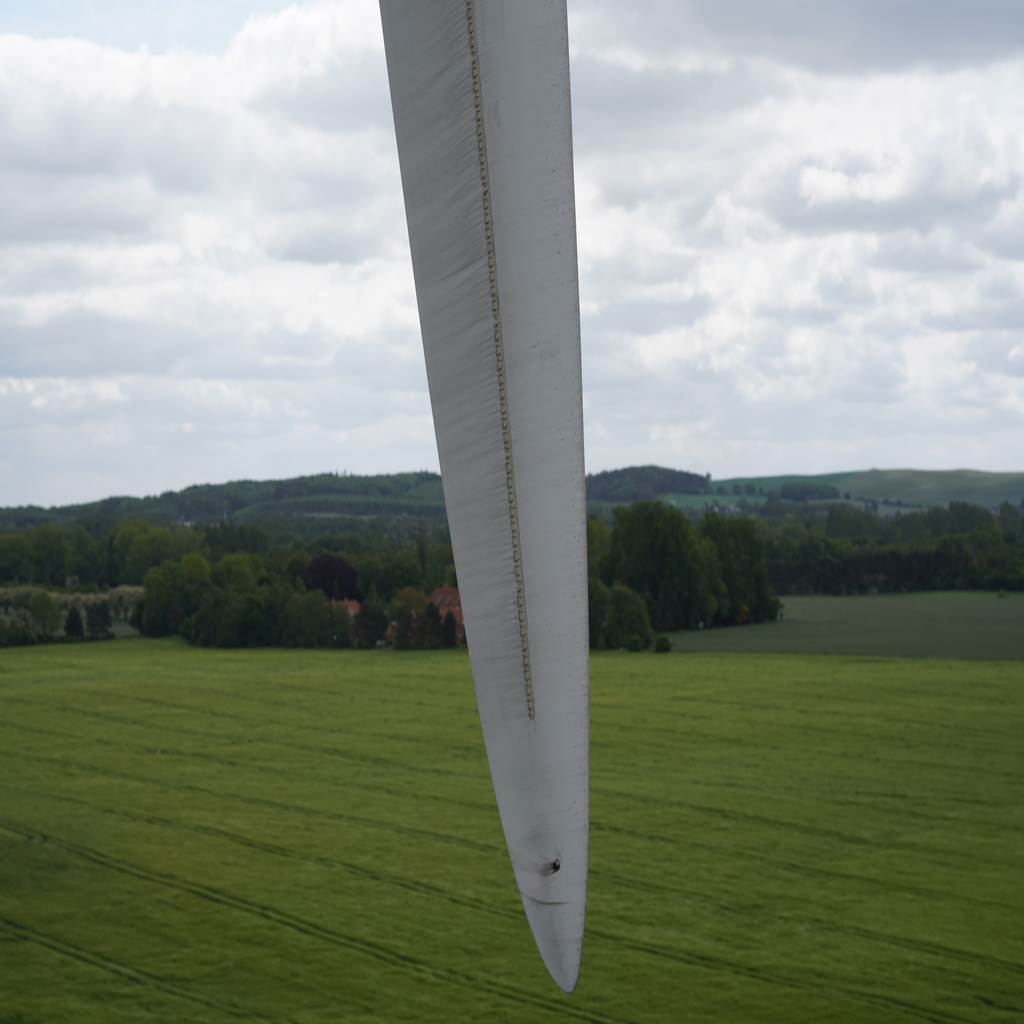}}}&
{\setlength{\fboxsep}{0pt}\fbox{\includegraphics[width=0.0833\linewidth]{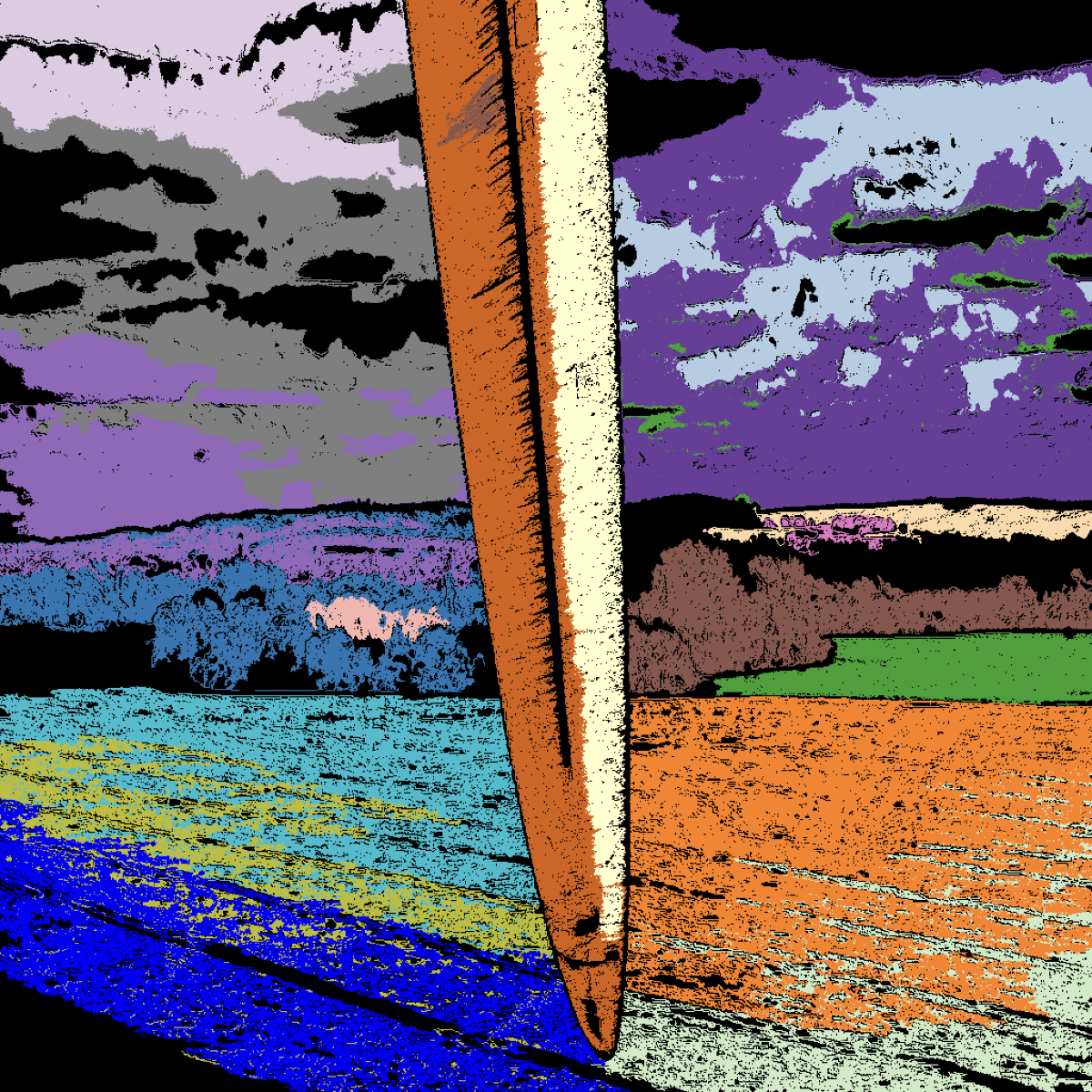}}}&
{\setlength{\fboxsep}{0pt}\fbox{\includegraphics[width=0.0833\linewidth]{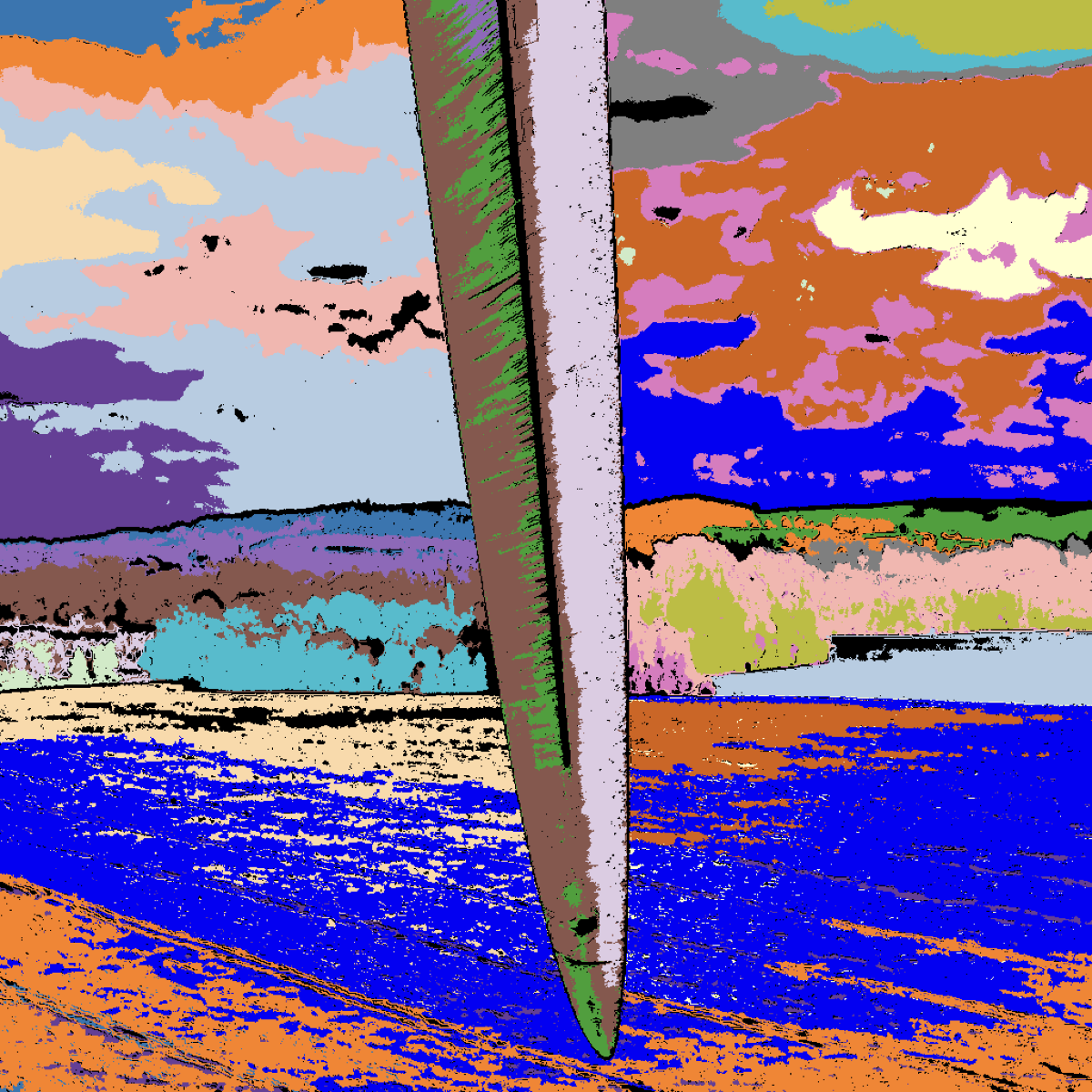}}}&
{\setlength{\fboxsep}{0pt}\fbox{\includegraphics[width=0.0833\linewidth]{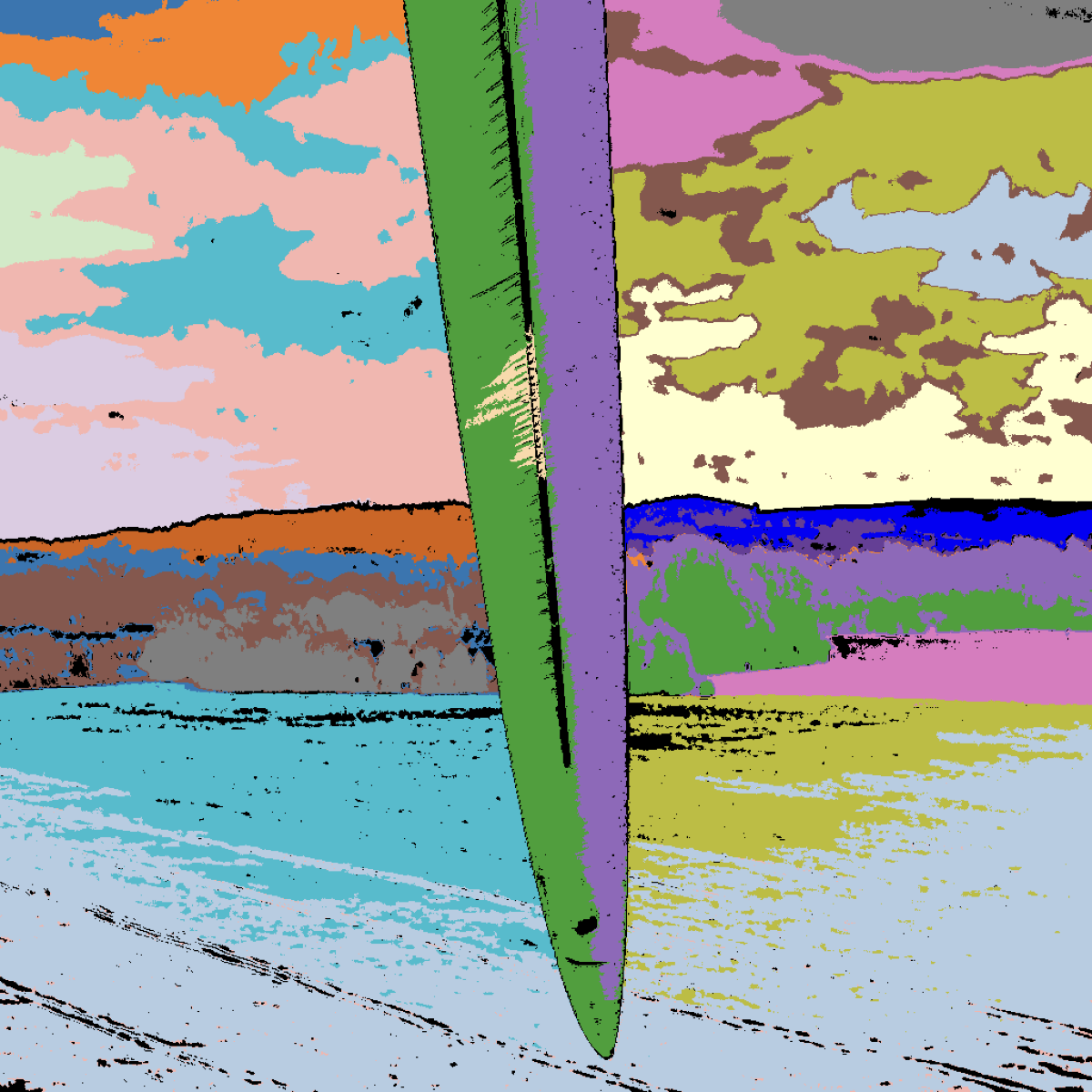}}}&
{\setlength{\fboxsep}{0pt}\fbox{\includegraphics[width=0.0833\linewidth]{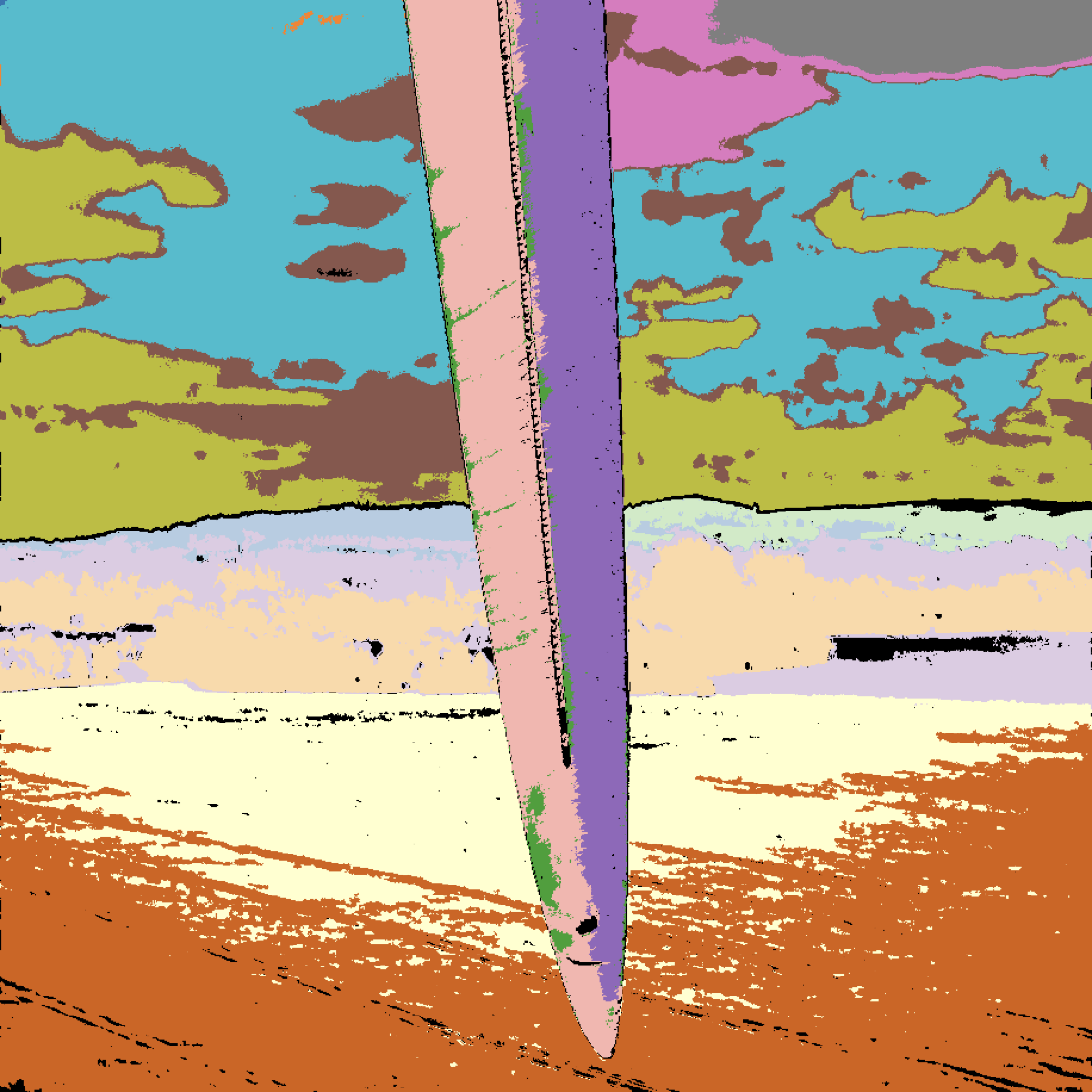}}}&
{\setlength{\fboxsep}{0pt}\fbox{\includegraphics[width=0.0833\linewidth]{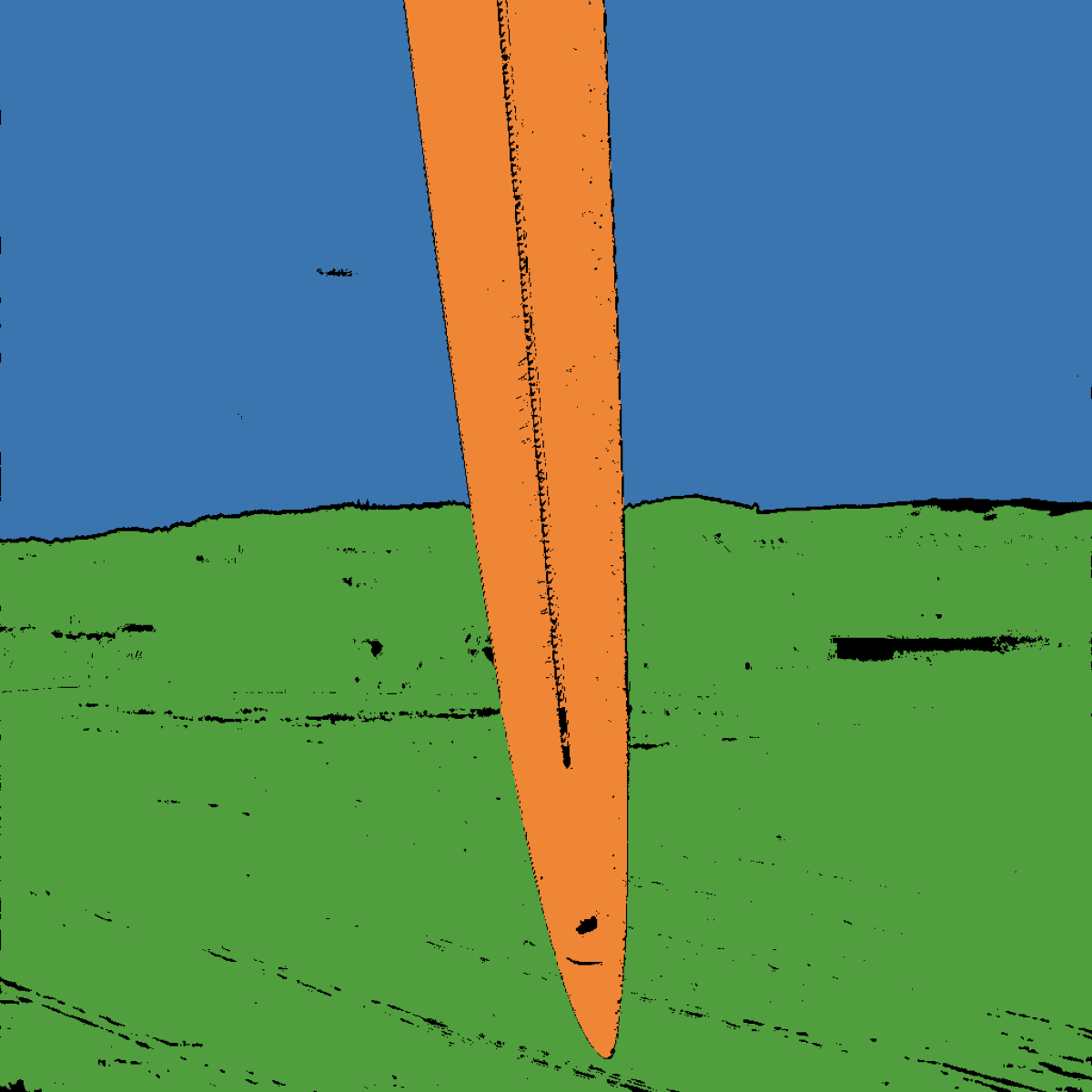}}} \vspace{-0.1cm} \\
{\setlength{\fboxsep}{0pt}\fbox{\includegraphics[width=0.0833\linewidth]{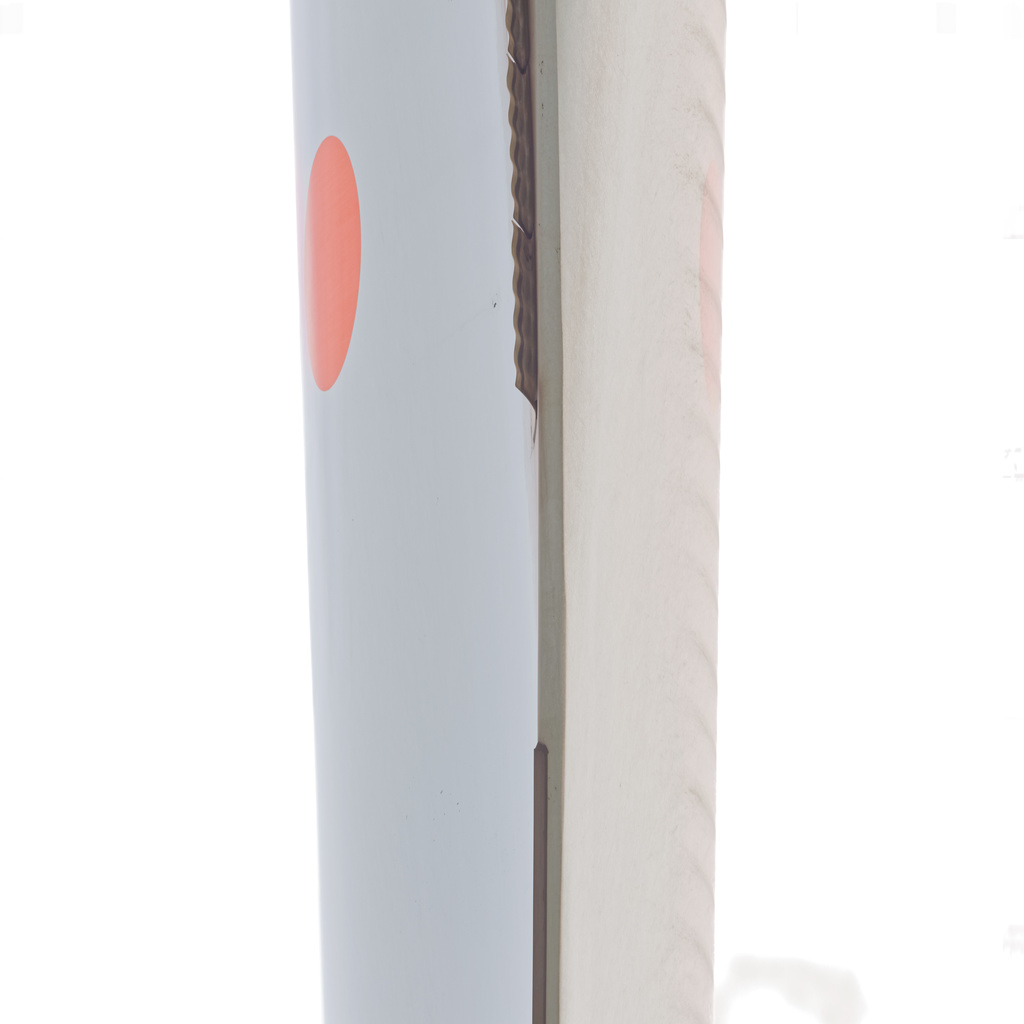}}}&
{\setlength{\fboxsep}{0pt}\fbox{\includegraphics[width=0.0833\linewidth]{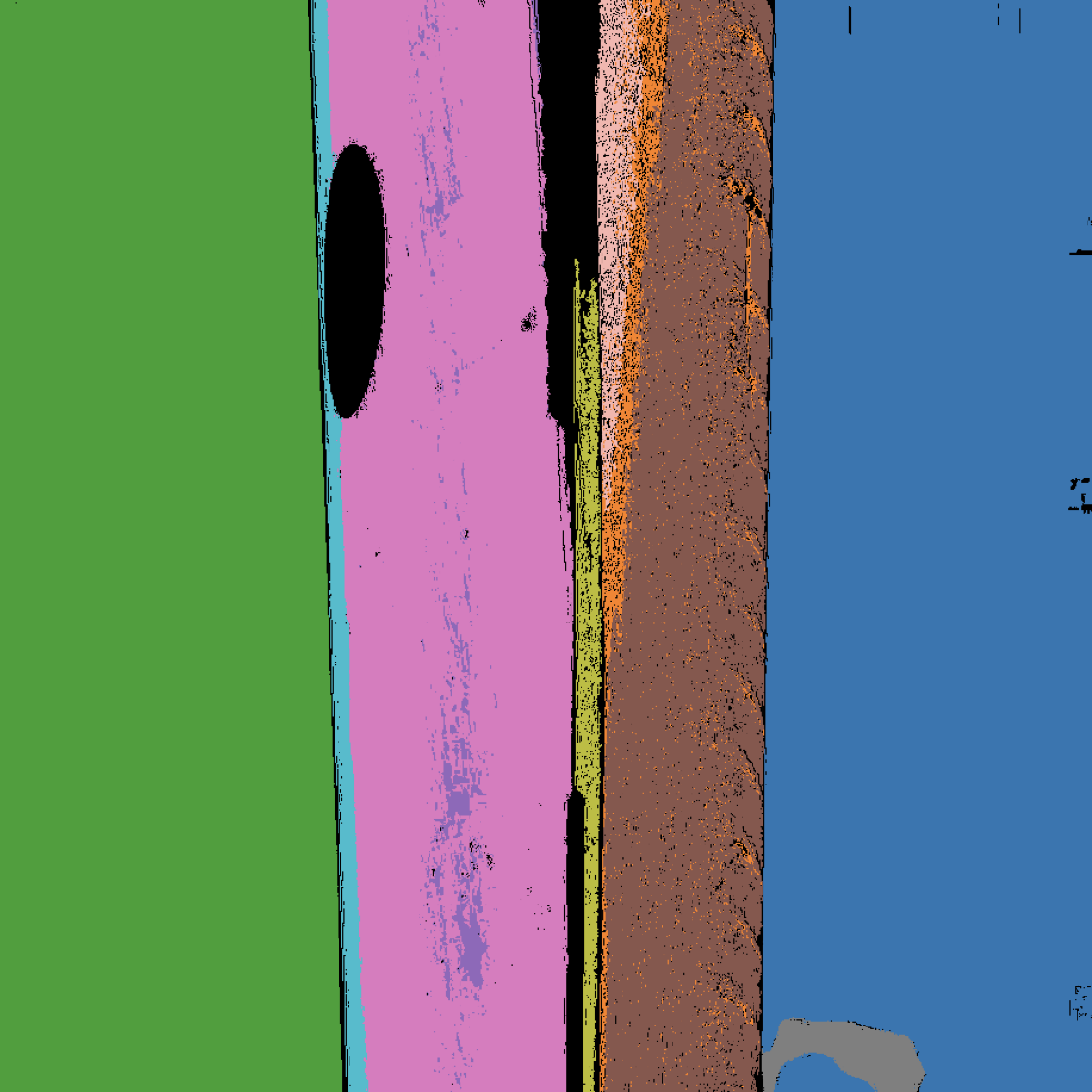}}}&
{\setlength{\fboxsep}{0pt}\fbox{\includegraphics[width=0.0833\linewidth]{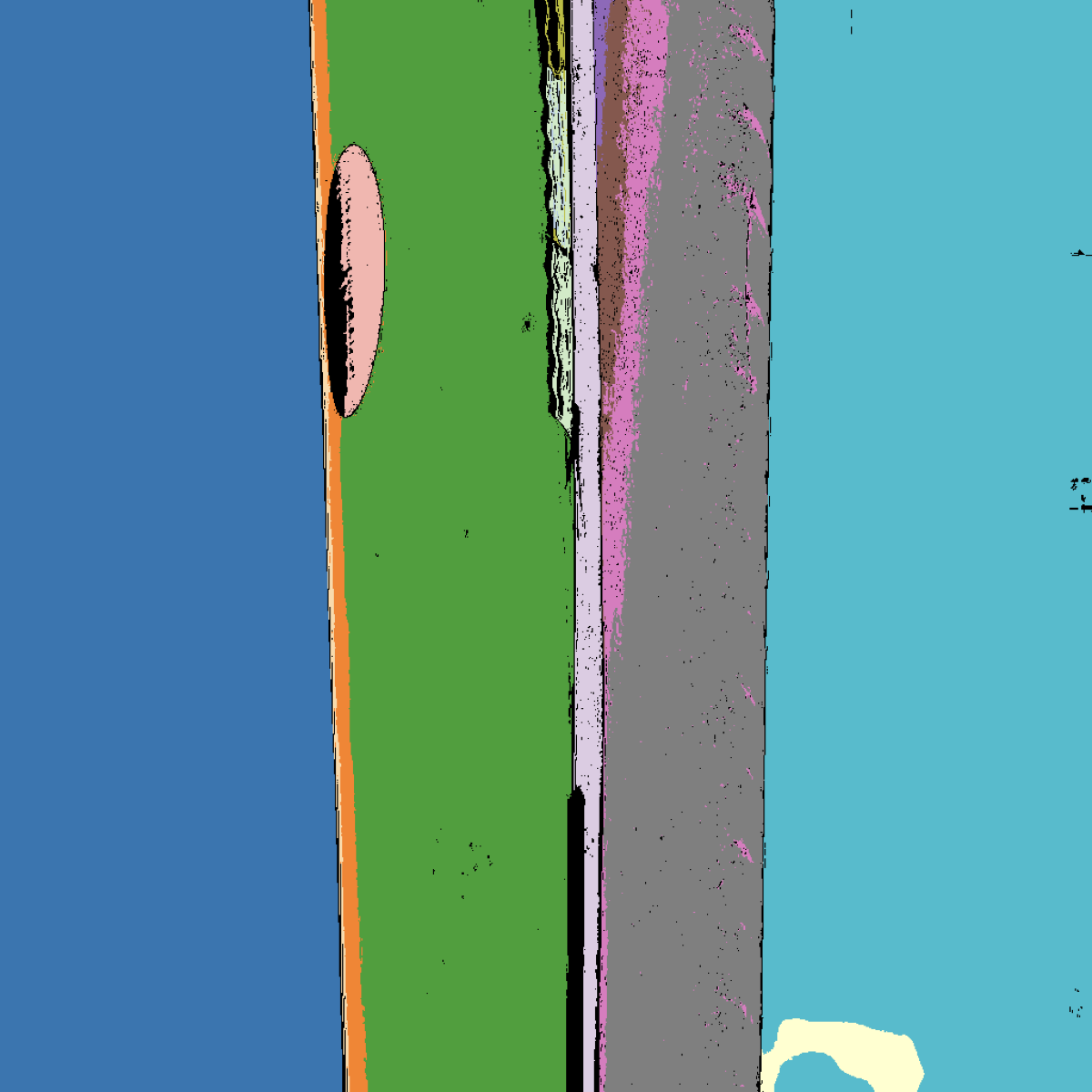}}}&
{\setlength{\fboxsep}{0pt}\fbox{\includegraphics[width=0.0833\linewidth]{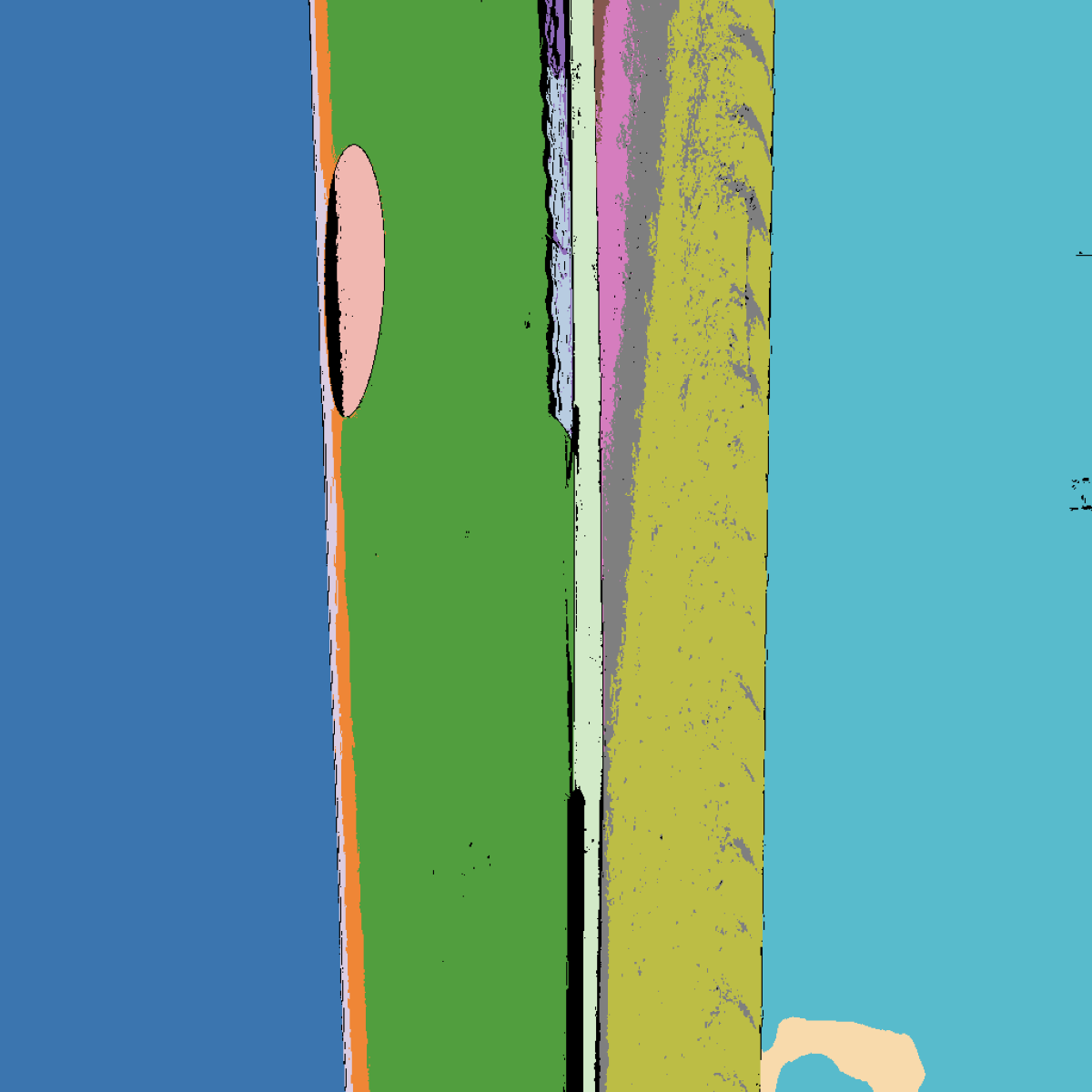}}}&
{\setlength{\fboxsep}{0pt}\fbox{\includegraphics[width=0.0833\linewidth]{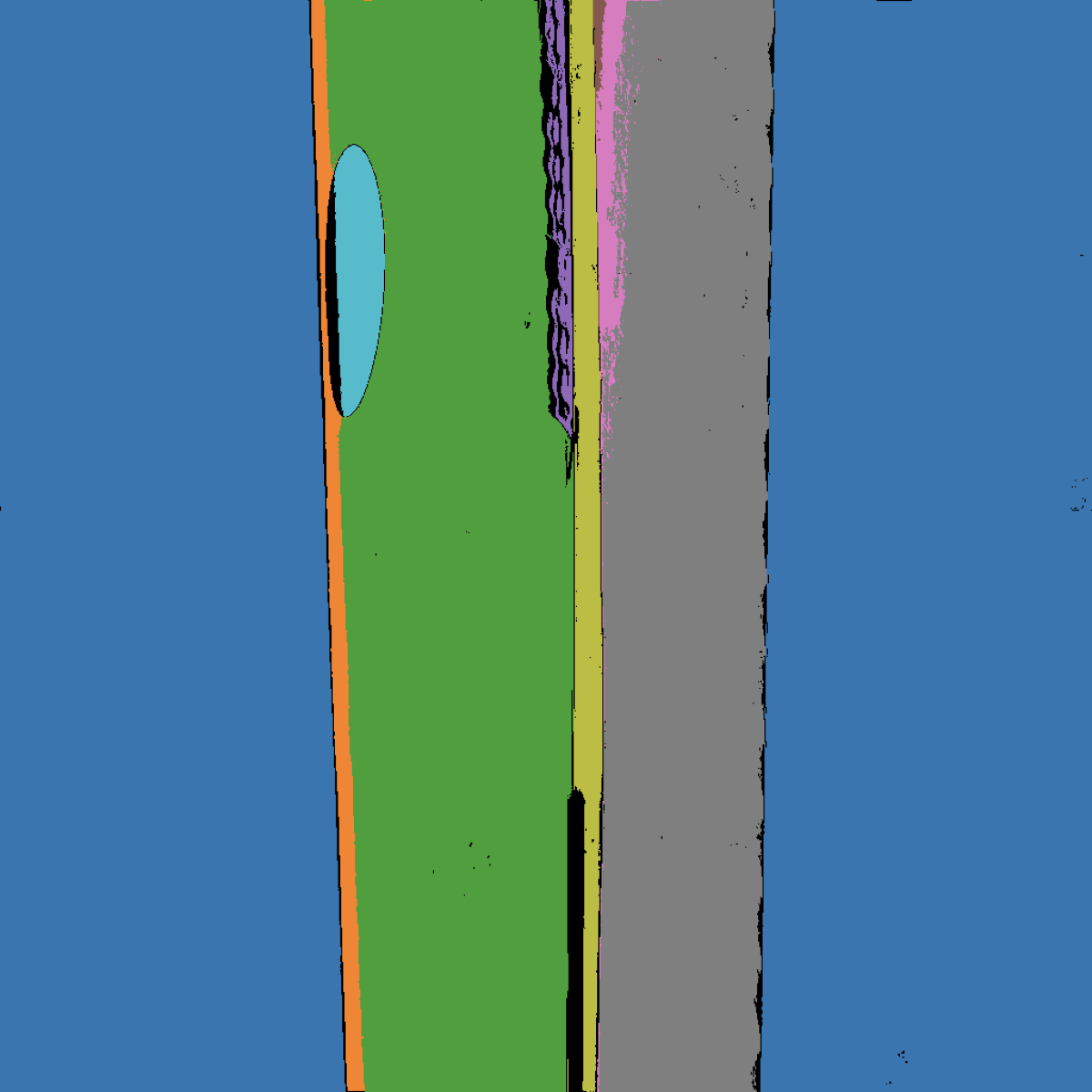}}}&\hspace{-0.17cm}
{\setlength{\fboxsep}{0pt}\fbox{\includegraphics[width=0.0833\linewidth]{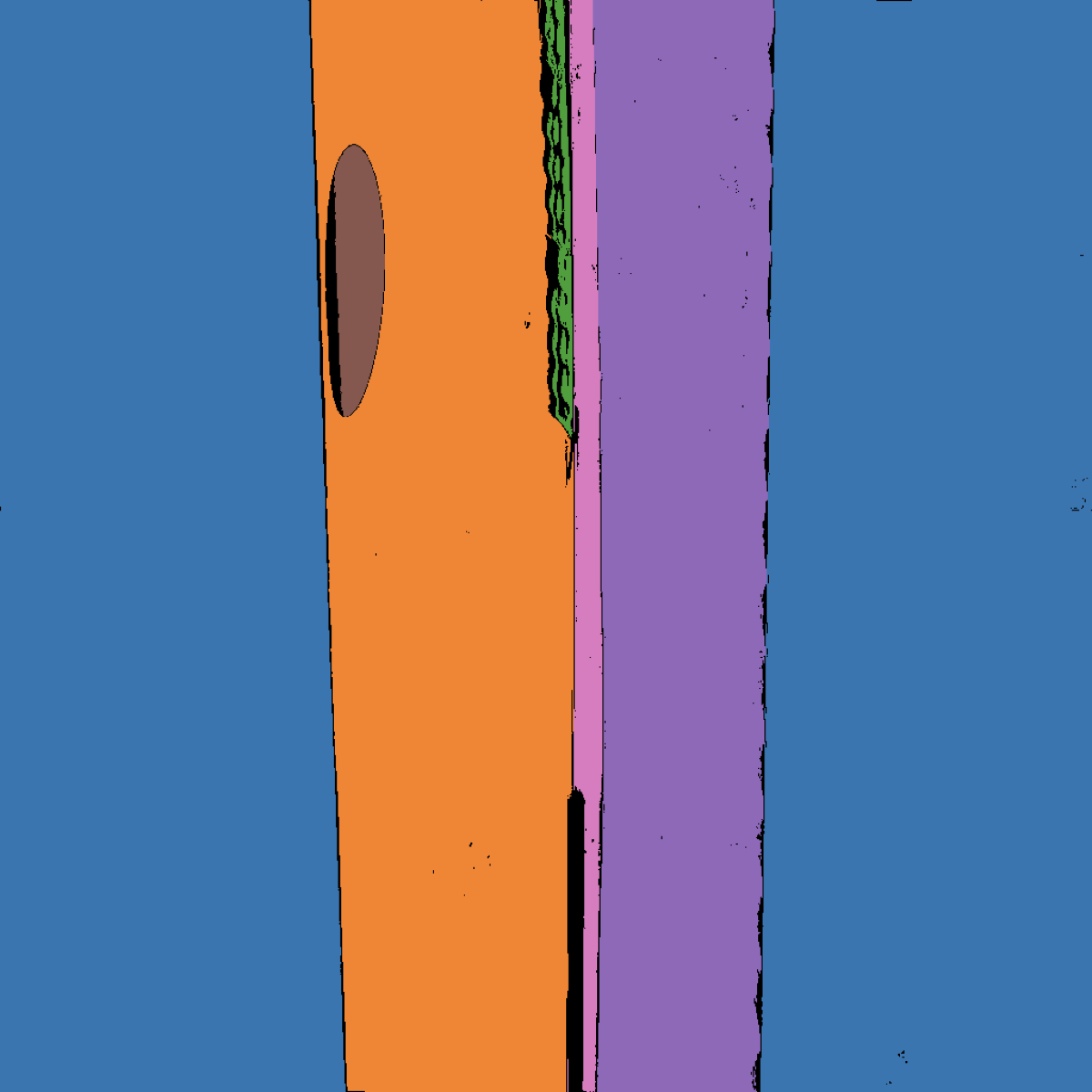}}}

\end{tabular}}
\vspace{-0.5cm}
\caption{\textbf{Region-growing algorithm progression across its modules.} From left to right: original image, RSRG, DTRG + GT (Global Thresholding), DTRG + AT (Adaptive Thresholding), DTMRG + AT, MARG.  Regions are depicted with different colors.}
\label{fig:reg-qualitative}
\vspace{-0.5cm}
\end{figure}

\Cref{fig:reg-qualitative} presents a visual progression of the distinct region-growing modules to provide a qualitative assessment. It substantiates the incremental enhancements from the baseline RSRG to the advanced MARG approach. RSRG (second column of \cref{fig:reg-qualitative}) cannot generate regions over all the image (low coverage $\Pi$), escaping from segmentation and leading to a loss of critical information. Notably, it fails to capture the blade's shape, as shown in the first instance.

%of the region-growing algorithm's performance is essential for discerning the tangible effects of various algorithmic enhancements on segmentation outcomes. Figure~\ref{fig:rg-qualitative} presents a visual progression of methods, providing the chance of performing a direct visual comparison of the changes.

The addition of Seed Selection (\cref{sec:seed-selection} ) yields marked improvements (third column of \cref{fig:reg-qualitative}), particularly in edge delineation and image coverage.  %This method exhibits few unassigned image portions, emphasizing Seed Selection's importance in encompassing the image's salient features.
With Adaptive Thresholding (fourth column of \cref{fig:reg-qualitative}, \cref{sec:adaptive-thresholding}), the algorithm gains a dynamic response to image contrast variances, enhancing boundary precision against diverse backgrounds. This precision, as demonstrated by superior blade segmentation, achieves commendable $\Pi$, leaving almost no pixel unclassified. Its downside is an increase in region granularity.

The integration of Modular Neighbors with the DTMRG method (\cref{sec:modular}) in the fifth alleviates fragmentation while more accurately reflecting the image's intrinsic characteristics. For instance, the fourth instance shows the sky on both sides of the blade coherently recognized as a single region.
    
%MARG's application (\cref{sec:reg-merging}) of Region Merging represents our full algorithm gathering all the presented modules. It slightly sacrifices certain metrics (\cref{tab:reg-ablation}) to facilitate subsequent classification. There is a significant reduction in the count of segmented regions, replaced by expansive, coherent segments that represent distinct features. This is exemplified in the last column of the first instance, where the background fields coalesce into a unified blue region. The algorithm's proficiency is also evident in the third instance, discerning and segments the sky, blade, and field with remarkable clarity.

MARG's with Region Merging (\cref{sec:reg-merging}) represents our full algorithm gathering all the presented modules. It slightly sacrifices certain metrics (\cref{tab:reg-ablation}) to facilitate subsequent classification. There is a significant reduction in the count of segmented regions, replaced by expansive, coherent segments that represent distinct features. This is exemplified in the last column of the first instance, where the background fields coalesce into a unified blue region. The algorithm's proficiency is also evident in the third instance, discerning the sky, blade, and field with remarkable clarity.

\begin{table*}[t!]
\centering
\resizebox{\linewidth}{!} {
\begin{tabular}{@{}ccc|ccccc|cccccc@{}}
\toprule
\multicolumn{3}{c|}{\textbf{Region-Growing Method}} & \multicolumn{5}{c|}{\textbf{Classification Metrics}} & \multicolumn{5}{c}{\textbf{Blade Segmentation Metrics}} \\
 \multicolumn{3}{c|}{} & $N_{\text{Test}}$ & Acc. & Precision & Recall & F1 & Acc. & Precision & Recall & F1 & mIoU & $N$ \\
 \textbf{Algorithm} & \textbf{RM} & \textbf{Labeling Type} & $\downarrow$ & $\uparrow$ [\%] & $\uparrow$ [\%] & $\uparrow$ [\%] & $\uparrow$ [\%] & $\uparrow$ [\%] & $\uparrow$ [\%] & $\uparrow$ [\%] & $\uparrow$ [\%] & $\uparrow$ [\%] & $\downarrow$ \\
\midrule
DTRG + AT & No & Binary  & 3109 & 86.55 & 90.05 & 79.62 & 84.52 & 92.46 & 90.04 & 89.54 & 93.17 & 84.23 & 15.54\\ 
DTRG + AT & Yes & Binary & 1334 & 91.53 & \textbf{99.05} & 88.87 & 93.68  & 96.82 & \textbf{99.02} & 95.23 & 96.51 & 93.46 & 6.67 \\ 
MARG & Yes & Binary & \textbf{1302} & 91.94 & 98.88 & 91.21 & 94.89 & 97.03 & 98.86 & 95.68 & 96.64 & 93.77 & \textbf{6.51}\\ 
MARG & Yes & CutMix~\cite{cutmix} & \textbf{1302} & 92.56 & 98.87 & 91.67 & 95.11 & 97.70 & 97.51 & 96.97 & 97.28 & 94.73  & \textbf{6.51}  \\
MARG & Yes & MixUp~\cite{mixup} & \textbf{1302} & 93.32 & 98.65 & 93.00 & 95.75 & 97.98 & 97.59 & 96.93 & 97.17 & 94.97 & \textbf{6.51} \\
MARG & Yes & RegionMix & \textbf{1302} & \textbf{93.86} & 99.01 & \textbf{93.45} & \textbf{96.15} & \textbf{98.01} & 98.10 & \textbf{96.70} & \textbf{97.52} & \textbf{95.05} & \textbf{6.51} \\ 
\bottomrule
\end{tabular}}
\vspace{-0.3cm}
\caption{\textbf{Performance metrics for blade segmentation using various region-growing configurations}, with ResNet50~\cite{resnet} architecture for the region classifier. Both region classification and blade segmentation metrics are included. $N_{\text{Test}}$ is the overall number of test regions.}
\label{tab:rg+class-ablation}
\vspace{-0.6cm}
\end{table*}
% 2 * (precision * recall) / (precision + recall)

\vspace{-0.2cm}
\section{Comprehensive Evaluation} \label{sec:comprehensive-eval}
\vspace{-0.2cm}

%This section comprehensively evaluates the model's performance on the test dataset, starting with classification efficacy through an ablation study of different region-growing approaches and labeling methods. It then examines the impact on blade segmentation both quantitatively and qualitatively, followed by an assessment of generalization capabilities across ten windfarms. Lastly, the proposed solution is compared with relevant state-of-the-art techniques.

This section evaluates the classification performance of different region-growing and labeling methods, then analyzes their quantitative and qualitative impact on blade segmentation, followed by an assessment of generalization. %Lastly, the proposed solution is compared with relevant state-of-the-art techniques.

\vspace{-0.2cm}
\subsection{Quantitative Evaluation} \label{sec:classification-efficacy}
\vspace{-0.15cm}

The classifier performance varies significantly with different region-growing algorithms used for training. These variations are detailed in \cref{tab:rg+class-ablation}. Introducing Region Merging (RM) significantly improves classification metrics by creating larger, coherent regions that the model can decipher more easily, reducing noise from smaller fragmented regions. Despite a slight decrease in region-growing metrics (\cref{tab:reg-ablation}), this method greatly enhances overall classification and segmentation performance. Additional improvements are seen with Modular Neighbors in the MARG method, and the best classification metrics are achieved with RegionMix labeling, which transitions from binary to continuous labeling, enhancing the model's ability to interpret nuanced information and perform robustly in realistic scenarios.

% \begin{table}[b!]
% \centering
% \resizebox{0.7\linewidth}{!}{
% \begin{tabular}{ccc|ccccc}
% \toprule
% \multicolumn{3}{c|}{\textbf{Region-Growing Method}} & \multicolumn{5}{c}{\textbf{Classification Metrics}} \\ 
% \midrule
% \textbf{Algorithm} & \textbf{RM} & \textbf{Labeling Type} & \textbf{$N_{DS}$}  & \textbf{Acc.}  & \textbf{Precision}  & \textbf{Recall}  & \textbf{F1-S.}  \\
% & & & $\downarrow$ & $\uparrow$ [\%] & $\uparrow$ [\%] & $\uparrow$ [\%] & $\uparrow$ [\%] \\
% \midrule
%  DTRG + AT & No  & Binary    & 3109 & 86.55 & 90.05 & 79.62 & 84.52 \\
%  DTRG + AT & Yes & Binary    & 1334 & 91.53 & \textbf{99.05} & 88.87 & 93.68 \\
%  MARG      & Yes & Binary    & \textbf{1302} & 91.94 & 98.88 & 91.21 & 94.89 \\
%  MARG      & Yes & RegionMix & \textbf{1302} & \textbf{93.86} & 99.01 & \textbf{93.45} & \textbf{96.15} \\
% \bottomrule
% \end{tabular}}
% \caption{\textbf{Classification performance for the distinct incorporated region-growing modules.} $N_{DS}$ represents the overall number of regions for the test dataset.}
% \label{tab:classification_metrics}
% \end{table}

\Cref{tab:rg+class-ablation} shows a direct correlation between region classification and segmentation quality. DTRG + AT provides a baseline recall of 89.54\%, which improves with RM by reducing the number of regions. Using MARG-defined regions boosts performance further, and RegionMix delivers the best results, outperforming other augmentation methods such as MixUp~\cite{mixup} and CutMix~\cite{cutmix}.
%The iterative advancements from DTRG to MARG with RegionMix not only increment the model's capability to classify regions with higher precision but also demonstrate a positive correlation to the fidelity of the final blade segmentation.

\begin{figure*}[t!]
\centering
\includegraphics[width=.804\linewidth]{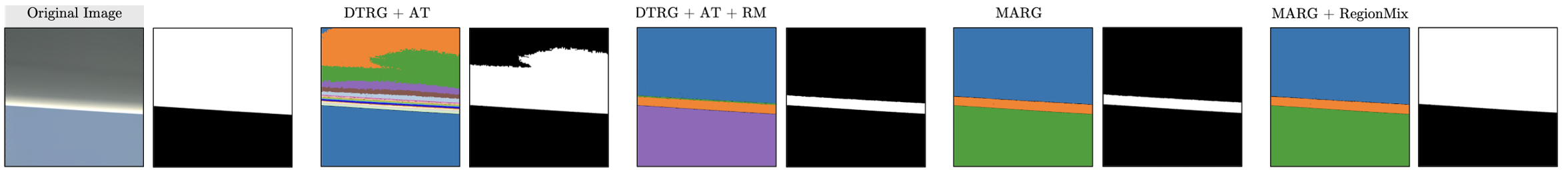}
\includegraphics[width=.8\linewidth]{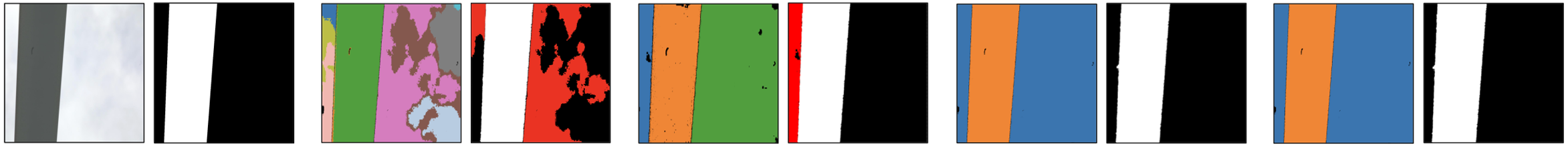}
\includegraphics[width=.8\linewidth]{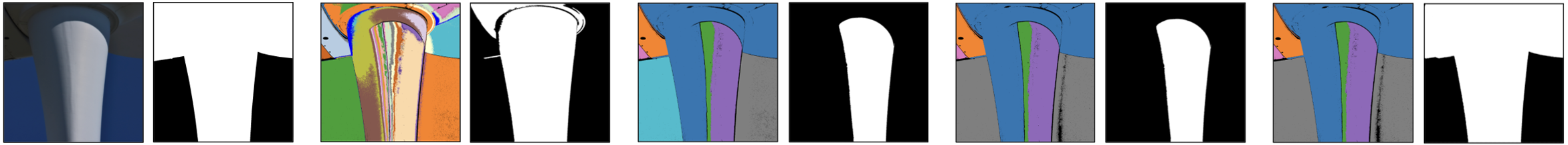}
\vspace{-0.35cm}
\caption{\textbf{Qualitative comparison of region-growing methods.} Each region-growing configuration displays the generated regions in distinct colors and on its right the estimated segmentation mask. Misclassified regions as blade are highlighted in red.}
\label{fig:reg-visual}
\vspace{-0.55cm}
\end{figure*}

\vspace{-0.2cm}
\subsection{Qualitative Evaluation} \label{sec:visual-dissection}
\vspace{-0.15cm}

\Cref{fig:reg-visual} provides a comparative visual analysis of various region-growing modules, showcasing the developed regions alongside the final blade segmentation mask. The second column shows that while DTRG + AT produces more fragmented regions, it results in a less accurate blade mask. The smaller, fragmented regions increase the likelihood of misclassifications, impacting negatively overall segmentation. In contrast, the larger regions generated by RM and MARG methods help the classifier perform better.% However, because these regions are larger, any misclassifications have a more significant impact on the overall segmentation, as seen in the fourth column, where the classifier is trained solely with the generated regions.

%The RegionMix labeling approach is crucial for increasing robustness and ensuring highly accurate classifications. This leads to precise final segmentation masks, as seen in the fifth column. In all three instances, RegionMix consistently produces near-perfect segmentation results. The method enhances segmentation performance by estimating blade coverage rather than relying strictly on binary classification, effectively managing challenges such as off-centered blades and varying blade widths.

RegionMix is crucial for ensuring highly accurate classifications. Its labeling approach leads to precise segmentation masks, see the fifth column. In all three instances, it consistently produces near-perfect results by estimating blade coverage rather than relying strictly on binary classification, effectively managing challenges such as off-centered blades and varying blade widths.

In challenging scenarios such as the second instance in \cref{fig:reg-visual}, where unusual color and contrast distribution confound the model trained on DTRG + AT, the incorporation of Modular Neighbors (MARG) creates discernible regions that simplify classification, improving segmentation.

Despite the general trend of improved classification with larger regions, some cases show that fragmented regions (DTRG + AT) yield better segmentation than RM and MARG with Binary labeling, as seen in the third instance. However, RegionMix consistently excels in demarcating blade regions, including rarely depicted blade roots, showcasing its advanced recognition capabilities.

\vspace{-0.25cm}
\subsection{State-of-the-Art Comparison}
\label{sec:marg-sota}
\vspace{-0.2cm}

\begin{table}[t!]%[h!]
\centering
\resizebox{\linewidth}{!} {
\begin{tabular}{lccccccccc}
\toprule
     \multicolumn{1}{c}{Method} & \multicolumn{1}{c}{Accuracy}  & \multicolumn{1}{c}{Precision} & \multicolumn{1}{c}{Recall} & \multicolumn{1}{c}{F1} & \multicolumn{1}{c}{mIoU}\\
    \multicolumn{1}{c}{} & {[\%]} & {[\%]} & {[\%]}  & {[\%]}  & {[\%]}  \\
    \midrule
    DeepLabv3+~\cite{deeplabv3+}  & 94.14 & 96.36 & 87.38 & 89.03 & 87.47 \\  %90.74
   SW~\cite{sw} & 93.48 & 93.57 & 91.71 & 91.37 & 87.44 \\ %89.71
   ResNeSt~\cite{resnest} & 94.23 & 96.84 &91.47 & 92.77 & 89.63  \\ %91.20
   U-NetFormer~\cite{unetformer} & 96.20 & 97.31 & 93.51 & 94.42 & 91.75\\ %93.65
   BiRefNet~\cite{birefnet} & 95.65 & 98.37 & 92.52 & 94.57 & 92.38 \\ %epoch_150
   SAM*~\cite{sam} & 94.36 & 97.29 & 91.22 & 92.60 & 91.66  \\ 
   SAM2*~\cite{sam2} & 97.86 & 97.46 & 94.82 & 95.84 & 94.82 \\
   CLIPSeg~\cite{clipseg} &  82.70 & 77.02 & 75.52 & 74.29 & 75.09 \\ 
   DiffSeg*~\cite{diffseg} & 96.37 & 89.74 & 85.73 & 86.40 & 72.14 \\ 
   EfficientFormer~\cite{efficientformer} & 96.42 & 95.47 &  93.63 & 94.55 & 93.81 \\ 
   MobileViT~\cite{mobilevit} & 96.14 & 95.44 & 93.33 & 94.38 & 93.47 \\
   Mask2Former~\cite{mask2former} & 96.68 & 95.63 & 93.89 & 94.76 & 93.72 \\
   BU-Net~\cite{bunet} & 97.39 & \textbf{99.42} & 93.35 & 95.73 & 93.80\\ %95.53
   Mask2Former-FreqFusion~\cite{freqfusion} & 97.35 & 98.02 & 96.12 & 96.67 & 94.65 \\ 
   MARG+RegionMix & \textbf{98.25} & 98.43 & \textbf{96.49} & \textbf{97.67} & \textbf{95.33}  \\ 
\bottomrule
\end{tabular}
}
\vspace{-0.35cm}
\caption{\textbf{Quantitative comparison on blade segmentation} using EfficientNet-B4~\cite{efficientnet} architecture for the region classifier. Models highlighted with an asterisk assume a perfect classifier. }
\label{tab:marg-unet-compare}
\vspace{-0.4cm}
\end{table}

Unlike zero-shot models, our algorithm is fully unsupervised, requiring no training data to generate regions. We benchmark its blade segmentation performance against supervised, zero-shot, and prior wind turbine segmentation methods. In contrast to MARG, zero-shot approaches typically rely on deep features; we compare SAM~\cite{sam}, SAM2~\cite{sam2}, and DiffSeg~\cite{diffseg} for region proposals under a perfect classifier assumption (following \cref{eq:marg-sim}), and report the best results for CLIPSeg~\cite{clipseg} across different prompts.

Our MARG+RegionMix approach simplifies segmentation by reducing it to region classification, facilitating its explainability since MARG does not rely on deep learning techniques and binary classification is inherently much simpler. \Cref{tab:marg-unet-compare} shows that our region classification approach achieves top-tier results, because it is less reliant on the quantity of training data compared to other approaches. Notably, it also offers a more balanced precision-recall trade-off compared to heavily data-reliant learning-based techniques. An detailed analysis of MARG+RegionMix's quantitative gains and its potential impact on defect detection is provided in Section 11 of Supplementary.

%These methods tend to overfit rapidly due to their reliance on large data volumes for robustness and generalizability. In contrast, the MARG+RegionMix approach, by continuously learning with diverse image-region combinations, mitigates overfitting and enhances generalization. This makes it a strong alternative for situations where interpretability and simplicity are crucial.

\vspace{-0.15cm}
\subsection{Generalization Across Windfarms} \label{sec:marg-windfarm}
\vspace{-0.1cm}

To analyze the robustness of our region-growing classifier, we analyze the test set composed of 200 images from ten diverse windfarms; see Section 12 of Supplementary. Evaluating each windfarm individually allows us to study generalization, simulating the conditions of a new blade inspection and providing insights into robustness and adaptability.

\begin{figure}[t!]%[h!]
\centering
\includegraphics[width=1\linewidth]{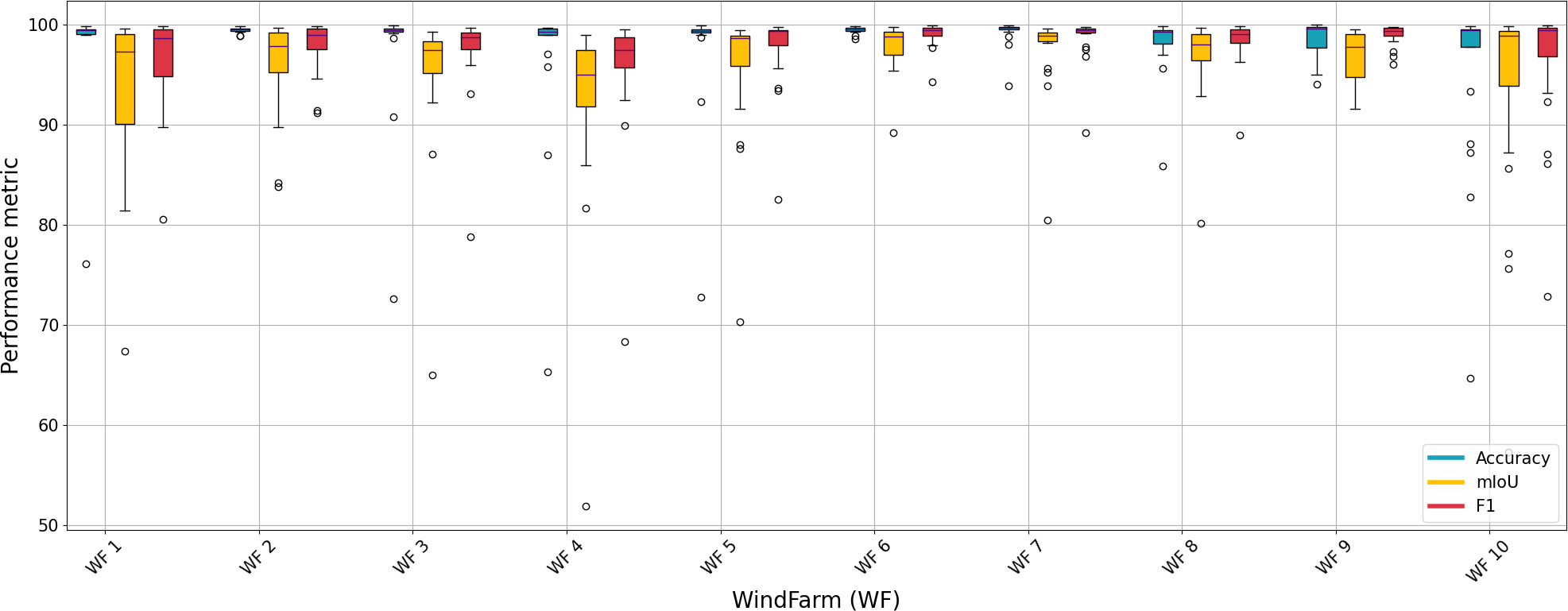} 
\vspace{-0.65cm}
\caption{\textbf{Segmentation results per windfarm on the test set}. Quantitative results are detailed in Table 3 of Supplementary.}
\label{fig:reg-windfarm}
\vspace{-0.6cm}
\end{figure}

\Cref{fig:reg-windfarm} illustrates the segmentation performance of MARG+RegionMix through boxplot distributions. The consistency observed across all windfarms underscores our framework’s effectiveness in blade segmentation, demonstrating the model’s adaptability and reliable performance under diverse environmental conditions.

While robust overall, some segmentation instances appear as outliers in \cref{fig:reg-windfarm}, reflecting occasional underperformance. These cases are examined in Section 7 of Supplementary, along with an analysis of the computational cost that is included in Section 8 of Supplementary. Notably, MARG is designed for CPU execution and lacks GPU acceleration, distinguishing it from other algorithms.% in terms of computational efficiency.

%% file: WACV26/chapters/Conclusion.tex
\vspace{-0.2cm}
\section{Conclusion}
\vspace{-0.15cm}

This work encapsulates the design and empirical substantiation of an image segmentation framework tailored for paving the way for wind turbine analysis. The methodology, characterized by a novel unsupervised region-growing algorithm with a CNN classifier, responds adeptly to the limitations posed by small datasets—a prevalent challenge in traditional pixel-dense segmentation that rely heavily on extensive annotated data, achieving better results under the same conditions while offering greater explainability.

The framework's components, including Dual-Threshold Modular Region-Growing, Adaptive Thresholding, and Region Merging, synergize to generate salient and expressive regions. By adeptly utilizing combined regions, RegionMix data augmentation proves effective in enhancing the classifier's discernment of regions.

The efficacy of the approach, which simplifies image segmentation to binary classification, is substantiated by the model’s exceptional generalization capabilities across diverse windfarm environments, confirming its robustness.

This research sets a precedent for blade image segmentation where limited data represents a challenge, underscoring its practical value and potential to advance predictive maintenance in the wind energy industry.

\textbf{Acknowledgements}: This work has been supported by the projects GRAVATAR PID2023-151184OB-I00 funded by MCIU/AEI/10.13039/501100011033 and ERDF, UE; and by GreenVAR of the Fundación Ramón Areces.

%% file: supplementary_arxiv.tex
\clearpage
% This command forces a new page and puts the argument across the full page width
\twocolumn[{%
   \null
   \vskip .375in
   \begin{center}
      % 1. The Title (Large and Bold)
      {\Large \bf Unsupervised Modular Adaptive Region Growing and RegionMix Classification for Wind Turbine Segmentation\\--Supplementary Material-- \par}

      % 3. Vertical space (matching WACV style)
      \vspace*{24pt}
      
      % 4. Author Block (Camera Ready Style)
      {
        \large
        \lineskip .5em
        \begin{tabular}[t]{c}
          Raül Pérez-Gonzalo$^{1,3,\dagger}$, \hspace{0.3cm}
          Riccardo Magro$^{1,2,\dagger}$, \hspace{0.3cm}
          Andreas Espersen$^{3}$, \hspace{0.3cm}
          Antonio Agudo$^{1}$
          \\[0.em] % Space between names and affiliations
          
          % Affiliations
          {\large $^{1}$Institut de Robòtica i Informàtica Industrial, CSIC-UPC, Barcelona, Spain} \\
          {\large $^{2}$Politecnico di Milano, Milan, Italy} \\
          {\large $^{3}$Wind Power LAB, Copenhagen, Denmark} \\
          {\tt\small \{rpg,ace\}@windpowerlab.com, riccardo.magro@mail.polimi.it, aagudo@iri.upc.edu}
        \end{tabular}
        \par
      }
      
      % 5. Final Spacing
      \vskip .5em
      \vspace*{12pt}
   \end{center}
}]

% 4. Reset your counters for the supplementary section
\setcounter{equation}{0}
\setcounter{figure}{0}
\setcounter{table}{0}
\setcounter{section}{0}

%\maketitle
\input{WACV26/chapters/Supplementary}

%% file: WACV26/chapters/Supplementary.tex
\section{Revisiting Region-Growing Segmentation}

Region-growing segmentation operates on the principle of pixel aggregation, where the grouping mechanism is based on homogeneity in a feature space that often includes intensity levels, color metrics, textural patterns, or spatial closeness. The algorithm commences with the selection of {\it seed} points, which serve as the starting locations for region growth. These seeds may be chosen based on a-priori knowledge of the image content, through automated processes that might involve statistical analyses, heuristic methods or in some cases can even be performed manually.

As the iterative process unfolds, the algorithm examines adjacent pixels or subregions and decides whether to merge them with the seed's growing region. This decision is contingent upon a similarity criterion, which vary substantially across proposed solutions. The criterion must be carefully defined to ensure that the resulting segmented regions are meaningful with respect to the image's content and the desired application.%~\cite{RevisitingRG}.

The generic outline of a region-growing algorithm is shown in \cref{Generic RG}. 
\begingroup % Start a group to localize changes
\renewcommand\thefootnote{\(\dagger\)} % Set the footnote mark to a dagger symbol
\footnotetext{These authors contributed equally.}
\endgroup % End the group, restoring the default footnote mark (usually numbers)

\begin{algorithm}[t!]
\caption{Generic Region-Growing Algorithm}
\label{Generic RG}
\begin{algorithmic}[1]
\STATE \textbf{Step 1: Seed Selection}
\STATE Choose an initial pixel to serve as the $seed$
\IF{interactive selection}
    \STATE User selects the seed pixel
\ELSE
    \STATE Seed pixel is chosen randomly or through some criteria
\ENDIF
\STATE \textbf{Step 2: Similarity Criterion for Pixels}
\STATE Define a criterion to determine similarity between two pixels
\STATE \textbf{Step 3: Initial Region Aggregation}
\STATE Add neighboring pixels similar to $seed$ to form an initial region 
\STATE \textbf{Step 4: Similarity Criterion for Regions}
\STATE Establish a criterion for similarity between a pixel and a region
\STATE \textbf{Step 5: Iteration}
\STATE Continue aggregation until no more pixels can be added
\end{algorithmic}
\end{algorithm}

\section{Dual-Threshold Seeded Region-Growing} \label{sec: DTRG} 

The Seed Selection strategy and the Dual-Threshold Region-Growing coalesce to form a segmentation algorithm, denoted by Dual-Threshold Seeded Region-Growing (DTRG) and summarized in \cref{DTRG}. 

\begin{algorithm}[t!]
\caption{Dual-Threshold Seeded Region-Growing}
\label{DTRG}
\begin{algorithmic}[1]
\STATE \textbf{Input:} Image $\mathbf{X}$
\STATE \textbf{Output:} Segmented regions \(\{\mathbf{R}_1, \mathbf{R}_2, \ldots, \mathbf{R}_N\}\)
 
\STATE Initialize $\mathbf{X}_{\mathbf{c}}$ as an empty set
\STATE Generate the set of candidate seed pixels \(C\)
\FORALL{\(\mathbf{c}_{h,w} \in C\)}
    \STATE Promote \(\mathbf{c}_{h,w}\) to seed pixel \(\mathbf{s}_{h,w}\) using Seed Promotion criteria 
    \STATE Initialize queue \(Q \leftarrow \{\mathbf{s}_{h,w}\}\)
    \STATE Initialize a new region \(\mathbf{R} \leftarrow \{\mathbf{s}_{h,w}\}\)
    \WHILE{\(Q\) is not empty}
        \STATE Dequeue \(\mathbf{x}_{h,w}\) from \(Q\)
        \FORALL{\(\mathbf{x}_{h',w'}  \in \aleph_{h,w}^{(1)}\)} 
            \STATE Calculate \(d_{l}\) and \(d_{s}\) for \(\mathbf{x}_{h',w'}\)
            \IF{\(d_{l} \leq \tau^{l}\) and \(d_{s} \leq \tau^{s}\)}
                \STATE Add \(\mathbf{x}_{h',w'}\) to \(R\) and enqueue \(\mathbf{x}_{h',w'}\) to \(Q\)
            \ENDIF
        \ENDFOR
    \ENDWHILE
    \STATE \(\mathbf{X}_{\mathbf{c}} \leftarrow \mathbf{X}_{\mathbf{c}} \cup \mathbf{R}\) 
    \STATE Mark \(\mathbf{R}\) as a completed region
\ENDFOR
\STATE \textbf{return} segmented regions \(\{\mathbf{R}_{1}, \mathbf{R}_{2}, \ldots, \mathbf{R}_{N}\}\)
\end{algorithmic}
\end{algorithm}

%\vspace{-0.7cm}
\section{Understanding Local and Seed Thresholds} \label{ThresholdVisuals}

This section elucidates the operational mechanics of local threshold (\(\tau^{l}\)) and seed threshold (\(\tau^{s}\)) within the presented Seeded Region-Growing algorithm, using visual aids to illustrate their impact on segmentation. To facilitate comprehension, a consistent color-coding scheme across all illustrations is employed:

\begin{itemize}
    \item \textbf{Seed Pixel}: Highlighted in red-violet, indicating the starting point for region growth.
    \item \textbf{Current Pixel}: Marked in yellow, representing the pixel currently under evaluation, surrounded by its 8-connected neighbors \(\aleph_{h,w}^{(1)}\).
    \item \textbf{Neighbors}: Shown in grey, these pixels are potential candidates for inclusion in the region.
    \item \textbf{Integrated Pixels}: Highlighted in orange, indicating pixels already assimilated into the region.
    \item \textbf{Color-Coded Circles}: Indicate the evaluation outcome based on local and seed thresholds: green for fulfillment of both criteria, red for non-fulfillment, blue for only seed threshold satisfaction, and purple for only local threshold satisfaction.

\end{itemize}

%This symbology, detailed in the legend (Fig.~\ref{fig:legend_up}), aids in visualizing the algorithm's decision-making process for pixel inclusion in a given region.

% \begin{figure}[t!]
%    \centering
%    \includegraphics{Plots/Legend.png}
%   \caption{\textbf{Illustration Key: Understanding Symbols and Colors}. This legend delineates the symbols and color codes used to visualize the segmentation algorithm's decision-making process, highlighting how local and seed thresholds influence pixel inclusion.}
%    \label{fig:legend_up}
%    \vspace{-0.4cm}
% \end{figure}

The following illustrations demonstrate how $\tau^{l}$ and $\tau^{s}$ activate, how their combined effect guides the region-growing process, and how the algorithm discerns which pixels to include in a region based on these criteria.

\begin{figure*}[t!]
    \centering
    \begin{tabular}{cccc}
        % First Image and Caption
        \begin{subfigure}[t]{0.235\textwidth}
            \centering
            \includegraphics[width=\linewidth]{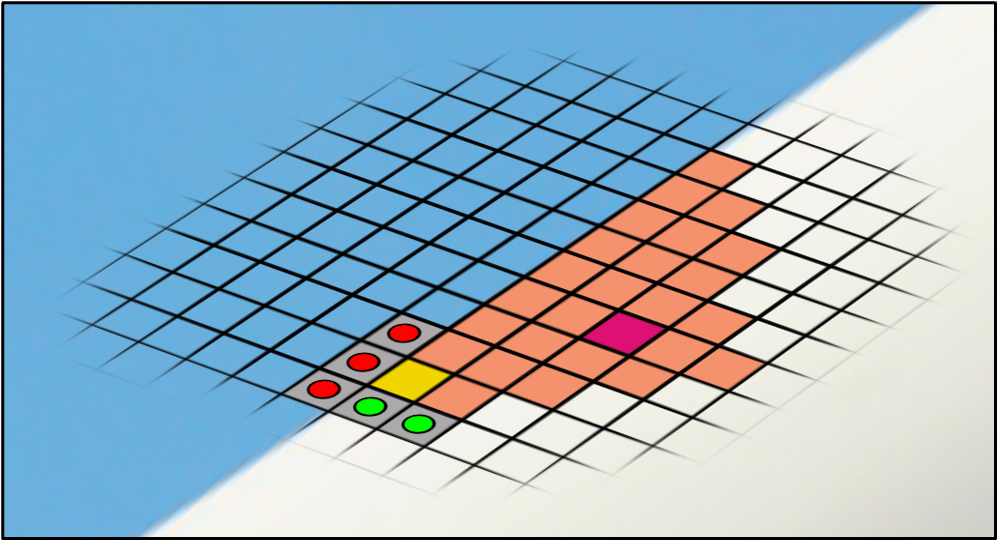}
            \caption{\textbf{Simultaneous activation of \(\tau^{l}\) and \(\tau^{s}\).} This scenario shows a seed pixel (red-violet) and the current pixel (yellow), surrounded by neighbors (grey) not yet included in the region. Neighbors satisfying both \(\tau^{l}\) and \(\tau^{s}\) appear with a green circle and red circle mark those failing the criteria.}
            \label{fig:scenario0}
        \end{subfigure} 
        &
        % Second Image and Caption
        \begin{subfigure}[t]{0.235\textwidth}
            \centering
            \includegraphics[width=\linewidth]{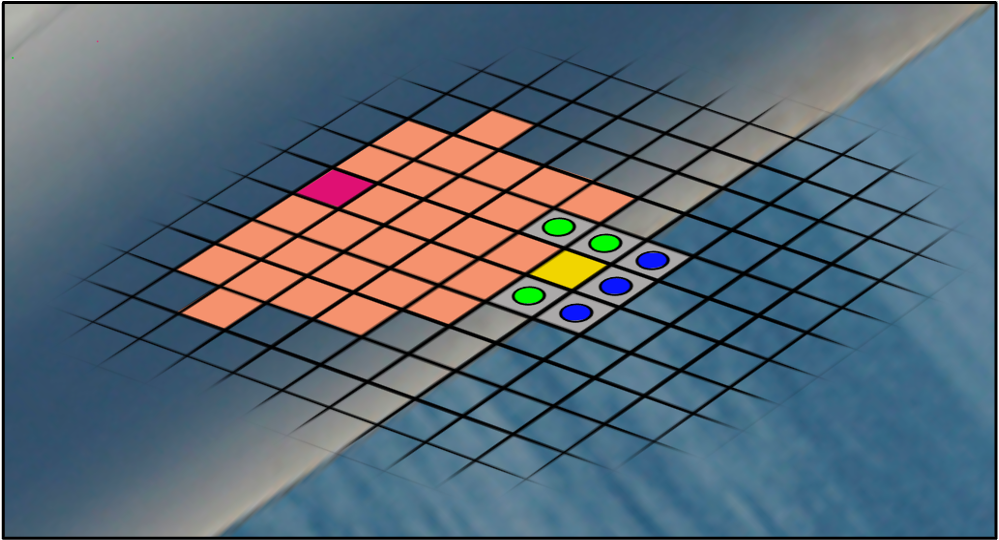}
            \caption{\textbf{Local threshold (\(\tau^{l}\)) dynamics.} This scenario shows a seed pixel (red-violet) and the current pixel (yellow), surrounded by neighbors (grey) not yet included in the region. Neighbors satisfying both \(\tau^{l}\) and \(\tau^{s}\) appear with a green circle. Those failing to satisfy the \(\tau^{l}\) appear with a blue circle.}
            \label{fig:scenario1}
        \end{subfigure} 
        &
        
        % Third Image and Caption
        \begin{subfigure}[t]{0.235\textwidth}
            \centering
            \includegraphics[width=\linewidth]{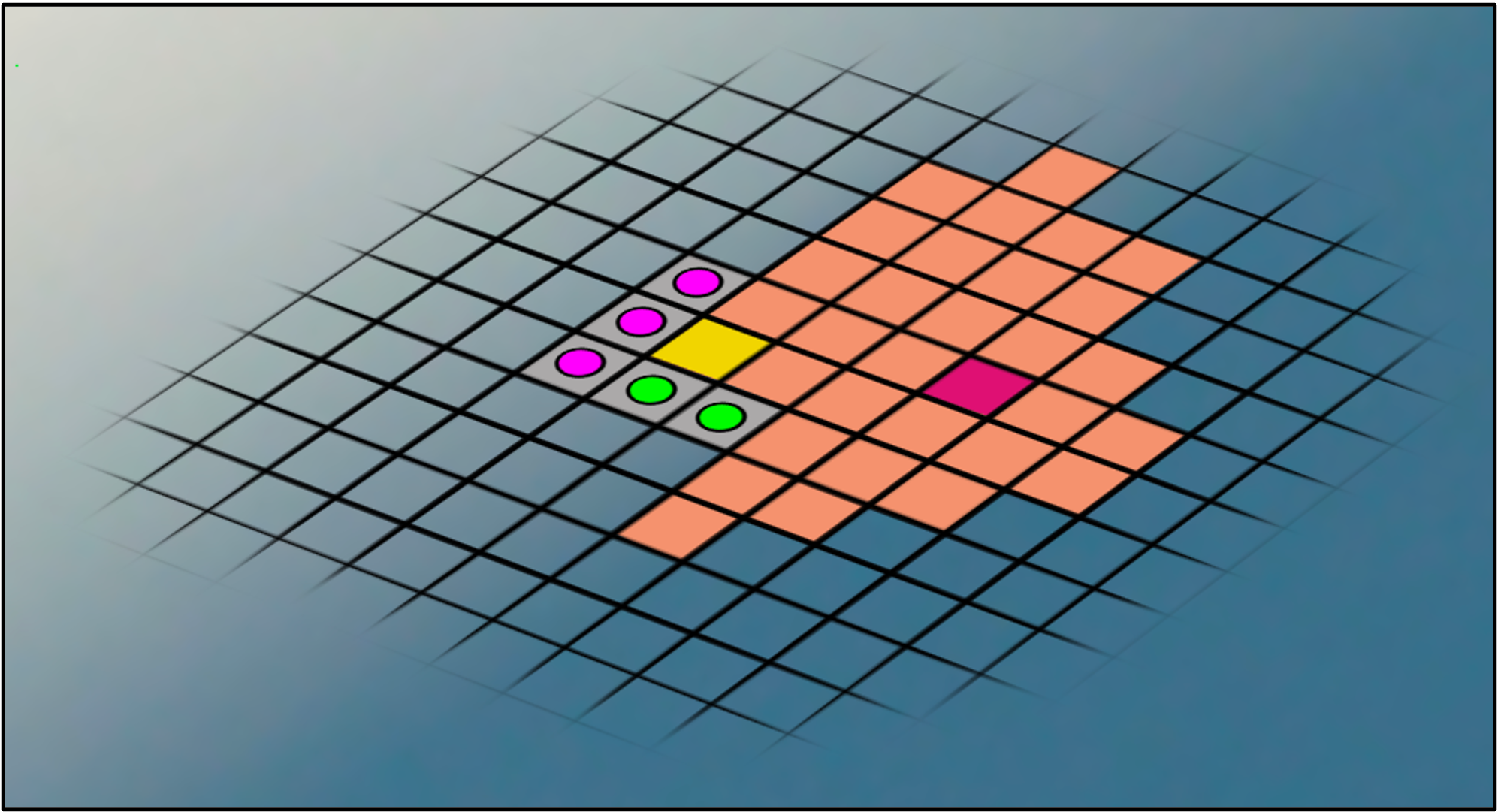}
            \caption{\textbf{Seed threshold (\(\tau^{s}\)) dynamics.} This scenario shows a seed pixel (red-violet) and the current pixel (yellow), surrounded by neighbors (grey) not yet included in the region. Neighbors satisfying both \(\tau^{l}\) and \(\tau^{s}\) appear with a green circle. Those failing to satisfy \(\tau^{s}\) appear with a purple circle.}
            \label{fig:scenario2}
        \end{subfigure} 
        &
        \begin{subfigure}[t]{0.2\textwidth}
            \centering
            \includegraphics[width=\linewidth]{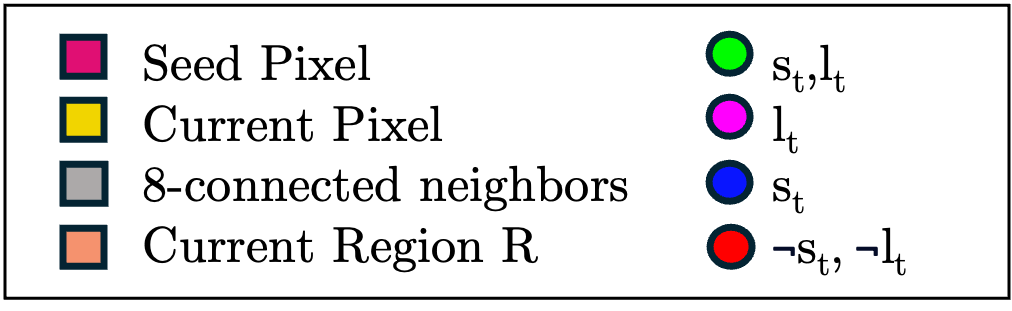}
            \caption{\textbf{Pixel color legend}.}
    %\vspace{5cm}
    \label{fig:legend}
        \end{subfigure} 
    \end{tabular}
    \vspace{-0.3cm}
    \caption{\textbf{Representative illustrations to understand the local \(\tau^{l}\) and seed \(\tau^{s}\) thresholds} in dual-threshold region-growing algorithm.} 
\end{figure*}

In \cref{fig:scenario0}, three neighboring pixels fail to be incorporated into the current region due to significant color differences with the seed and current pixels, thus violating both conditions ($d_{s} \leq \tau^{s}$, $d_{l} \leq \tau^{l}$). Conversely, two neighbors, congruent in color with the blade's RGB values, satisfy both thresholds and are assimilated into the region.

\Cref{fig:scenario1} depicts a situation where three neighbors meet the criteria for inclusion, exhibiting color similarities with both the seed and current pixels, thus fulfilling conditions ($d_{s} \leq \tau^{s}$, $ d_{l} \leq \tau^{l}$). The three excluded neighbors align with the seed pixel in terms of color; however, they diverge substantially from the current pixel, breaching the local threshold. Here, \(\tau^{l}\) adeptly delineates a boundary in an image with subtle transitions.

In \cref{fig:scenario2}, two neighbors are integrated into the region, whereas three are rejected due to their significant color variance from the seed pixel. This case exemplifies the seed threshold's utility in addressing the limitations of the local threshold, which may not detect gradual but significant color transitions, by providing a global reference for color consistency.

In conclusion it is possible to assess that $\tau^{l}$ primarily functions to maintain local color consistency during the region-growing process. For any pixel at the boundary of a growing region in fact, $\tau^{l}$ assesses the color difference between this pixel and its neighborhood. By imposing the condition $d_{l} < \tau^{l}$, we restrict the addition of neighboring pixels to the region only if their color intensity does not significantly deviate from that of the bordering pixel. This constraint effectively mitigates the erroneous expansion of a region across the edge of an object, a common issue in scenarios where object boundaries are not distinctly marked.

However, relying solely on $\tau^{l}$ is insufficient due to the presence of gradual color gradients within an image. Such gradients can lead to over-expansion of a region, as the local threshold criterion may continually be satisfied for adjacent pixels despite a considerable overall deviation from the original region color. Here it is where $\tau^{s}$ becomes crucial. It compares the color of a candidate pixel not just with its neighborhood but with the original seed one of the region. By enforcing the condition $d_{s} < \tau^{s}$, it is ensured that the pixels added to a region are not only locally consistent but also globally representative of the seed pixel's characteristics. In other words, $\tau^{s}$ solves cases where iteratively each pixel added the deviation from the seed pixel becomes higher and higher without any limits.

\begin{algorithm}[t!]
\caption{Adaptive Threshold for Seeded Region-Growing Segmentation}
\label{AT}
\begin{algorithmic}
\STATE \textbf{Input:} Image $\mathbf{X}$
\STATE \textbf{Output:} Adaptive $\tau^{s}$, $\tau^{l}$

\STATE Initialize $\tau^{s}$ to a preliminary value.
\REPEAT
    \STATE Increment $\tau^{s}$.
    \STATE Compute $\Pi$.
\UNTIL{There is no further increase in $\Pi$.}
\STATE Establish ${\tau^{s*}} = \tau^{s}$.
\REPEAT
    \STATE Increment $\tau^{l}$.
    \STATE Compute $\Pi$.
\UNTIL{There is no further increase in $\Pi$.}
\STATE Update $\tau^{l}$ to its last value: ${\tau^{l}} = {\tau^{l*}}$ 
\STATE \textbf{return} ${\tau^{s*}}$, ${\tau^{l*}}$
\end{algorithmic}
\end{algorithm}

\section{Adaptive Thresholding}

\begin{figure}[t!]
  \centering
 \includegraphics[width=\linewidth]{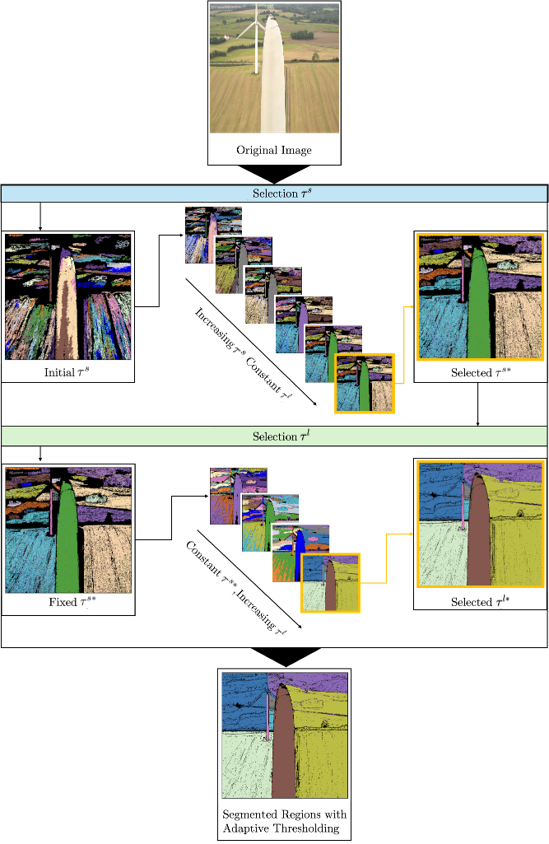}
 \vspace{-0.6cm}
  \caption{\textbf{Visualization of the Adaptive Algorithm:} The process starts from an image and returns the adaptive ${\tau^{s*}}$ and ${\tau^{l*}}$ using $\Pi$ as an unsupervised metric for gauging segmentation quality.}
\label{fig:full_adapt} \vspace{1cm}
\end{figure}

Adaptive Thresholding (AT) algorithm seeks the pair of thresholds based on local image characteristics, fundamental to address a highly diversified dataset. This unsupervised algorithm that adaptively determines $\tau^{s}$ and $\tau^{l}$ based on a dual-thresholding criterion is presented in \cref{AT}. A qualitative example of AT algorithm is illustrated in \cref{fig:full_adapt}.

\section{Region Merging}

\begin{figure}[t!]
  \centering
  \includegraphics[width= \linewidth]{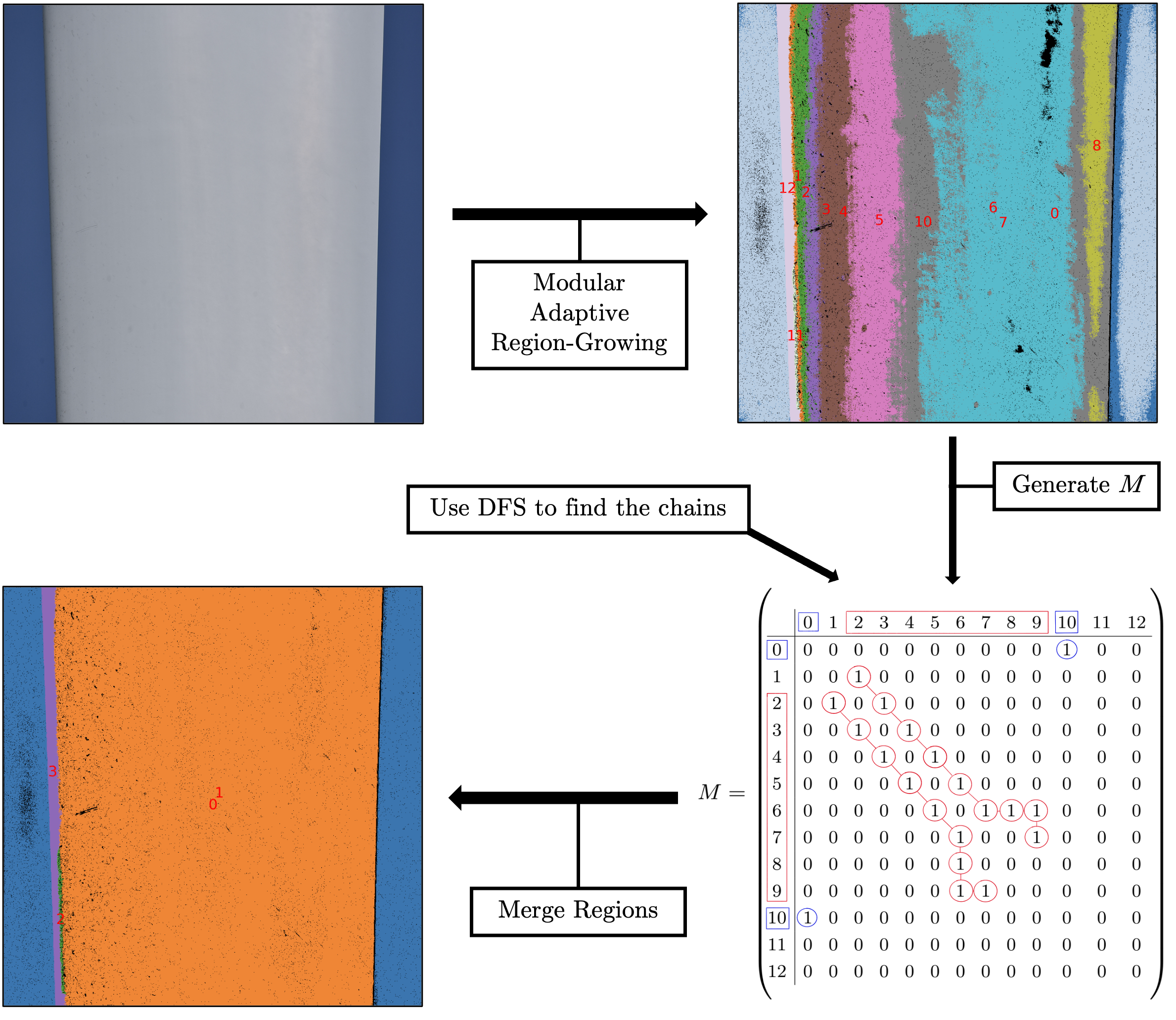}
  \vspace{-0.5cm}
  \caption{\textbf{Comprehensive Overview of the Region Merging Strategy.} {\textbf{Top-right:} Initial segmented regions obtained using DTMRG. \textbf{Bottom-right:} The merging matrix $\mathbf{M}$, where non-zero entries suggest potential merging of corresponding regions. \textbf{Bottom-left:} Final image after the Region-Merging algorithm is applied, illustrating the aggregation of initial segments into larger homogeneous areas. In this figure, the segmented regions are numbered (in red) to highlight the reduction of regions from before to after merging process.}}
  \label{fig:full_RM}
\end{figure}

The Region Merging process effectively consolidates overlapping regions into cohesive segments, greatly reducing the overall number of regions (see \cref{fig:full_RM}). This approach not only streamlines the classification task but also produces segmentation results that more accurately reflect the image's actual structure, minimizing fragmentation and improving interpretability.

\subsection{Full Region-Growing Algorithm Outline}

All the previous ideas are now combined to propose the comprehensive Modular Adaptive Region-Growing (MARG) algorithm summarized in \cref{full_algo}. This method can directly split the image $\mathbf{x}$ into different regions \(\{\mathbf{R}_1, \mathbf{R}_2, ..., \mathbf{R}_N\}\) in an unsupervised fashion, without assuming any training data at all. 

\begin{algorithm}[t!]
\caption{Modular Adaptive Seeded Region-Growing Algorithm (MARG)}
\label{full_algo}
\begin{algorithmic}[1]
\STATE \textbf{Input:} Image $\mathbf{X}$
\STATE \textbf{Output:} Segmented regions \(\{\mathbf{R}_1, \mathbf{R}_2, \ldots, \mathbf{R}_N\}\)

\STATE \textbf{Step 1: Adaptive Thresholding }
\STATE Determine ${\tau^{s*}}$ and ${\tau^{l*}}$

\STATE \textbf{Step 2: Dual-Threshold Modular Region-Growing }
\STATE Use ${\tau^{s*}}$ and ${\tau^{l*}}$, perform DTMRG on $\mathbf{X}$

\STATE \textbf{Step 3: Region Merging }
\end{algorithmic}
\end{algorithm}

\section{Hyperparameter Sensitivity}

\begin{figure}[t!]
  \centering
  \begin{tabular}{@{}c@{}}
  \includegraphics[width=0.8\linewidth]{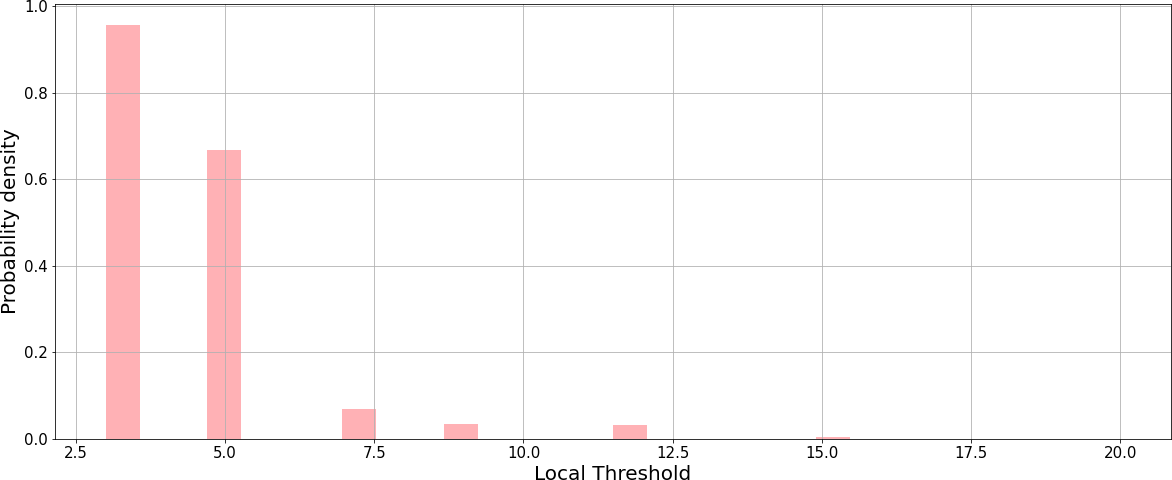}     \\
  \includegraphics[width=0.8\linewidth]{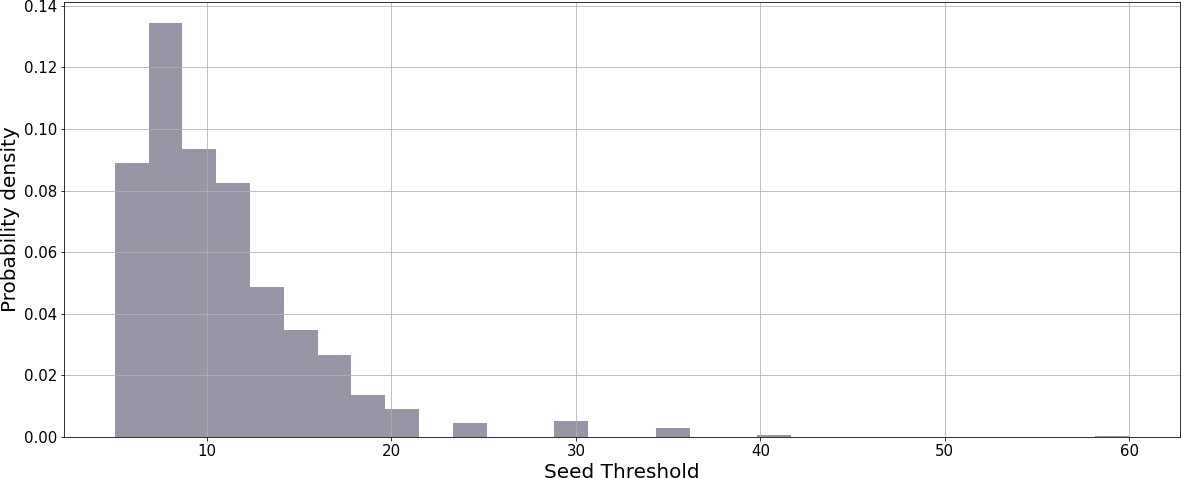}
  \end{tabular}  
  \vspace{-0.3cm}
   \caption{\textbf{Optimal local $\tau^{l*}$ and seed $\tau^{s*}$ threshold distributions over the test dataset.}} 
   \label{fig:thresh_dist}
   \vspace{-0.4cm}
\end{figure}

In \cref{fig:thresh_dist}, we depict the distributions in optimal local and seed thresholds for AT. The high variability in optimal thresholds demonstrates the method's flexibility in adapting to diverse image characteristics, improving region segmentation in challenging conditions.
 
Moreover, we analyzed the sensitivity of the MARG algorithm to its hyperparameters by assessing how variations in these parameters affect segmentation performance across the test dataset. Specifically, we examined the range within which a hyperparameter could be adjusted without reducing the mIoU and F1-score by more than 1\%. The window size $k$ defined in $\aleph_{h,w}^{(k=2)}$ from Seed Selection strategy (Section~3.2 of the main manuscript) can be flexibly modified to $k \in \{2,\ldots,6\}$. Similarly, the merging overlap from Equation~5 of the main manuscript achieves consistent results within the range $[0.08,0.32]$. 

\section{Failure Analysis} \label{sec:marg-failure}

\begin{figure}[t!]%[h!]
\centering
\hspace{-0.1cm}\begin{tabular}{@{}c@{}@{}c@{}}
\includegraphics[width=0.5\linewidth]{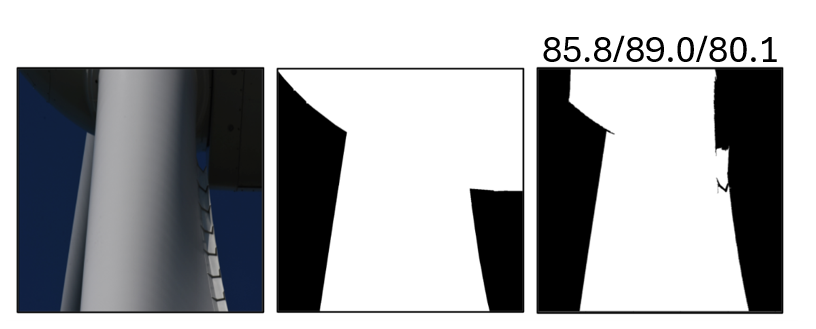}&
\includegraphics[width=0.5\linewidth]{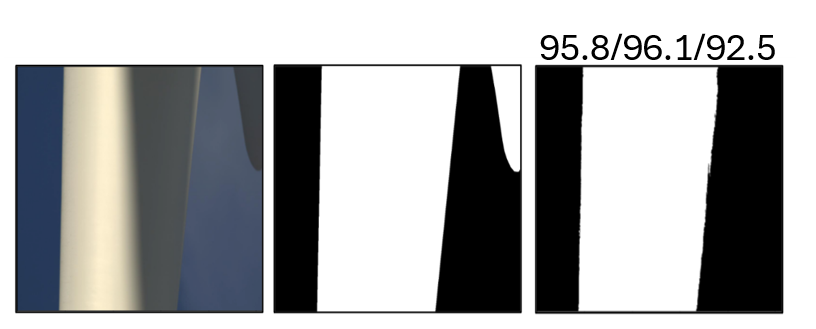} \vspace{-0.5cm}\\

\includegraphics[width=0.5\linewidth]{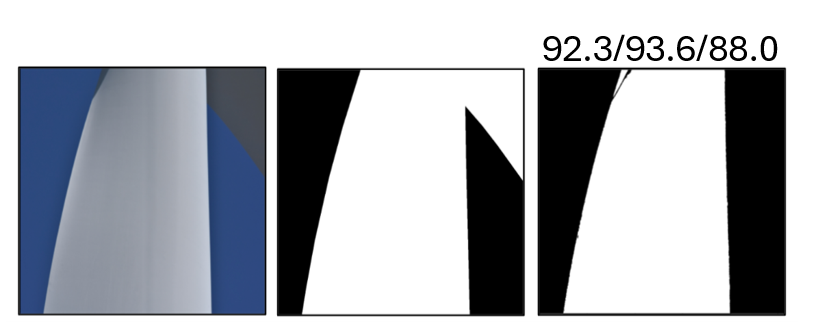}&
\includegraphics[width=0.5\linewidth]{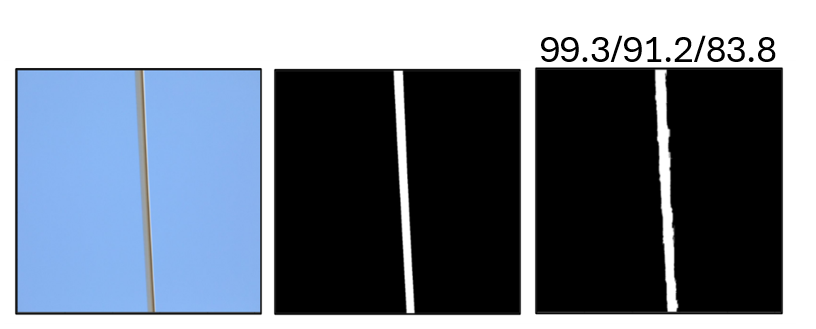}\vspace{-0.5cm}\\

\includegraphics[width=0.5\linewidth]{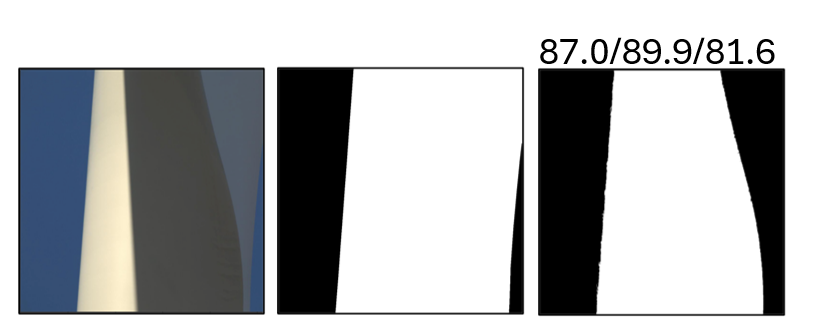}&
\includegraphics[width=0.5\linewidth]{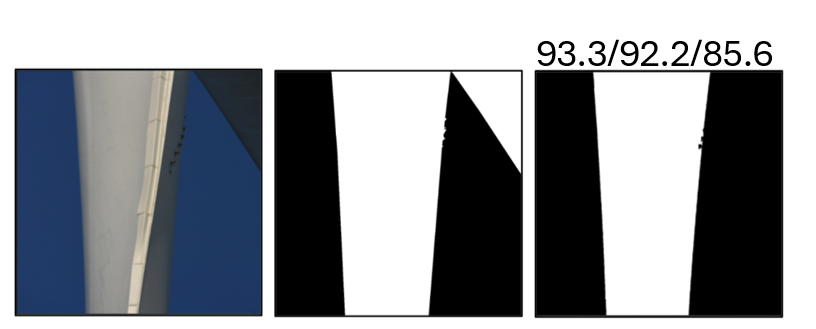}

\end{tabular}
\vspace{-0.5cm}
\caption{\textbf{Typical failure cases on test images of MARG+RegionMix classifier.} The first column presents the input color image, the second displays the ground-truth mask, and the third shows the segmentation estimation. On top of each estimation, the accuracy, F1 and mIoU performance metrics are showcased, respectively. Both sides display the same information.} \label{fig:marg_failure}
\end{figure}

\begin{table}[t!]
\centering
\resizebox{\linewidth}{!} {
\begin{tabular}{l cccc}
\toprule
Failure mode & Percentage of & Accuracy & F1 & mIoU \\
& outliers {[\%]} & {[\%]} & {[\%]} & {[\%]} \\
\midrule
Blade root & 43.8 & 87.89 & 85.46 & 82.86 \\
Multiple blades & 42.1 & 91.45 & 88.72 & 85.13\\
Slender blades & ~7.5 & 98.76 & 92.95 & 86.32 \\
Atypical lighting & ~6.6 & 82.56 & 82.34 & 72.64 \\
\bottomrule
\end{tabular}}
\vspace{-0.3cm}
\caption{\textbf{Distribution of failure modes among outlier cases}. These outliers are taken from the boxplot, refer to Fig.~10 of the main manuscript.}
\label{tab:failure_modes}
\end{table}

In \cref{fig:marg_failure}, we illustrate some representative instances where our MARG+RegionMix classifier may struggle. This analysis aims to facilitate future research to build on top of our algorithm and improve wind turbine blade segmentation. 
When analyzing the typical failure cases of our MARG classifier, we observe that the algorithm typically struggles with similar cases as other image segmentation algorithms - despite adopting a total different approach with region classification. These failure cases can be categorized into four main groups: images of the blade's root, images with the presence of multiple blades, slender blade images, and those with atypical lighting conditions.

To provide an in-depth analysis, we studied the outliers present in the windfarm boxplots from Fig.~10 of the main manuscript to analyze robustness. As seen in \cref{tab:failure_modes}, around 43\% of the outliers depict the blade’s root. The model struggles to correctly segment the hub (first row, left side instance), often misclassifying it as background due to the lack of hub images in the training data. These cases correspond to a reduction in mIoU from the overall average to 82.86\%. Another 43\% of outliers feature multiple blades within the frame (first row, right side; second row, left side instances), leading the model to omit additional blades and classify them as background, as the training set primarily includes single-blade images. This failure mode results in a mIoU of 85.13\%. A smaller fraction (~7.5\%) of outliers correspond to slender blade images (second row, right side instance), where the regions tend to slightly overestimate the area of slender blades, which visually does not seem alarming, but end up dropping significantly the mIoU performance to 86.32\%. Lastly, around 6.6\% of outliers arise from unusual lighting conditions (third row instances), such as shadows, which can mislead the model into segmenting a single blade into multiple parts. These scenarios show the most severe degradation, with IoU dropping to 72.6\%, and occur primarily because such illumination cases are rare in both training and test sets, reducing the classifier’s ability to generalize. 

These failures are primarily attributable to dataset limitations and the region classifier, rather than the proposed MARG region-growing algorithm itself. In particular, insufficient training samples for hub and multi-blade configurations limit classifier generalization. Collecting more diverse training data (including hub and multi-blade instances), augmenting with varied lighting conditions, and refining boundary-aware post-processing would mitigate most errors. Furthermore, when compared with competitive segmentation models (\cref{fig:practical-impact}), we find that these methods exhibit these weaknesses in blade root and shadowed cases, suggesting that such difficulties are intrinsic to the task rather than specific to our approach.

\begin{figure}[t!]
\centering
\includegraphics[width=0.9\linewidth]{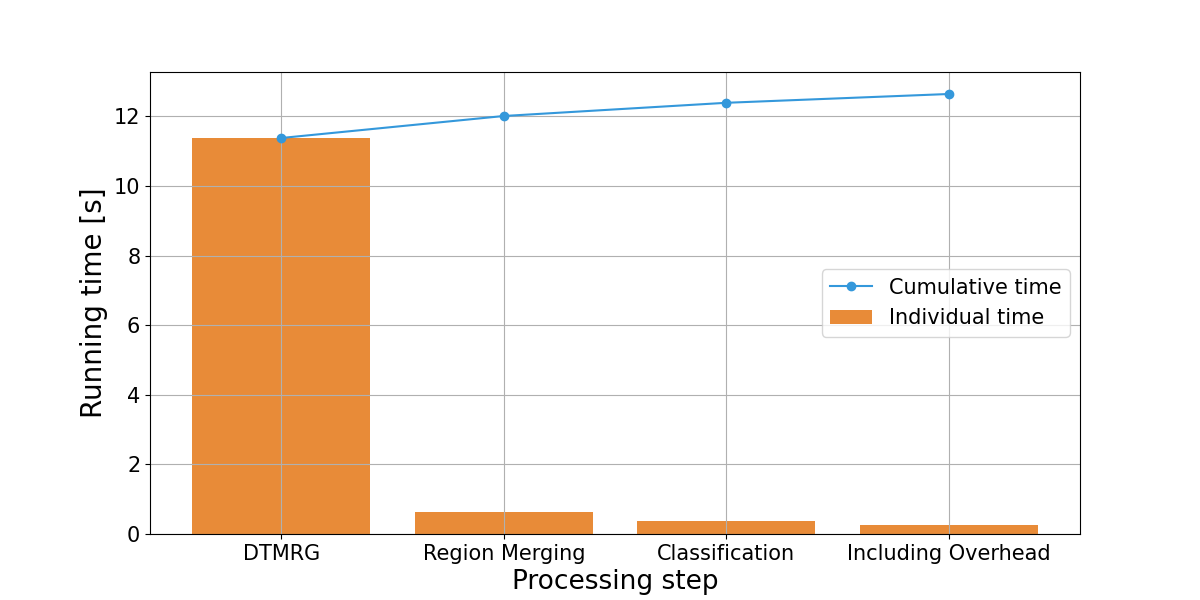}
    \vspace{-0.3cm}
    \caption{\textbf{Computational time of a single image for each step in MARG + Classifier segmentation algorithm}. Computational time for MARG, Region Merging and classification steps are reported. The blue line represents the cumulative time.}
    \label{fig:marg-time}
    \vspace{-0.4cm}
\end{figure}

\begin{table}[t!]
    \centering
    \resizebox{0.95\linewidth}{!} {
    \begin{tabular}{lrr}
    \toprule
     \multicolumn{1}{c}{Method} & \multicolumn{1}{c}{GPU Inference} & \multicolumn{1}{c}{CPU Inference}\\
     \multicolumn{1}{c}{} & {Time [s] $\downarrow$} & {Time [s] $\downarrow$}\\
    \midrule
    BiRefNet~\cite{birefnet} & 4.93 & 13.56 \\
    SAM~\cite{sam} & 6.40 & 69.42 \\ 
    CLIPSeg~\cite{clipseg} & 2.39 & 3.57 \\ 
    DiffSeg~\cite{diffseg} & 26.07 & 79.21 \\ 
    EfficientFormer~\cite{efficientformer} & 4.51 & 7.32 \\ 
    MobileViT~\cite{mobilevit} &  6.18 & 13.19 \\  %inventao
    BU-Net~\cite{bunet} & 4.89 & 10.42 \\ %inventao
    Mask2Former-FreqFusion~\cite{freqfusion} & 5.11 & 42.47 \\
    MARG+RegionMix & - & 12.23 \\ 
   
    \bottomrule
    \end{tabular}
    }
    \vspace{-0.3cm}
    \caption{\textbf{Computational time comparison on wind turbine blade segmentation} of the distinct state-of-the-art methods.} 
    \label{tab:silora-sota-time}
    \vspace{-0.4cm}
\end{table}

\begin{table*}[t!]%[h!]
\centering
\resizebox{\linewidth}{!} {
\begin{tabular}{c:ccc|ccc|ccc}
\toprule
 & \multicolumn{3}{c|}{\textbf{Accuracy}} & \multicolumn{3}{c|}{\textbf{mIoU}} & \multicolumn{3}{c}{\textbf{F1}} \\
Windfarm &          Mean ± Std &   CI (t$_{\text{student}}$) & CI (boot) &          Mean ± Std & CI (t$_{\text{student}}$) & CI (boot) &       Mean ± Std &   CI (t$_{\text{student}}$) & CI (boot) \\
\midrule
     WF1 & 98.19 ± 5.20 & 95.76–100.63 &   95.81–99.44 &  93.44 ± 8.30 & 89.55–97.33 &    89.43–96.64 & 96.41 ± 4.83 & 94.15–98.67 &  94.07–98.25 \\
     WF2 & 99.45 ± 0.24 &  99.34–99.56 &   99.34–99.55 &  95.82 ± 4.95 & 93.51–98.14 &    93.49–97.75 & 97.83 ± 2.70 & 96.57–99.10 &  96.56–98.89 \\
     WF3 & 97.53 ± 6.37 & 94.46–100.60 &   94.23–99.44 &  94.74 ± 7.78 & 90.98–98.49 &    90.75–97.31 & 97.12 ± 4.71 & 94.85–99.39 &  94.72–98.62 \\
     WF4 & 96.28 ± 8.56 & 91.88–100.68 &   91.75–99.17 & 91.43 ± 11.22 & 85.66–97.20 &    85.56–95.53 & 95.12 ± 7.37 & 91.33–98.91 &  91.25–97.69 \\
     WF5 & 97.68 ± 6.08 & 94.84–100.52 &   94.62–99.42 &  95.53 ± 6.98 & 92.26–98.80 &    92.07–98.03 & 97.58 ± 4.04 & 95.69–99.47 &  95.55–98.99 \\
     WF6 & 99.45 ± 0.32 &  99.30–99.61 &   99.31–99.58 &  97.82 ± 2.51 & 96.62–99.03 &    96.59–98.77 & 98.94 ± 1.31 & 98.31–99.57 &  98.28–99.41 \\
     WF7 & 99.25 ± 1.34 &  98.62–99.87 &   98.59–99.67 &  97.41 ± 4.28 & 95.41–99.42 &    95.36–98.79 & 98.64 ± 2.37 & 97.53–99.75 &  97.50–99.39 \\
     WF8 & 98.10 ± 3.09 &  96.66–99.55 &   96.59–99.10 &  96.65 ± 4.29 & 94.64–98.65 &    94.51–98.11 & 98.24 ± 2.38 & 97.13–99.36 &  97.06–99.04 \\
     WF9 & 98.53 ± 1.87 &  97.65–99.41 &   97.70–99.25 &  96.84 ± 2.56 & 95.64–98.04 &    95.74–97.88 & 98.96 ± 1.07 & 98.46–99.46 &  98.47–99.36 \\
    WF10 & 95.43 ± 8.85 &  91.29–99.57 &   91.08–98.67 & 93.26 ± 11.35 & 87.95–98.57 &    87.79–97.59 & 96.12 ± 6.94 & 92.87–99.37 &  92.76–98.73 \\
\bottomrule
\end{tabular}
}
\vspace{-0.35cm}
\caption{\textbf{Quantitative generalization results across test windfarms}. We report mean ± standard deviation, together with 95\% confidence intervals computed using both Student-$t$ and bootstrap resampling. Boxplot results are shown in Fig.~10 of the main manuscript. }
\label{tab:windfarms}
\vspace{-0.4cm}
\end{table*}

\section{Computational Cost} \label{sec:marg-time}

Similar to previous segmentation algorithms, we analyze the computational cost of the inference process using a system equipped with an NVIDIA RTX 3080 Ti GPU and a 20-core Intel Core i9-10900 processor. To accelerate the selection of optimal local thresholds $\tau^{l*}$ and seed thresholds $\tau^{s*}$ for AT, the coverage $\Pi$ of each seed threshold $\tau^s$ and of each local threshold $\tau^l$ (given a fixed $\tau^{s*}$) are computed in parallel. %The reported results assume that the model weights are pre-loaded.

\Cref{fig:marg-time} shows the average runtime for each step of the algorithm, including the cumulative time. The total runtime for segmenting a single image is 12.23 seconds, with the most time-consuming step being the DTMRG region-growing algorithm, which takes 11.10 seconds. This is primarily due to the computation of the optimal thresholds in the AT process. Despite parallelization, generating distinct region-growing images remains the primary bottleneck in inference speed.

The remaining steps of the algorithm are significantly faster: Region Merging takes 0.61 seconds, while the classification of each region takes just 0.31 seconds. There is also a small overhead of 0.21 seconds, which includes tasks such as image loading and saving, memory transfers between the CPU and GPU, region assembly, and hole filling.

Our algorithm is desined to run on the CPU, except for the region classification, which can be run with GPU acceleration. We include in \cref{tab:silora-sota-time} the computational time comparison for segmenting a single image for distinct state-of-the-art algorithms. While it does not achieve the highest speed in CPU timing, our framework provides a highly practical solution when GPU hardware is not available, as it accomplishes top-performing segmentation results and reaches the third highest timing on this hardware compared to state-of-the-art segmentation algorithms. In addition to this, it offers advantages in terms of interpretability, adaptability, and, as mentioned, ease of deployment in environments where GPU resources are limited or unavailable.

\section{Generalization Across Windfarms}

\Cref{tab:windfarms} reports performance on each windfarm in terms of mean $\pm$ standard deviation, together with 95\% confidence intervals estimated using both a Student-$t$ model and a non-parametric bootstrap. This results are also illustrated in the main manuscript using boxplot distributions in Fig.~10. Results show consistently strong generalization across sites, with accuracies above 95\% for all windfarms. The mIoU and F1 scores follow the same trend, with only modest variability depending on windfarm-specific conditions. The close agreement between $t$-based and bootstrap intervals indicates that the reported means are robust and not strongly affected by distributional assumptions. Notably, performance remains high even on windfarms with greater variability (e.g., WF4 and WF10), highlighting the method’s resilience to domain shifts. This demonstrates that our approach generalizes reliably to previously unseen windfarms, addressing a key challenge for deployment in real-world scenarios.

\section{Region Classifier Architectures} \label{sec:region-classifiers}

Our MARG framework first employs our proposed region-growing algorithm (MARG) to generate candidate regions, which are then classified into either blade or background using a CNN- or transformer-based binary classifier; see Fig.~1 and Fig.~7 of the main manuscript. In this section, we compare a range of candidate backbones for the region classifier, summarized in \cref{tab:architectures}.

We experimented with convolutional models (VGG~\cite{vggmodel}, ResNet~\cite{resnet}, EfficientNet~\cite{efficientnet}) and transformer-based classifiers (ViT~\cite{vit_b_16}, Swin~\cite{swin}, DeiT~\cite{deit}), trained either from scratch or with official pretrained weights. To adapt these architectures to our task, we extend the input layer with an additional fourth channel that encodes the binary region mask, while leaving the remainder of each model unchanged. The results in \cref{tab:architectures} show that most architectures achieve high accuracy on the binary classification task ($>$97\%), yet some differences emerge when balancing validation robustness and efficiency.

Overall, transformer models such as ViT-B/16~\cite{vit_b_16} and DeiT-S~\cite{deit} reach the highest test accuracy (98.33\% and 98.29\%, respectively) and F1-score (98.03\% and 97.88\%). However, they require considerably more parameters and training resources, and their validation accuracy fluctuates more strongly, suggesting potential overfitting to the test set if chosen solely on test performance. By contrast, EfficientNet-B4~\cite{efficientnet} with pretrained weights provides a strong trade-off: it achieves the best validation accuracy (98.38\%) and lowest validation loss (0.1972), while still reaching competitive test metrics (98.25\% accuracy, 97.79\% F1, 95.33\% mIoU). We therefore selecteFailure Analysisd EfficientNet-B4 as the backbone for our MARG+RegionMix classifier. This choice ensures that model selection is based on validation performance rather than test set tuning, thereby avoiding overfitting, and at the same time yields an efficient architecture in terms of computational cost.

We also explored whether other CNN- and transformer-based classifiers, such as ResNet-50, ResNet-101~\cite{resnet}, ViT-B/16~\cite{vit}, Swin-S~\cite{swin}, and DeiT-S~\cite{deit}, would substantially improve binary classification. As seen in \cref{tab:architectures}, although some of these models slightly outperform EfficientNet-B4~\cite{efficientnet} on isolated test metrics, they do not consistently improve validation performance. Moreover, transformer-based models incur higher memory and runtime costs without a clear margin of improvement, which opposes the idea of our algorithm to be deployed on CPU. This suggests that, for the binary classification step in MARG, EfficientNet-B4 offers the best balance between accuracy, generalization, and efficiency.

We observed that the use of pretrained weights generally improves both convergence stability and validation performance for CNN-based models. For instance, EfficientNet-B4 and EfficientNet-B0~\cite{efficientnet} benefit significantly from initialization with pretrained weights, reaching higher validation accuracy and lower loss compared to training from scratch. In contrast, transformer-based models such as ViT-B/16~\cite{vit} and Swin-S~\cite{swin} are less consistent: while ViT~\cite{vit} improves with pretraining in terms of accuracy, Swin-S~\cite{swin} shows reduced performance, likely due to domain mismatch between ImageNet pretraining and our region mask–augmented input. Specifically, Swin-S is the only model whose test performance is significantly lower than its validation accuracy, indicating overfitting caused by the combination of a large model capacity and the relatively small size of our dataset. This highlights that while pretraining is advantageous for convolutional backbones, its benefit for transformer-based classifiers is less clear in this task.

\begin{table}[t]
\centering
\resizebox{\linewidth}{!}{
\begin{tabular}{ll:ccccc|cc}
\toprule
& & \multicolumn{5}{c}{\textbf{Test}} & \multicolumn{2}{|c}{\textbf{Validation}} \\
Architecture & Pretr. & Acc. & Prec. & Recall & F1 & mIoU & Loss & Acc. \\
& & {[\%]} & {[\%]} & {[\%]}  & {[\%]}  & {[\%]} & & {[\%]} \\
\midrule
VGG16~\cite{vggmodel} & No                 & 98.13 & 98.23 & 96.74 & 97.78 & 95.28 & 0.3487 & 95.25 \\
ResNet34~\cite{resnet} & No               & 97.70 & 97.91 & 96.46 & 97.25 & 94.69 & 0.2525 & 97.93 \\
ResNet34~\cite{resnet} & Yes              & 98.01 & \underline{98.49} & 96.29 & 97.64 & 95.10 & 0.2669 & 97.79 \\
ResNet50~\cite{resnet} & No               & 97.84 & 97.95 & 96.62 & 97.42 & 94.89 & \underline{0.2273} & 96.97 \\
ResNet50~\cite{resnet} & Yes              & 98.01 & 98.10 & 96.70 & 97.52 & 95.05 & 0.2522 & \underline{97.97} \\
ResNet101~\cite{resnet} & No              & 97.40 & 96.61 & \underline{97.32} & 96.84 & 94.31 & 0.2475 & 97.03 \\
ResNet101~\cite{resnet} & Yes             & 97.52 & 97.74 & 96.47 & 97.12 & 94.52 & 0.2452 & 97.03 \\
EfficientNet-B0~\cite{efficientnet} & No        & 97.67 & 98.22 & 96.12 & 97.31 & 94.66 & 0.3541 & 96.79 \\
EfficientNet-B0~\cite{efficientnet} & Yes       & 97.94 & 98.45 & 96.15 & 97.48 & 94.98 & 0.3007 & \underline{97.98} \\
EfficientNet-B4~\cite{efficientnet} & No        & 97.88 & 97.46 & 97.14 & 97.48 & 94.97 & \underline{0.2419} & 95.84 \\
EfficientNet-B4~\cite{efficientnet} & Yes       & \underline{98.25} & 98.43 & 96.49 & \underline{97.79} & \underline{95.33} & \textbf{0.1972} & \textbf{98.38} \\
ViT-B/16~\cite{vit_b_16} & No  & 97.32 & 98.28 & 95.41 & 96.72 & 94.06 & 0.3361 & 91.69 \\
ViT-B/16~\cite{vit_b_16} & Yes & \textbf{98.33} & \textbf{98.67} & 96.57 & \textbf{98.03} & \textbf{95.49} & 0.3442 & 95.33 \\
Swin-S~\cite{swin} & No  & 92.75 & 89.20 & \underline{97.53} & 90.69 & 87.42 & 0.2420 & 97.32 \\
Swin-S~\cite{swin} & Yes & 88.74 & 85.06 & \textbf{97.59} & 88.84 & 83.14 & 0.3685 & 97.71 \\
DeiT-S~\cite{deit} & No   & \underline{98.29} & \underline{98.63} & 96.40 & \underline{97.88} & \underline{95.29} & 0.3055 & 95.86 \\
DeiT-S~\cite{deit} & Yes  & 98.14 & 98.09 & 96.94 & 97.75 & 95.28 & 0.2612 & 96.49 \\
\bottomrule
\end{tabular}}
\vspace{-0.35cm}
\caption{\textbf{Quantitative comparison of architectures for our region classifier} with and without public pretrained weights, which is denoted in the Table as ``Petr.''. We highlight the top-performing metric in bold and underline the second and third best models.}
\label{tab:architectures}
\vspace{-0.4cm}
\end{table}

\section{In-depth Comparative Comparison}

A priori limitation of our method is that the improvements over strong state-of-the-art models, while consistent, are numerically modest when averaged over the test set. All competing algorithms already exceed 90\% in standard metrics, and although ours is the top performer (see Table~3 of the main manuscript), it is only the third fastest on a CPU-only environment (\cref{tab:silora-sota-time}). It is therefore important to assess whether this tradeoff is justified, especially since segmentation serves as a precursor for downstream defect analysis.

\begin{figure}[t!]
\centering
\begin{subfigure}{\linewidth}
\centering
\resizebox{\linewidth}{!}{
\begin{tabular}{c:cccccccc}
\hline
& \multicolumn{8}{c}{\textbf{mIoU Threshold}} \\
Model & 0.2 & 0.3 & 0.4 & 0.5 & 0.6 & 0.7 & 0.8 & 0.9 \\
\hline
BiRefNet~\cite{birefnet}             & 2   & 5   & 6   & 11  & 14  & 20  & 24  & 31 \\
Mask2Former-FreqFusion~\cite{freqfusion}   & 0   & 1   & 2   & 7   & 8   & 14  & 18  & 24 \\
MARG+RegionMix                        & 0   & 0   & 1   & 2   & 5   & 6   & 13  & 23   \\
\hline
\end{tabular}}
\subcaption{mIoU below the given threshold.}
\label{tab:miou-thresholds}
\end{subfigure}
\hfill
\vspace{0.cm}
\begin{subfigure}{\linewidth}
\centering
\vspace{-0.3cm}
\includegraphics[width=\linewidth]{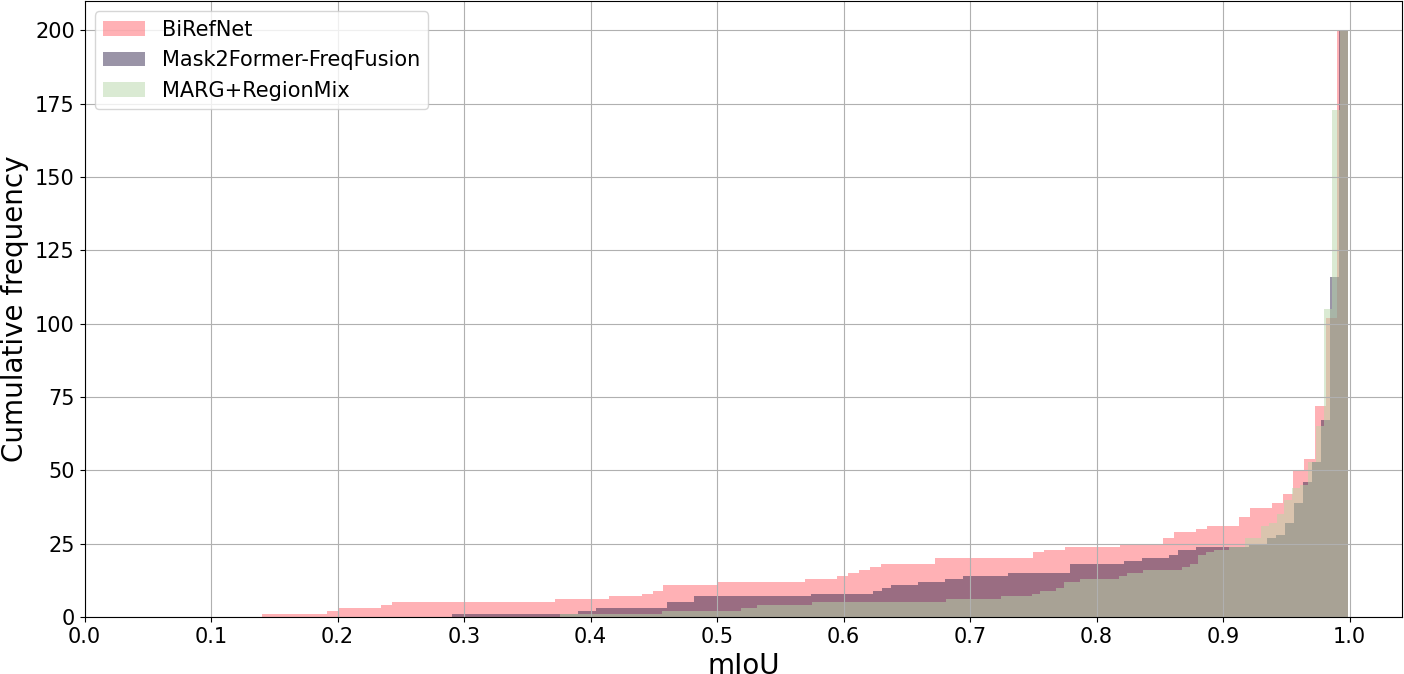}
\subcaption{Cumulative histogram of mIoU.}
\label{fig:miou-cumhist}
\end{subfigure}
\vspace{-0.65cm}
\caption{\textbf{Distribution of mIoU across the test set} (200 images) for BiRefNet~\cite{birefnet}, Mask2Former-FreqFusion~\cite{freqfusion}, and MARG+RegionMix.}
\label{fig:miou-distribution}
\end{figure}

\cref{fig:miou-distribution} provides further insight. While average mIoU differences appear small(95.05\% for MARG+RegionMix vs. 94.65\% for Mask2Former-FreqFusion~\cite{freqfusion} and 92.38\% for BiRefNet~\cite{birefnet}), our method markedly reduces poor-quality segmentations. Only 2 of 200 test images fall below an mIoU of 0.5 with MARG+RegionMix, compared to 7 for Mask2Former-FreqFusion and 11 for BiRefNet. This robustness at the lower tail is crucial: downstream defect detection pipelines are highly sensitive to under-segmentation, as missing pixels at blade boundaries can obscure or distort small cracks. Thus, even a 1–2\% mean IoU gain translates into substantially fewer failure cases in practice.

\begin{figure*}[t!]
\centering
\begin{tabular}{@{}c@{}c}
\multicolumn{1}{c}{\hspace{-0.25cm}
  \begin{tabular}{@{\hspace{0.6cm}}c@{\hspace{1.cm}}c@{\hspace{0.7cm}}c@{\hspace{0.7cm}}c@{\hspace{0.6cm}}c}
  \tiny Input & \tiny Ground-truth & \tiny BiRefNet~\cite{birefnet} & \tiny  Mask2Former- & \tiny MARG+RegionMix
  \vspace{-0.2cm} \\
  & & & \tiny FreqFusion~\cite{freqfusion} & \vspace{0.02cm}\\
  \end{tabular}
} & \multicolumn{1}{c}{\hspace{-0.25cm}
  \begin{tabular}{@{\hspace{0.6cm}}c@{\hspace{1.cm}}c@{\hspace{0.7cm}}c@{\hspace{0.7cm}}c@{\hspace{0.6cm}}c}
  \tiny Input & \tiny Ground-truth & \tiny BiRefNet~\cite{birefnet} & \tiny  Mask2Former- & \tiny MARG+RegionMix
  \vspace{-0.2cm} \\
  & & & \tiny FreqFusion~\cite{freqfusion} & \vspace{0.02cm}\\
  \end{tabular}} \\

\includegraphics[width=0.49\linewidth]{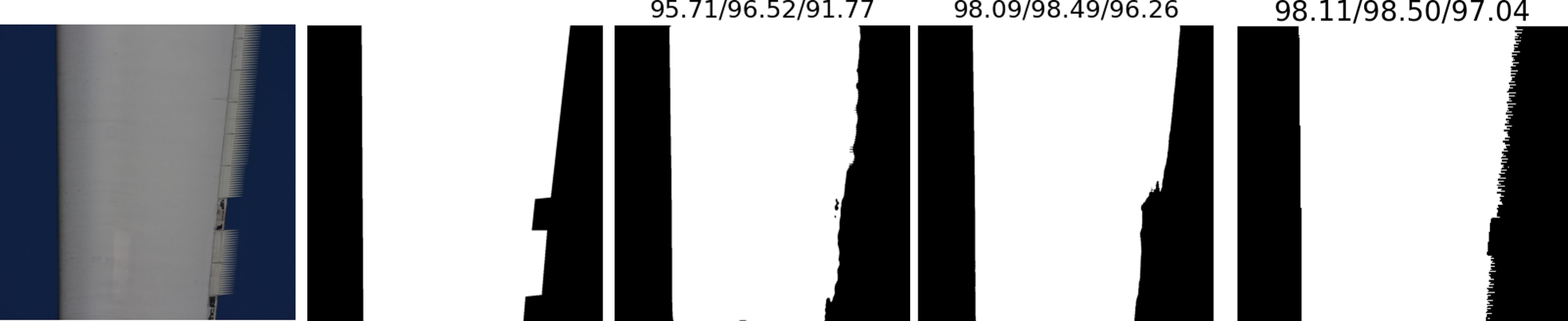}&\hspace{-0.05cm}
\includegraphics[width=0.49\linewidth]{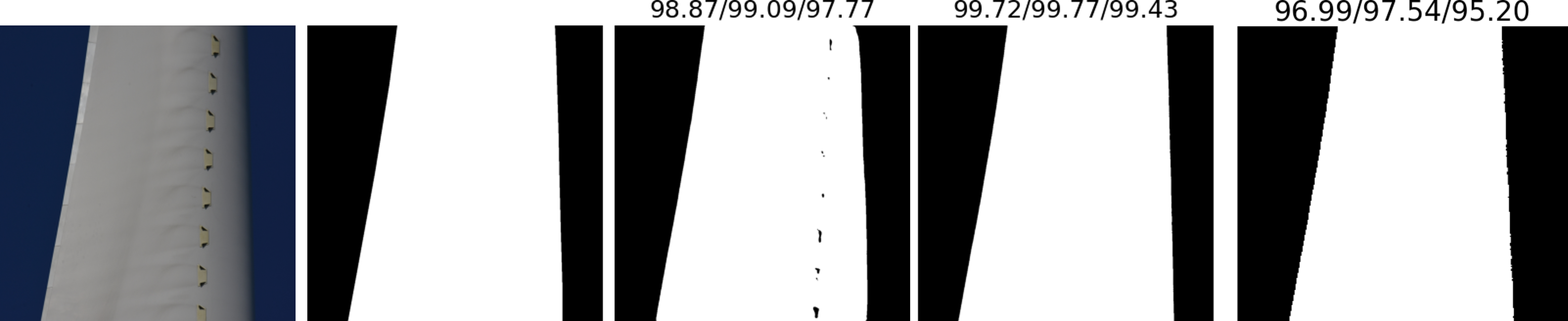}\\
\includegraphics[width=0.49\linewidth]{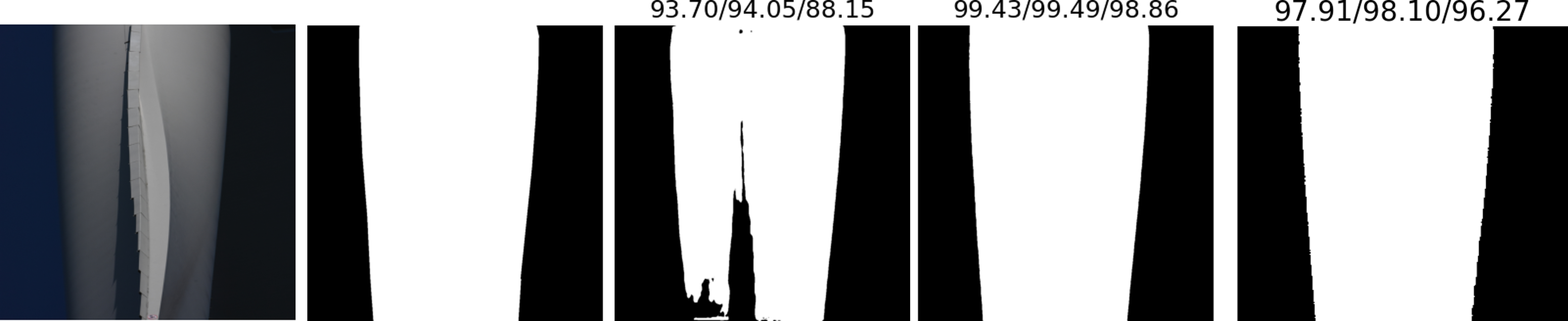}&\hspace{-0.05cm}
\includegraphics[width=0.49\linewidth]{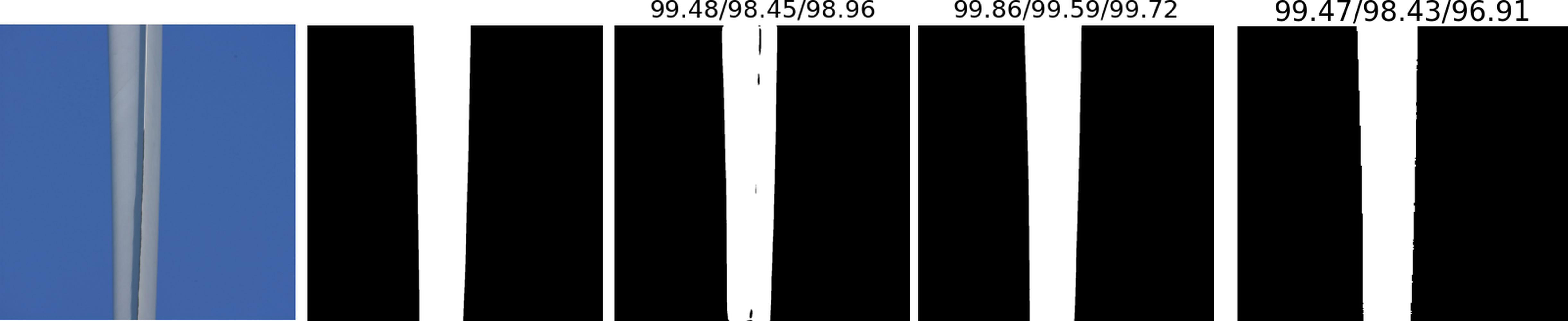}\\
\includegraphics[width=0.49\linewidth]{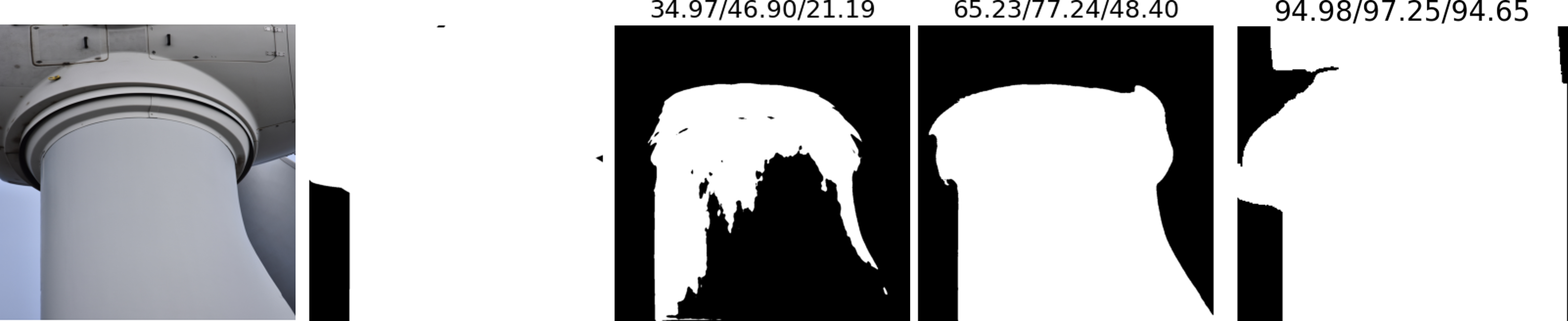}&\hspace{-0.05cm}
\includegraphics[width=0.49\linewidth]{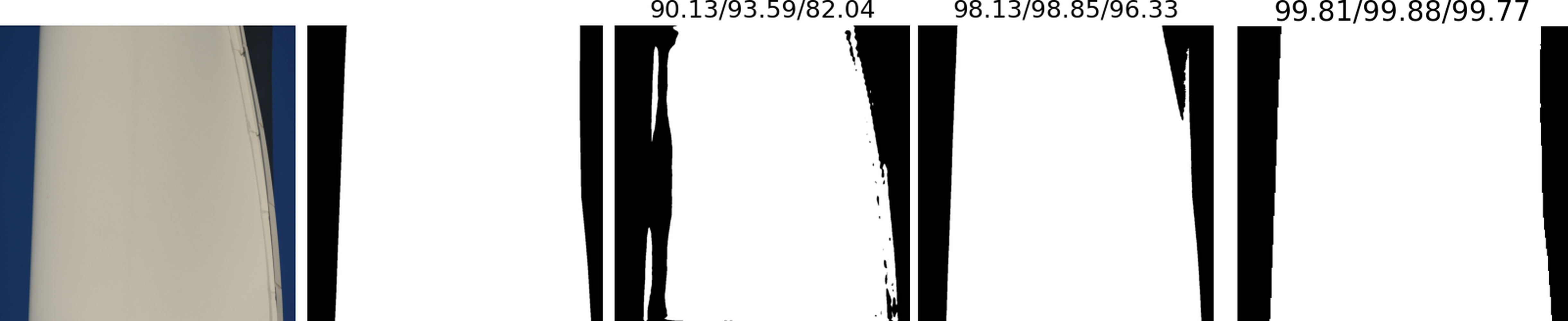}\\
\includegraphics[width=0.49\linewidth]{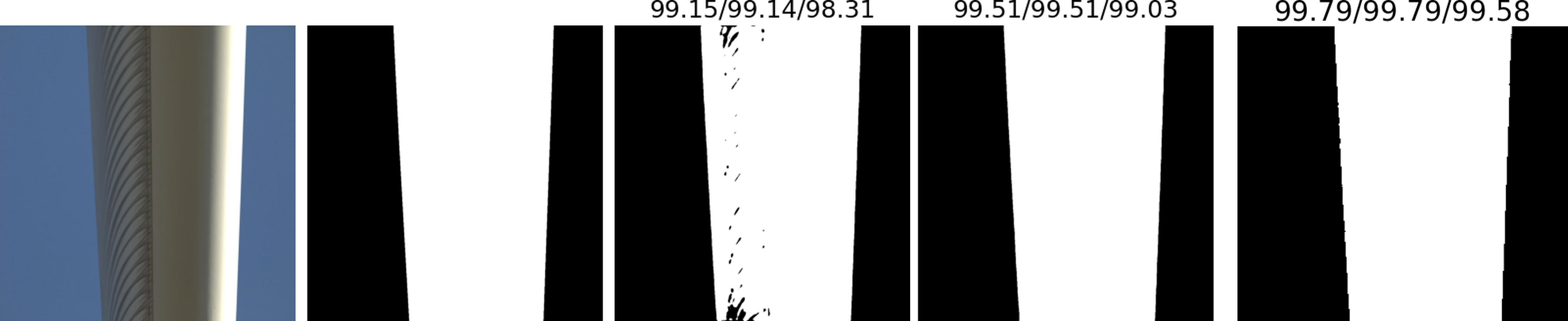}&\hspace{-0.05cm}
\includegraphics[width=0.49\linewidth]{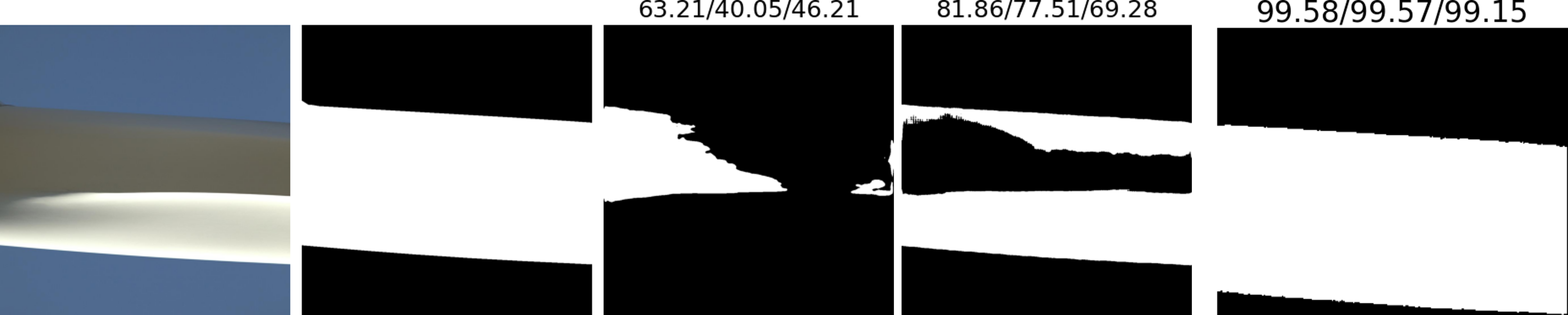}\\
\includegraphics[width=0.49\linewidth]{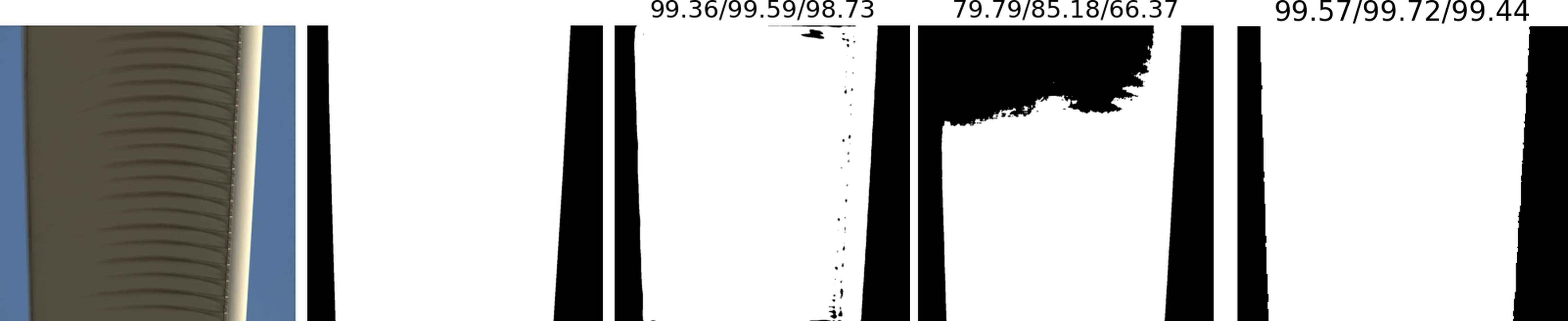}&\hspace{-0.05cm}
\includegraphics[width=0.49\linewidth]{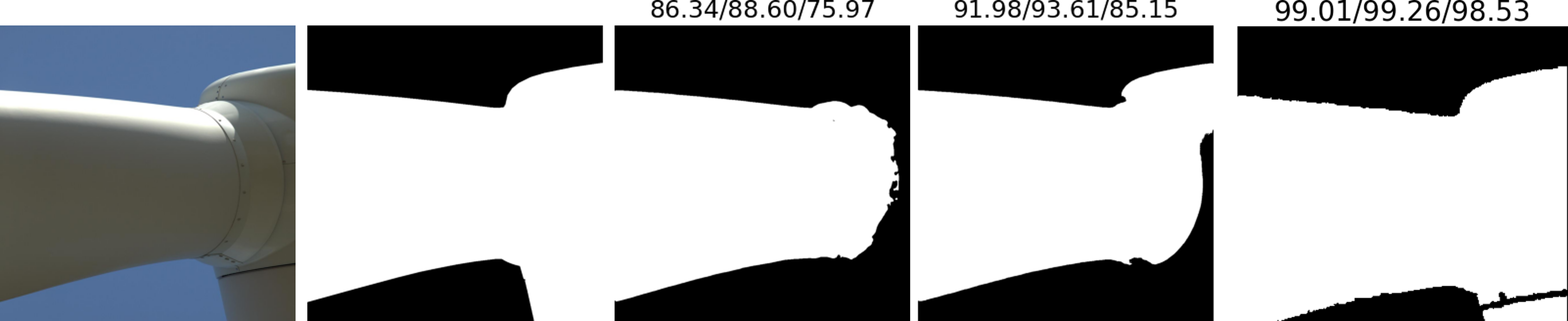}\\
\includegraphics[width=0.49\linewidth]{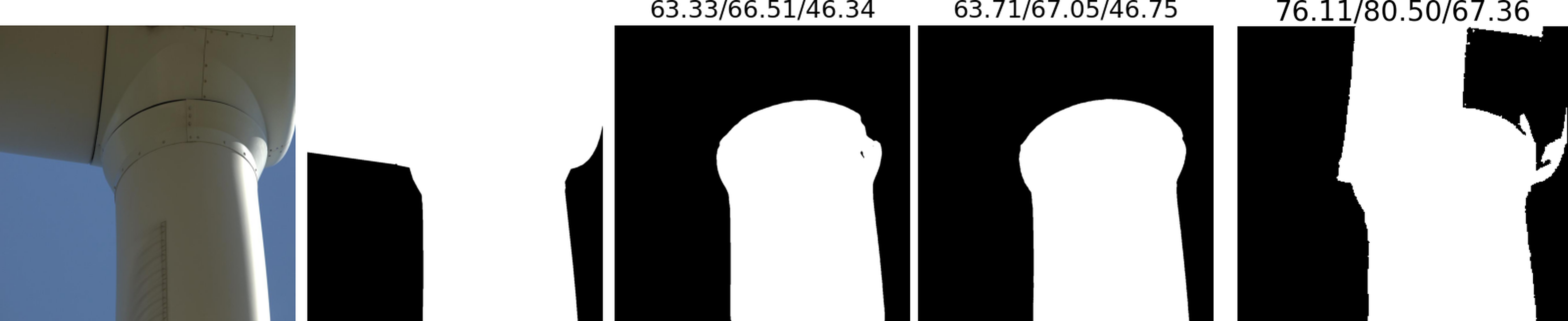}&\hspace{-0.05cm}
\includegraphics[width=0.49\linewidth]{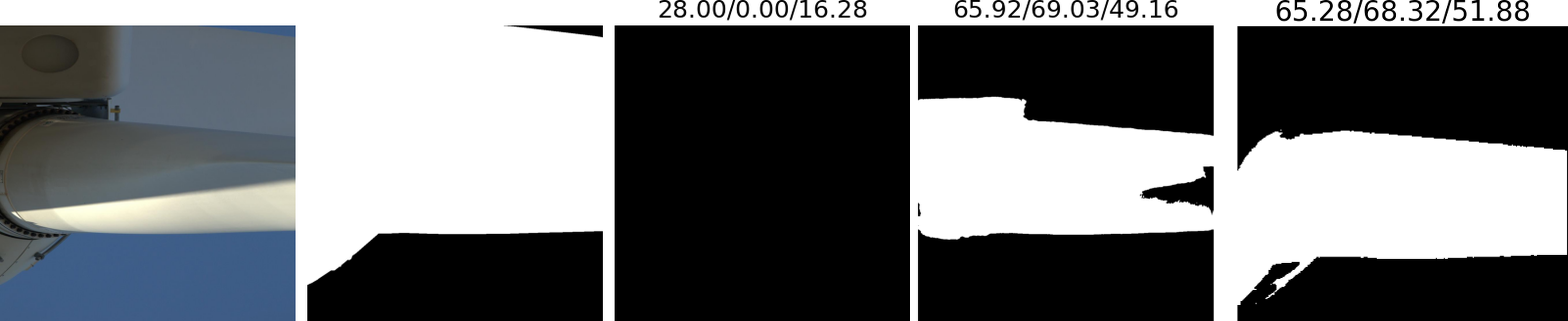}\\
\includegraphics[width=0.49\linewidth]{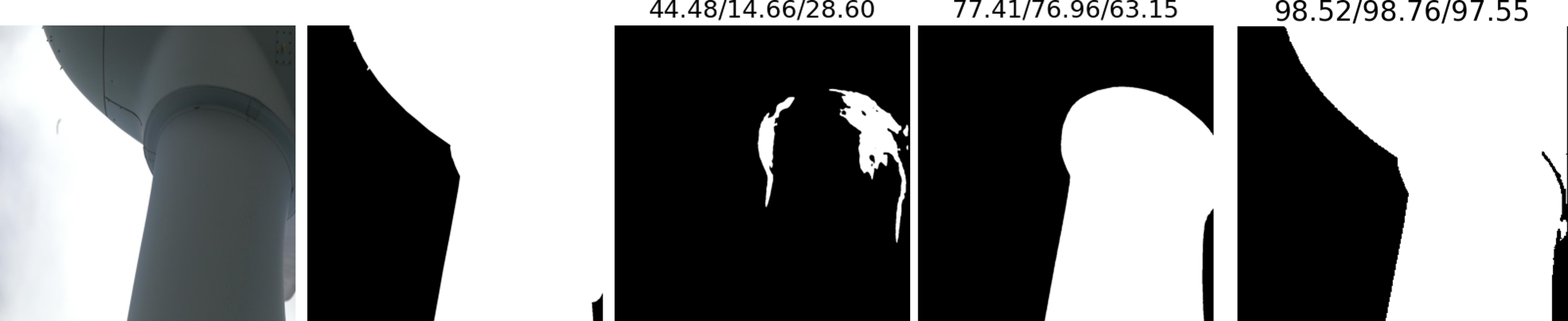}&\hspace{-0.05cm}
\includegraphics[width=0.49\linewidth]{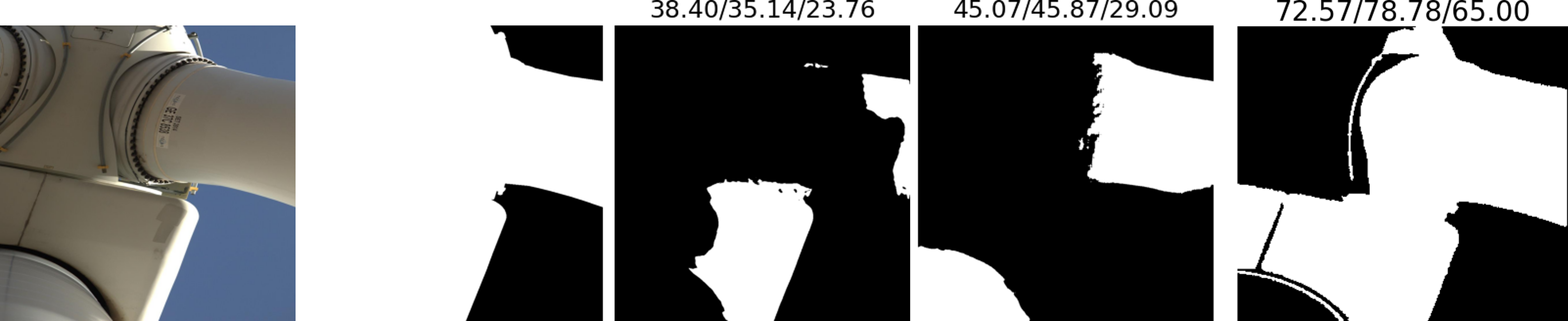}\\
\includegraphics[width=0.49\linewidth]{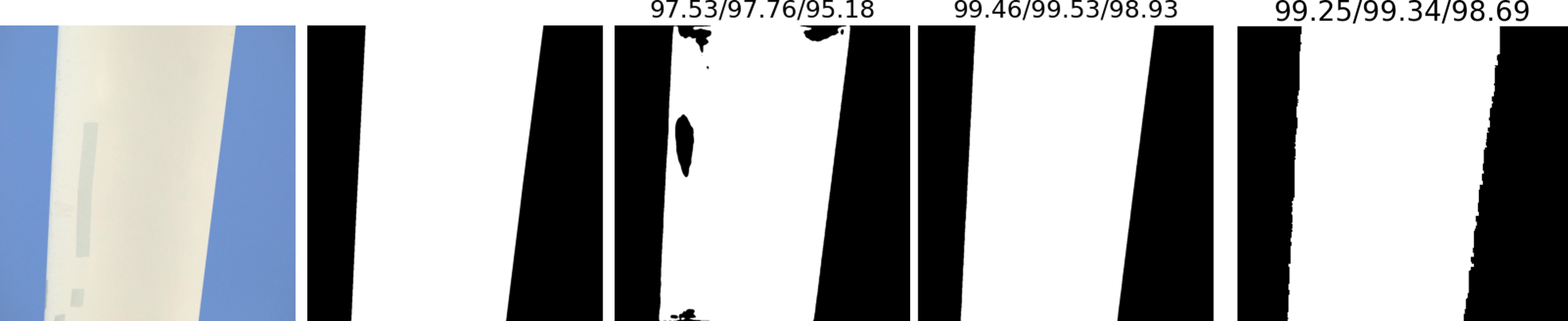}&\hspace{-0.05cm}
\includegraphics[width=0.49\linewidth]{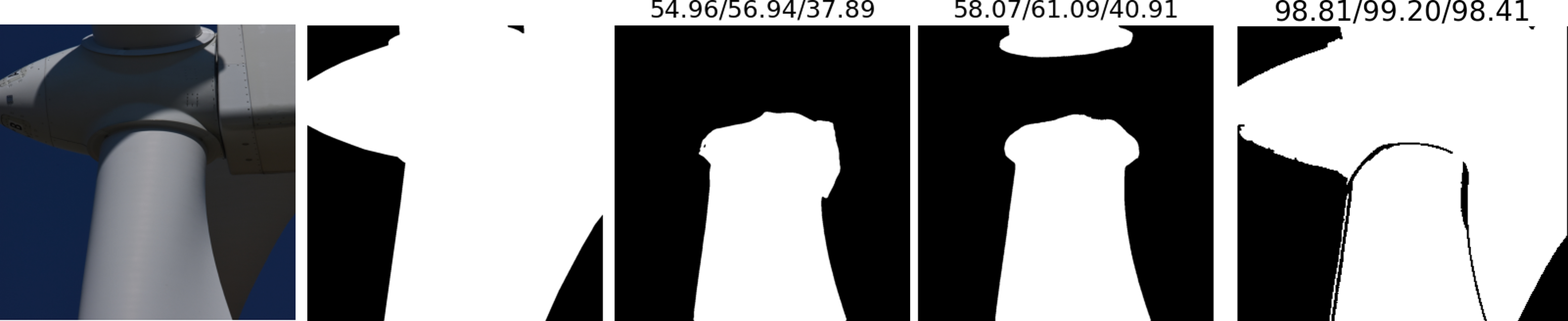}\\

\end{tabular}
\vspace{-0.25cm}
\caption{\textbf{Qualitative comparison of MARG+RegionMix with competing algorithms}. The first column presents the input color image, the second displays the ground-truth mask, and, from the third to the fifth, it shows the segmentation estimation of BiRefNet~\cite{birefnet}, Mask2Former-FreqFusion~\cite{freqfusion} and MARG+RegionMix, respectively. On top of each estimation, the accuracy, F1 and mIoU performance metrics are showcased, respectively. Both sides display the same information.} \label{fig:practical-impact}
\end{figure*}

\cref{fig:practical-impact} highlights these differences qualitatively. In challenging cases with shadows, cluttered backgrounds, or repairs, baselines often under-segment blades while still reporting high mIoU scores. For example, in the top-left case, Mask2Former-FreqFusion~\cite{freqfusion} and BiRefNet~\cite{birefnet} miss a serrated-edge defect, despite achieving 96.26\% and 91.77\% mIoU, respectively. Similarly, in the bottom-left case, BiRefNet~\cite{birefnet} fails to capture a repair patch, yet still reports 95.18\% of mIoU. By contrast, MARG+RegionMix produces more complete masks, preserving defect-relevant regions that would otherwise be lost.

In summary, although the average gains of MARG+RegionMix may appear modest, they correspond to a tangible reduction of problematic segmentations. For downstream applications such as defect detection or repair monitoring, robustness in these difficult cases is more critical than raw throughput. Moreover, our approach remains interpretable and competitive in runtime (top three on CPU), making the tradeoff justifiable for real-world use.

\section{Wind Turbine Image Segmentation Data} \label{sec:data_seg} 

The dataset used in this work is the same as that introduced in BU-Net~\cite{bunet}, and we expand its description here for clarity and completeness. It comprises color images with a resolution of $1024\times1024$, captured from different sections of the blade. The corresponding masks have been annotated manually to obtain the ground-truth solution. Each image is guaranteed to contain at least one wind turbine blade instance, as all acquisitions were carried out during targeted inspection campaigns. The dataset is therefore curated specifically for the task of wind turbine blade segmentation, ensuring relevance to real inspection scenarios.

Wind turbine blade imagery is inherently difficult to obtain due to safety restrictions, the high cost of inspections, and the sensitive nature of the data. Many images contain proprietary information about the operator's assets, such as turbine design, location, or operational conditions; a significant defect in the data could potentially compromise the reliability or confidentiality of these assets. Unlike generic segmentation benchmarks, this dataset reflects a highly specialized domain where both scarcity and confidentiality are major constraints.

Due to be taken in the wild, the images highly vary in visual appearance, leading to a challenging dataset. See some instances of the wind turbine blade segmentation data in \cref{fig:practical-impact}. The blade can diverge in size, shape or illumination conditions. Furthermore, the background is completely distinct depending on the engine used to take the picture. In general, the expected relation between the background and the blade area is 2:1, but it is worth noticing that in many cases the blade area could be very reduced or, conversely, cover almost entirely the image. 

The dataset was collected from multiple wind turbine operators and inspection companies, capturing images while the turbines were stationary to ensure high-detail acquisition. This diversity was essential to ensure generalization across geographic locations and turbine types. Two type of imaging engines were employed. The first engine is based on ground-based robotic inspections which consist of a robotic arm equipped with a high-resolution industrial camera placed in the floor. The camera was typically positioned 50 to 100 meters from the blade and primarily captures the blade against a sky background. On the other side, aerial drone images were obtained from a closer perspective - typically operated at distances between 2 and 5 meters from the blade surface. These images typically included broader landscape backgrounds, which varied depending on the wind farm's location (onshore or offshore). 

The training set is comprised of 1712 images, while the validation and test sets contain 120 and 200 images, respectively. These pictures have been gathered from different windfarms and inspection campaigns, ensuring their independence and that the test emulates new data acquired, so we can fairly analyze the generalization of our approach. More specifically, the validation and test sets are formed from randomly selecting 20 images per windfarm. Moreover, the training data was selected from a pool of different windfarms, prioritizing the under-sampled instances that are harder to infer. These images cover the blade sections located at the root, tip or max-cord, commonly known as the shoulder of the blade. In this way, we ensure that we capture the structural diversity of blades, along its different in geometry and appearance.